%% file: main.tex
\documentclass[11pt]{article}

\usepackage[preprint]{acl}

\usepackage{times}
\usepackage{latexsym}

\usepackage[T1]{fontenc}

\usepackage[utf8]{inputenc}

\usepackage{microtype}

\usepackage{inconsolata}

\usepackage{graphicx}

\usepackage{amsmath}
\usepackage{amssymb}
\usepackage{xcolor}
\usepackage{pgfplots}
\pgfplotsset{compat=1.18}
\usepackage{pgfplotstable}
\usepackage{tikz}
\usepackage{caption}
\usepgfplotslibrary{groupplots}
\definecolor{skyblue}{RGB}{135,206,235}
\definecolor{salmon}{RGB}{250,128,114}
\colorlet{skyblue_trans}{skyblue!90}
\colorlet{salmon_trans}{salmon!60}
\usepackage{tcolorbox}
\usepackage{algorithmic}
\usepackage{algorithm}
\usepackage{booktabs}
\usepackage{arydshln}
\usepackage{subcaption}
\usepackage{enumitem}

\title{Can LLMs Track Their Output Length?\\A Dynamic Feedback Mechanism for Precise Length Regulation}

\author{
 \textbf{Meiman Xiao\textsuperscript{1}}\thanks{These authors contributed equally.},
 \textbf{Ante Wang\textsuperscript{1}}\footnotemark[1],
 \textbf{Qingguo Hu\textsuperscript{1}}\footnotemark[1],
 \textbf{Zhongjian Miao\textsuperscript{3}}, 
 \textbf{Huangjun Shen\textsuperscript{1}}
 \\
 \textbf{Longyue Wang\textsuperscript{2}},
 \textbf{Weihua Luo\textsuperscript{2}},
 \textbf{Jinsong Su\textsuperscript{1}}\thanks{Corresponding author.}
\\
 \textsuperscript{1}School of Informatics, Xiamen University, China \\
 \textsuperscript{2}Alibaba International Digital Commerce Group~~\textsuperscript{3}Li Auto Inc.
 \\
\texttt{\small{\{xiaomeiman,wangante,huqingguo\}@stu.xmu.edu.cn}}~~\texttt{\small{jssu@xmu.edu.cn}}
}

\begin{document}
\maketitle
\input{section/0_abs}
\input{section/1_intro}
\input{section/2_pre}
\input{section/3_method}
\input{section/4_exp}
\input{section/5_related}
\input{section/6_conclu}

\section*{Limitations}
This study incorporates length feedback from an external counter to improve LLMs’ ability to regulate output length. Theoretically, this introduces only minor latency due to the efficiency of length computation; however, it interrupts the batched generation process, which can lead to latency issues in large-scale deployment. One promising solution is to selectively decide when to invoke the length counter and to optimize for efficacy~\cite{xu2025alignment}, which we leave for future work.

\bibliography{custom}

\input{section/7_appendix}

\end{document}

%% file: section/0_abs.tex
\begin{abstract}
Precisely controlling the length of generated text is a common requirement in real-world applications. However, despite significant advancements in following human instructions, Large Language Models (LLMs) still struggle with this task.
In this work, we demonstrate that LLMs often fail to accurately measure their response lengths, leading to poor adherence to length constraints. To address this issue, we propose a novel length regulation approach that incorporates dynamic length feedback during generation, enabling adaptive adjustments to meet target lengths.
Experiments on summarization and biography tasks show our training-free approach significantly improves precision in achieving target token, word, or sentence counts without compromising quality. Additionally, we demonstrate that further  supervised fine-tuning allows our method to generalize effectively to broader text-generation tasks.
\end{abstract}

%% file: section/1_intro.tex
\section{Introduction}

The use of products built upon Large Language Models (LLMs,~\citealt{achiam2023gpt,team2023gemini,yang2025qwen3}) to process various text generation tasks has grown prevalent in daily human study and work. Benefiting from pretraining on massive corpora and subsequent post-training, such as Supervised Fine-Tuning (SFT,~\citealt{taori2023alpaca,chiang2023vicuna}) and Reinforcement Learning (RL,~\citealt{ouyang2022training,bai2022training}), current LLMs demonstrate strong text comprehension and instruction-following capabilities, enabling them to effectively handle diverse user requests.

Despite their significant progress, it remains challenging for LLMs to precisely control their outputs according to specific requirements. For instance, when given a query such as ``Please write a 500-word composition,'' current LLMs often struggle with strictly adhering to the length constraint.
To address this issue, previous research~\cite{jie_prompt-based_2024,yuan2024following,li2024ruler,butcher_precise_2025} has trained LLMs to produce outputs that adhere to specified length requirements through SFT or advanced RL algorithms.
However, these approaches often demand massive training data to cover a wide range of length preferences, and may degrade model performance after intensive post-training. More critically, these techniques are typically tailored to a single type of length constraint, hindering generalization to different types of length control (e.g., tokens, words, or sentences).

\begin{figure}[t]
\centering
\includegraphics[width=\linewidth, trim={0.29in 0.98in 0.20in 0.0in}, clip]{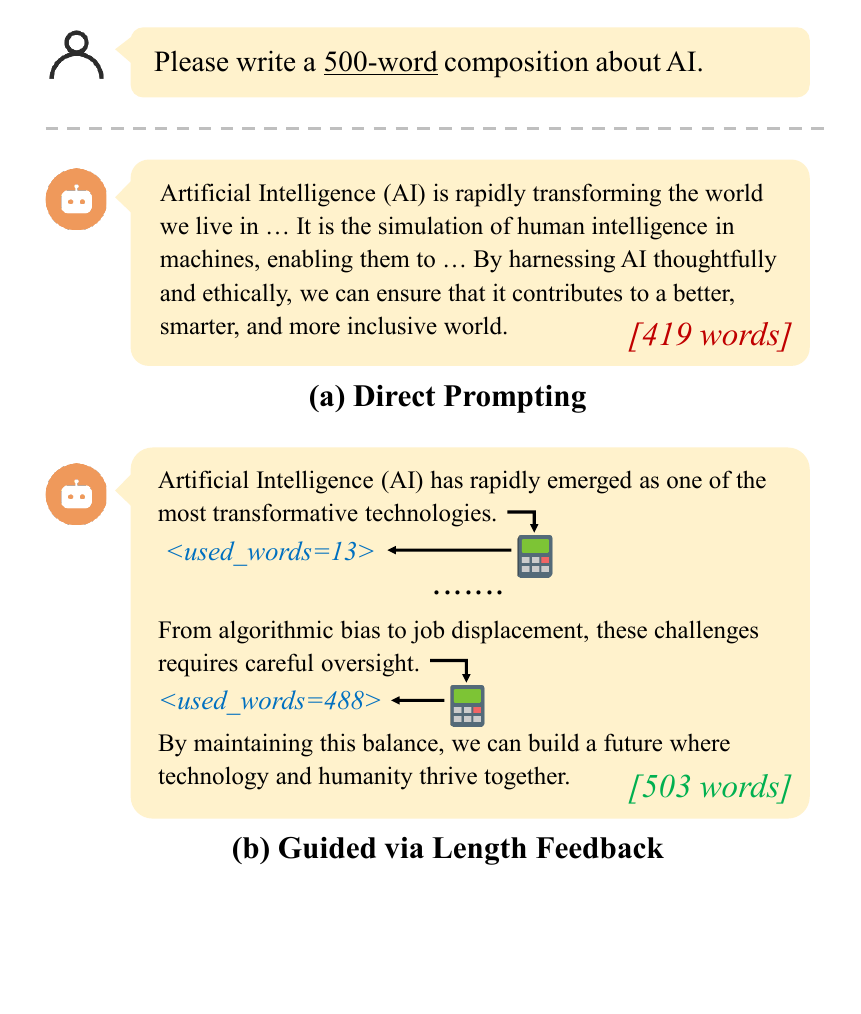}
\caption{A comparison between the conventional prompt-based approach and our feedback-guided approach for controlling text generation length.}
\label{fig:example}
\end{figure}

Intuitively, an LLM should track its output length to precisely control the length of its generated content.
To explore this issue further, we first conduct a preliminary study to investigate: \textit{Do LLMs know how long their generated texts are?} Our results indicate that while LLMs can accurately measure the length of short texts, this ability declines significantly as text length increases. This deficiency mirrors human tendencies: when writing, people typically focus on content creation and planning rather than precisely tracking exact length. However, unlike LLMs, humans can leverage tools such as text editors to obtain external feedback for monitoring counts, which frees them from manual estimation and allows them to use this feedback to adjust their strategies for subsequent content.

Inspired by these findings, we explore the use of length feedback to help LLMs track and dynamically adjust their output length.
Previous research~\cite{markergen} successfully incorporated word-level length guidance through a three-stage pipeline, inserting length feedback after tokens with an exponential decay strategy. However, this method incurs high inference overhead, and its iterative process can lead to error propagation, potentially harming response quality.
In contrast, our approach has LLMs generate text in one pass but intermittently interrupts the generation process to insert feedback. To ensure the LLM understands the feedback, we enrich the prompt with explanations of its meaning and usage. To minimize disruption, we insert feedback only at sentence boundaries, thereby preserving semantic coherence.
More importantly, our method offers flexible adaptation to different length constraints by simply adjusting the feedback style and input prompt. A comparison with the conventional prompt-based approach is shown in Figure~\ref{fig:example}.

We first validate our method on summarization~\cite{huang-etal-2021-efficient} and biography generation~\cite{min2023factscore} tasks.
Without specialized fine-tuning, our method demonstrates superior length control across different constraint types when applied to LLaMA-3.1-8B-Instruct~\cite{dubey2024llama} and Qwen3~\cite{yang2025qwen3} family models. Evaluations based on DeepSeek-V3~\cite{deepseekai2024deepseekv3technicalreport} indicate that the text quality remains robust while meeting these constraints.

To assess our method’s effectiveness on general-domain tasks, we further conduct experiments on ELI5~\cite{eli5}, a benchmark for long-form question answering. While our training-free approach can better meet length requirements, it may compromise quality for tasks requiring answers within a strict length range. However, by applying SFT only to LLM-generated responses and without any data filtering, our method effectively maintains performance while enhancing length control. Further analysis shows that our method achieves at least 4$\times$ greater training efficiency than baselines without feedback incorporation, avoiding performance degradation from excessive fine-tuning.

%% file: section/2_pre.tex
\section{Preliminary Study}
\label{sec:preliminary_study}
Generating text with a desired length inherently requires the model to accurately track its output length, yet the LLM's ability to estimate length remains unclear.
In this section, we design experiments to investigate this question: \textit{Can LLMs accurately estimate the length of their generated texts?}
Specifically, we first collect responses of varying lengths from several mainstream LLMs and evaluate their length-tracking ability by prompting them to estimate the lengths of their own outputs (\S \ref{pilot_setup}). We then present the evaluation results and highlight the key findings (\S \ref{pilot_result}).

\input{table/length_estimation}

\subsection{Experimental Setup}
\label{pilot_setup}

\paragraph{Data Preparation}  
Following prior work~\cite{jie_prompt-based_2024,butcher_precise_2025}, we conduct experiments on the summarization task, where length constraints are commonly specified.
We use the GovReport dataset~\cite{huang-etal-2021-efficient} and evaluate the following LLMs: LLaMA-3.1-8B-Instruct~\cite{dubey2024llama}, as well as Qwen3-4B and Qwen3-8B~\cite{yang2025qwen3}, chosen for their strong reasoning capabilities on challenging tasks.
To obtain responses of varying lengths, we prompt the LLMs to generate summaries for each instance with a randomly selected target length between 100 and 400 tokens.
In total, 1,000 generated summaries are collected.

\paragraph{Length Estimation}
To evaluate the LLMs' ability to track output length, we feed each generated summary back into the same model and prompt it to estimate its length.
Specifically, we use the prompt in Figure~\ref{fig:token_count} to ask the LLM for measuring the output length in tokens.
We use the Mean Absolute Error (MAE) between the LLM-estimated and actual generated length to quantify the LLMs' prediction errors in counting output length. Additionally, we report the MAE between the LLM-generated and user-specified lengths to provide a more direct assessment of the LLMs' length controlling capability.

\subsection{Results and Analysis}
\label{pilot_result}

Results in Figure~\ref{fig:pilot_study} offer several key observations.

\paragraph{LLMs perform worse when counting longer texts}
Considering the MAE between the LLM-estimated and generated text length, we find a strong positive correlation between prediction errors and text length across all three LLMs.
This suggests that as text length increases, LLMs' estimations deviate more significantly from the true length, revealing a fundamental limitation in their length-tracking capability. These trends also persist across other types of length constraints (e.g., words and sentences). Detailed results are provided in the Appendix~\ref{appendix:preliminary}.

\paragraph{Length control is closely correlated with length estimation}
We observe the same trend when comparing the prediction errors of LLM-generated and LLM-estimated lengths against the corresponding target length.
For the Qwen3 series models, those that achieve more accurate length estimation also tend to perform better in length control.
This conclusion holds across different types of length constraints.
For example, when estimating output length, LLaMA-3.1-8B-Instruct performs significantly better at word counting than token counting, leading to greater compliance with word-length constraints over token-based ones in generation.
These results underscore the importance of length tracking in length control.

%% file: table/length_estimation.tex
\newcommand{\sevenpt}{\fontsize{7.6pt}{9pt}\selectfont}
\pgfplotsset{compat=1.3,
    /pgfplots/ybar legend/.style={
    /pgfplots/legend image code/.code={%
       \draw[##1,/tikz/.cd,yshift=-0.25em]
        (0cm,0cm) rectangle (7pt,0.8em);},
   },
}
\begin{figure*}[t]
\hspace*{-0.5cm}
\pgfplotsset{width=6.5cm, height=4.5cm}
\centering
\scalebox{0.90}{\begin{tikzpicture}
\begin{groupplot}[
    group style={
        group name=my plots,
        group size=3 by 1,
        horizontal sep=0.8cm,
    },
]

\nextgroupplot[
    ybar,
    ymin=0,
    ymax=135,
    ytick={0,30,...,120},
    ylabel={MAE~$\downarrow$},
    y label style={yshift=-0.6em},
    symbolic x coords={{[50,150)}, {[150,250)}, {[250,350)}, {[350,450)}},
    xticklabel style={font=\sevenpt},
    xtick=data,
    enlarge x limits=0.2,
    xtick pos=bottom,
    ytick pos=left,
    bar width=11pt,
    title style={yshift=-11.50em, font=\small},
    title={(a) Qwen3-4B},
    axis lines*= left,
]
\addplot[color=purple!80!black, fill=purple!20!white] coordinates {
    ({[50,150)},28.23) ({[150,250)},56.89) ({[250,350)},73.30) ({[350,450)},94.82)
};
\addplot[color=teal!80!black, fill=teal!20!white] coordinates {
    ({[50,150)},20.40) ({[150,250)},56.01) ({[250,350)},56.28) ({[350,450)},69.99)
};

\nextgroupplot[
    ybar,
    ymin=0,
    ymax=135,
    ytick={0,30,...,120},
    symbolic x coords={{[50,150)}, {[150,250)}, {[250,350)}, {[350,450)}},
    xticklabel style={font=\sevenpt},
    xtick=data,
    enlarge x limits=0.2,
    xtick pos=bottom,
    ytick pos=left,
    bar width=11pt,
    title style={yshift=-11.50em, font=\small},
    title={(b) Qwen3-8B},
    axis lines*= left,
    legend style={
            at={(0.5,1)},
            anchor=south,
            column sep=0.5ex,
            /tikz/every even column/.append style={column sep=1.5em},
            font=\small,
            draw=none,
            legend columns=2,
        },
]
\addplot[color=purple!80!black, fill=purple!20!white]  coordinates {
    ({[50,150)},26.81) ({[150,250)},38.68) ({[250,350)},52.58) ({[350,450)},108.86)
};
\addplot[color=teal!80!black, fill=teal!20!white] coordinates {
    ({[50,150)},21.43) ({[150,250)},44.24) ({[250,350)},49.03) ({[350,450)},52.89)
};
\legend{Estimated vs. Generated, Generated vs. Specified}

\nextgroupplot[
    legend style={at={(0.5,-0.35)}, anchor=north, legend columns=2},
    ybar,
    ymin=0,
    ymax=135,
    ytick={0,30,...,120},
    symbolic x coords={{[50,150)}, {[150,250)}, {[250,350)}, {[350,450)}},
    xticklabel style={font=\sevenpt},
    xtick=data,
    enlarge x limits=0.2,
    xtick pos=bottom,
    ytick pos=left,
    bar width=11pt,
    title style={yshift=-11.50em, font=\small},
    title={(c) LLaMA-3.1-8B-Instruct},
    axis lines*= left
]
\addplot[color=purple!80!black, fill=purple!20!white] coordinates {
    ({[50,150)},27.33) ({[150,250)},41.20) ({[250,350)},58.85) ({[350,450)},79.71)
};
\addplot[color=teal!80!black, fill=teal!20!white] coordinates {
    ({[50,150)},30.37) ({[150,250)},54.40) ({[250,350)},94.96) ({[350,450)},122.86)
};

\end{groupplot}
\end{tikzpicture}}
\caption{Mean Absolute Error (MAE) between (1) the estimated and actual generated length, and (2) the generated and user-specified length, across token ranges for (a) Qwen3-4B, (b) Qwen3-8B, or (c) LLaMA-3.1-8B-Instruct.}
\label{fig:pilot_study}
\end{figure*}

%% file: section/3_method.tex
\section{Methodology}

In our preliminary study, we observe that existing LLMs struggle to accurately track their output length, resulting in poor control over text generation.
To overcome this challenge, we consider how humans handle similar length control problems during writing. Rather than relying on manual estimation, humans typically use external tools such as text editors to precisely monitor the length of their ongoing work. Abandoning inefficient multi-stages used in Markergen, we propose a novel length regulation approach that integrates length feedback from an external counter at the end of sentences to help LLMs track the length of their generated output more effectively and efficiently.

As shown in Figure~\ref{fig:example}(b), we intermittently interrupt the LLM’s generation to insert length feedback. Formally, given an unfinished output \(\mathbf{\hat{y}}\), we append a feedback string (e.g., ``\textit{\text{\textless}used\_tokens=\(f(\mathbf{\hat{y}})\)\text{\textgreater}}'') before resuming generation, where \(f(*)\) is a predefined length calculation function. Depending on the desired length constraint type (e.g., tokens, words, sentences), \(f(*)\) can be easily implemented using the model's tokenizer or external tools like \texttt{NLTK}~\cite{loper2002nltk}. Specifically, during generation, we detect sentence boundaries and invoke the length-calculation tool based on the specified constraint type. This design minimizes disruption to the natural flow of generation while maintaining semantic coherence.

To ensure the LLM correctly interprets this length feedback, we augment the prompt with instructions explaining its meaning and usage, as illustrated below:

\begin{tcolorbox}[  
    colback=gray!5!white,   
    colframe=gray!50!black,   
    coltitle=white,  
    fonttitle=\bfseries\normalsize,  
    fontupper=\small,  
    title=Prompt of Token-Counting Tool Usage  
]
You can accurately calculate your output length using a provided tool. To call the tool, please output ``\texttt{<used\_tokens=}'' and the number of generated tokens will be automatically computed.
\end{tcolorbox}
Leveraging the strong comprehension capabilities of LLMs, we find that they can correctly interpret the feedback string without disrupting the natural generation fluency. In experiments, we further construct a feedback-augmented dataset and demonstrate that SFT enables LLMs to actively use the tool and further enhances response quality.

%% file: section/4_exp.tex
\section{Experiments}

In this section, we first detail the experimental setup (\S \ref{sec:setup}). Next, we validate our training-free approach on specific tasks (\S \ref{sec:train_free}) and finally extend our method to the general domain via SFT (\S \ref{sec:train_based}).

\subsection{Setup}
\label{sec:setup}

\paragraph{Datasets}
\label{sec:dataset}  

We evaluate our approach across three different models on two tasks suited for length constraints, as well as a general-domain long-form generation task, to assess method robustness.

\begin{itemize}[leftmargin=*]
    \item \textbf{GovReport} \cite{huang-etal-2021-efficient}: A long-document summarization dataset of 973 U.S. government reports with expert-written abstractive summaries for testing. Unlike conventional short-form summarization datasets \cite{cohan-etal-2018-discourse, kornilova-eidelman-2019-billsum, sharma-etal-2019-bigpatent}, GovReport allows better evaluation across various target lengths.

    \item \textbf{Biographies} \cite{min2023factscore}: A biography generation dataset of 183 individuals for testing, originally proposed to evaluate the factuality of LLM-generated text. Unlike summarization, this task relies on LLMs to generate content based on their parametric knowledge.

    \item \textbf{ELI5} \cite{eli5}: A general-domain long-form question-answering dataset.
    We use a publicly available dataset\footnote{\url{https://huggingface.co/datasets/sentence-transformers/eli5}}, randomly sampling 10,240 instances for training and 500 for testing.
\end{itemize}

\input{table/main}

\paragraph{Models and Hyperparameters}
We use the same models as in the preliminary study (\S \ref{sec:preliminary_study}): Qwen3-4B, Qwen3-8B, and LLaMA-3.1-8B-Instruct. For all experiments, we conduct inference using the \texttt{vLLM} engine \cite{kwon2023efficient}, which provides high-throughput serving for LLMs. We employ nucleus sampling with temperature set to 0.8 and top-p value of 0.95 for text generation.
We define target lengths across three linguistic granularities: token-level (100 to 400 tokens, in increments of 50), word-level (100 to 400 words, in increments of 50), and sentence-level (5 to 30 sentences, in increments of 5).
To quantify generation length across different linguistic units, we use each model's native tokenizer for token-level counts and the \texttt{NLTK}~\cite{loper2002nltk} for word- and sentence-level measurements.

\paragraph{Evaluation Metrics}
\label{sec:main_metrics}
Following prior studies \cite{butcher_precise_2025, li2024ruler}, we evaluate length‐generation error and text quality at these target lengths using the metrics defined below.

\begin{itemize}[leftmargin=*]
    \item \textbf{Mean Absolute Error (MAE)}: This metric computes the average absolute difference between the predicted length \(\hat{L}_i\) and the ground-truth user-specified length \(L_i\) across all \(N\) samples:  
    \begin{equation}  
    \text{MAE} = \frac{1}{N} \sum_{i=1}^{N} \left| \hat{L}_i - L_i \right|.  
    \end{equation}  

    \item \textbf{Precise Match (PM)}: This metric calculates the fraction of predictions whose absolute error falls within a predefined tolerance threshold \(\epsilon\):  
    \begin{equation}  
    \text{PM} = \frac{1}{N} \sum_{i=1}^{N} \mathbb{I}\left( \left| \hat{L}_i - L_i \right| \leq \epsilon \right),  
    \end{equation}  
    where \(\mathbb{I}(\cdot)\) denotes the indicator function.  
    In our experiments, we set \(\epsilon = 10\) for token and word counts, and \(\epsilon = 0\) for sentence-level measurements.  
    
    \item \textbf{Quality Score (Qual.)}: 
    We adopt the LLM-as-a-Judge paradigm~\cite{zheng2023judgingllmasajudgemtbenchchatbot}, using DeepSeek-V3\footnote{DeepSeek-V3-0324} across all datasets for more accurate assessments.
    Specifically, for GovReport, we follow~\citet{liu-etal-2023-g} to evaluate summaries on three dimensions: coherence, consistency, and relevance. For Biographies, we assess coherence and factual accuracy. For ELI5, we follow ~\citet{zheng2023judgingllmasajudgemtbenchchatbot} by focusing on correctness and helpfulness. To manage costs, for each linguistic unit and specified length, we randomly select 100 instances for summary or biography evaluation and 200 instances for general-domain text generation. To ensure reliability, we further conduct a human evaluation (Appendix~\ref{appendix:human_evaluation}), which validates strong consistency between human and model judgments. The \textbf{Quality Score} is reported as the average of the relevant metrics for each dataset.
\end{itemize}

\subsection{Results on Training-Free Length Control}
\label{sec:train_free}
Given the high cost of fine-tuning LLMs, we first compare training-free approaches. In addition to conventional prompting with target length constraints (Baseline), we investigate the effectiveness of in-context learning (ICL), which exhibits strong performance across various tasks without task-specific fine-tuning~\cite{wies2023learnability,dong2024survey}.
To implement this, we collect instances of varying lengths and construct a one-shot demonstration for each instance.
Besides, we implement Markgen~\cite{markergen} as a competitive baseline, given that its official code is not publicly available.

\begin{figure*}[t!]
\centering
    \begin{subfigure}{.340\textwidth}
        \includegraphics[width=\linewidth]{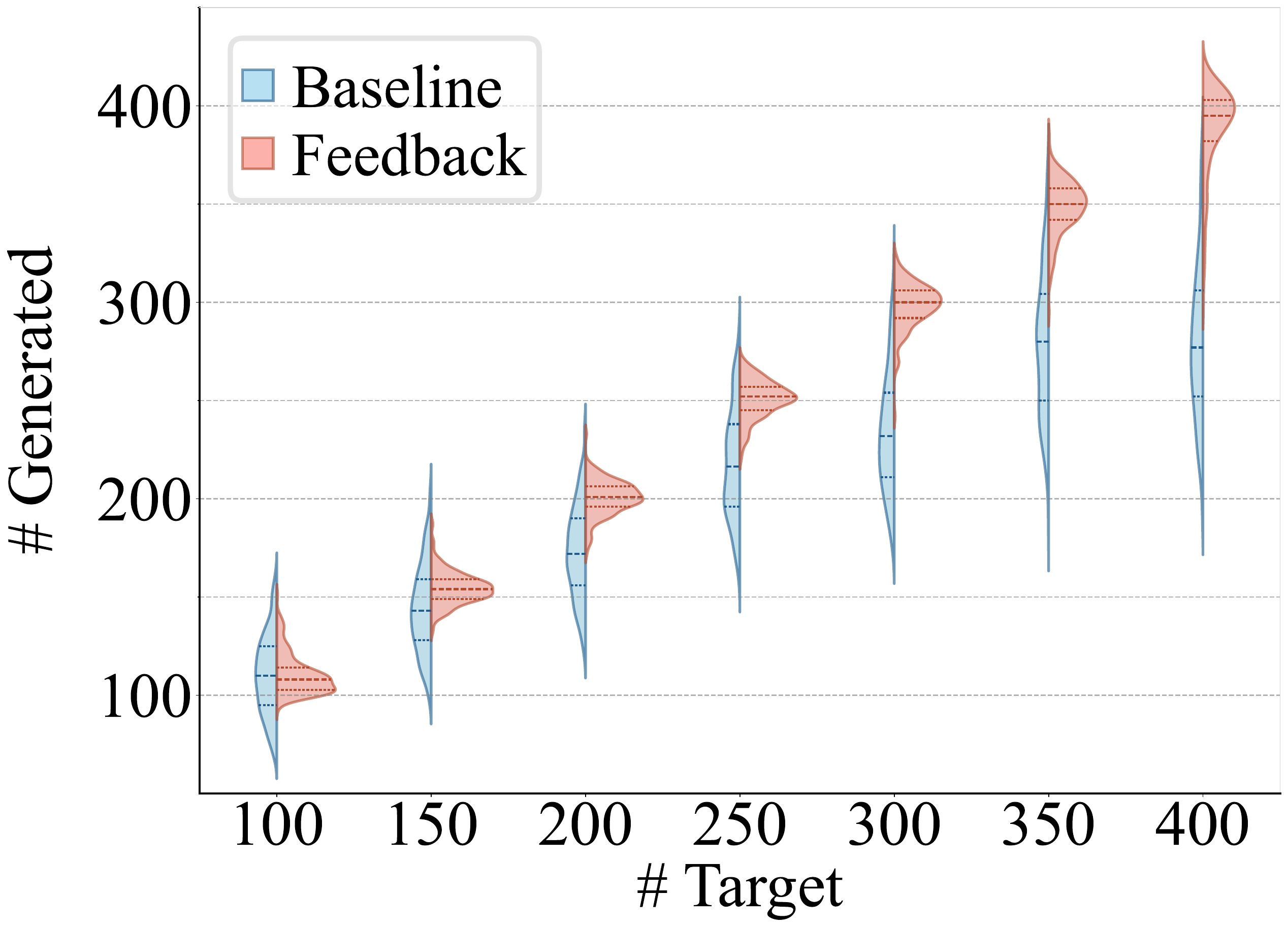} 
        \caption{Token}
        \label{subfig:summary_prompt_a}
    \end{subfigure}
    \hfill
    \begin{subfigure}{.32\textwidth}
        \includegraphics[width=\linewidth]{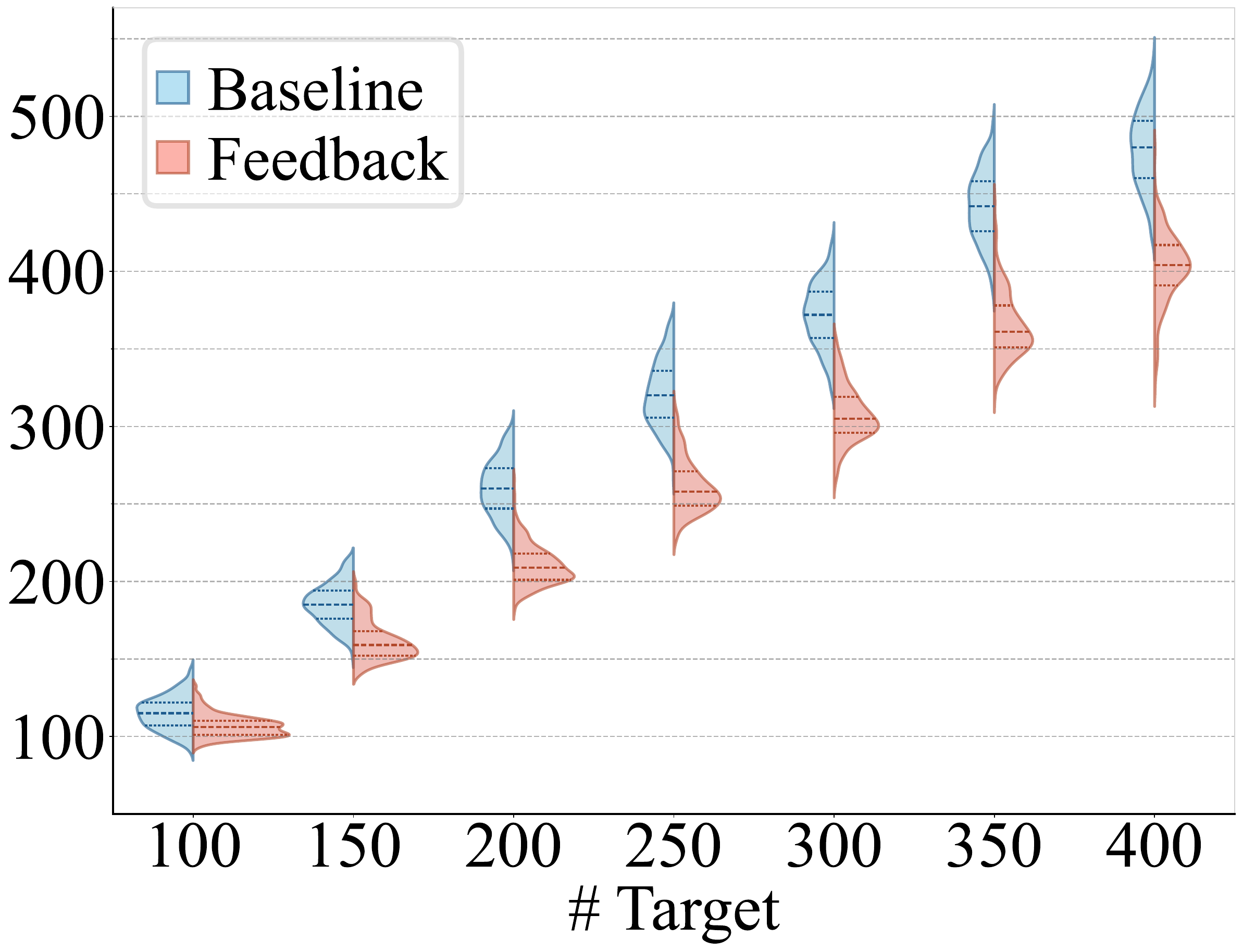} 
        \caption{Word}
        \label{subfig:summary_prompt_b}
    \end{subfigure}
    \hfill
    \begin{subfigure}{.32\textwidth}
        \includegraphics[width=\linewidth]{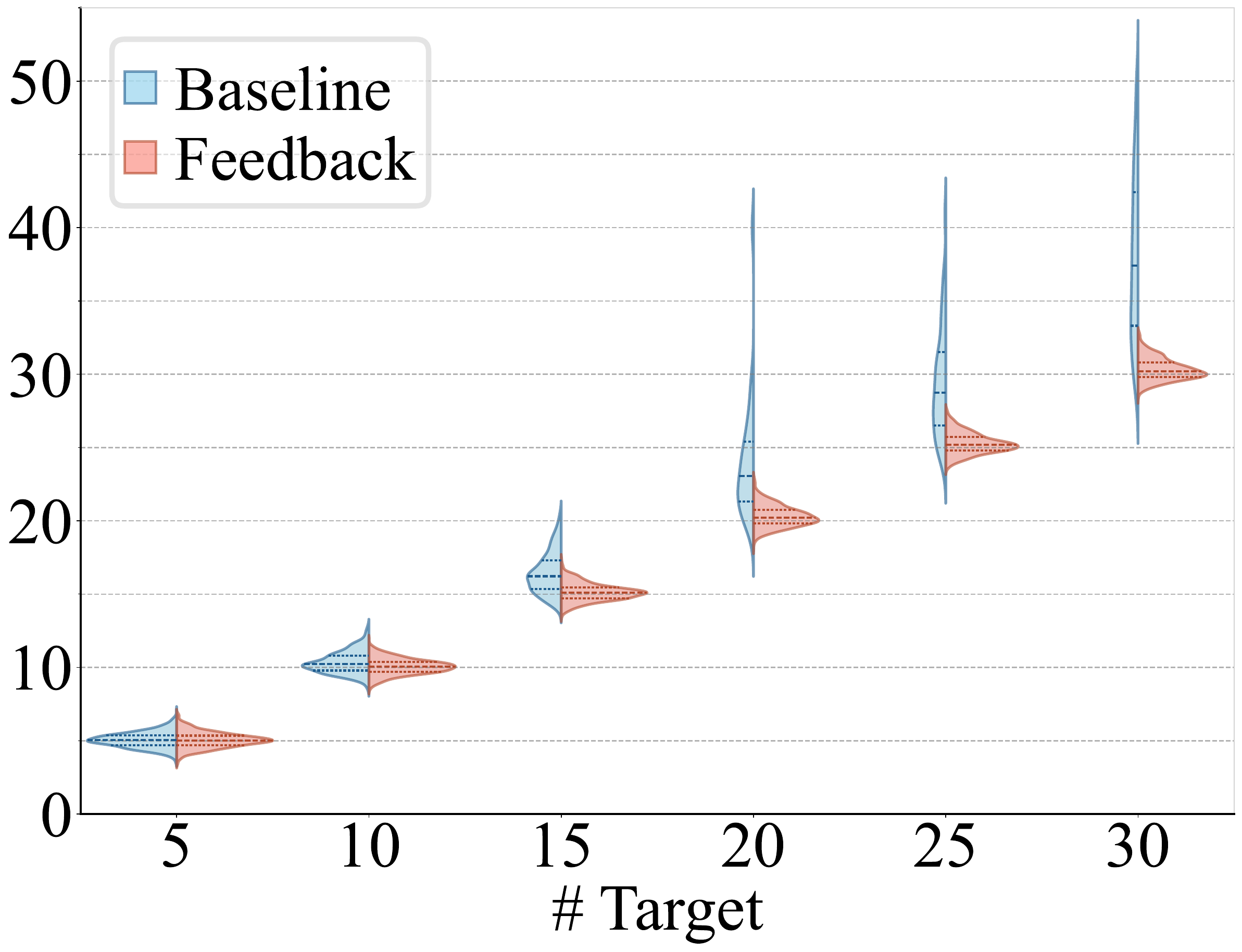} 
        \caption{Sentence}
        \label{subfig:summary_prompt_c}
    \end{subfigure}
    \caption{Generated length distributions under varying target lengths on GovReport using Qwen3-8B, comparing Baseline and Feedback at (a) token-level, (b) word-level, and (c) sentence-level granularities. Due to space constraints, please refer to the Appendix~\ref{appendix:train-free} and~\ref{appendix:train-based} for results on biographies and other models.}
    \label{fig:summary_align}
\end{figure*}

Results on GovReport and Biographies are in Table~\ref{tab:main}. Our key findings are as follows:
\paragraph{Incorporating length feedback effectively improves length control while maintaining text quality}
Comparisons of the Baseline and Feedback demonstrate consistent and significant improvements in both MAE and PM across all datasets, linguistic units, and LLMs.
To further illustrate this, Figure~\ref{fig:summary_align} presents violin plots comparing the two methods with Qwen3-8B on the GovReport dataset at various target lengths.
The results show that Feedback yields significantly tighter length distributions than Baseline across all granularities, with a higher density around the target lengths.
This confirms the effectiveness of our method in enhancing length control.
Additionally, most Qual. scores remain competitive across both datasets for different LLMs, suggesting that text quality is effectively preserved in our method.
For Markergen, we also observe a significant improvement in length control. However, our method achieves relatively better results without expensive multi-stage inference. In terms of efficiency, our method is approximately 5$\times$ more efficient than Markergen in token consumption. Furthermore, Markergen exhibits noticeably poorer quality scores than the Baseline. A closer examination suggests that the inserted feedback within the body of a sentence can disrupt the content fluency, and LLMs may not adequately follow the prompt at each stage, leading to an error propagation issue.

\paragraph{Generating with length feedback benefits more from ICL than vanilla prompting}
We observe that ICL has an unstable impact on the Baseline. Generally, it degrades the performance on GovReport, while bringing some benefits to Biographies. In contrast, for ICL+Feedback, we observe a more consistent improvement. This is because our method enables better leveraging of the input demonstration to learn how to use feedback information for controlling the generation length. However, we still observe some obvious decline when using Qwen3-8B, which may be due to its capability of leveraging additional contexts.
Nevertheless, these findings suggest that our method can be further improved even with simple ICL, without requiring specific fine-tuning.

\paragraph{Stronger models gain more from incorporating length feedback}
Generally, larger models possess stronger language understanding capabilities, enabling them to more effectively leverage length feedback to adjust subsequent outputs.
When comparing Qwen-4B and Qwen-8B, we observe that larger models consistently achieve greater improvements in MAE and PM scores while typically experiencing only minor negative effects on text quality when incorporating length feedback.
This suggests that our method can achieve better performance when integrated into LLM products, which generally have stronger baseline capabilities.

\subsection{Results on Training-Based Length Control}
\label{sec:train_based}

Though incorporating length feedback can effectively improve length control, we notice that it sometimes blindly reaches the target length, ignoring the response quality for general-domain instances from ELI5. To address this issue, we further investigate SFT for LLMs to balance the length and quality goals.

\input{table/eli5}

\begin{figure*}[t!]
\centering
    \begin{subfigure}{.340\textwidth}
        \includegraphics[width=\linewidth]{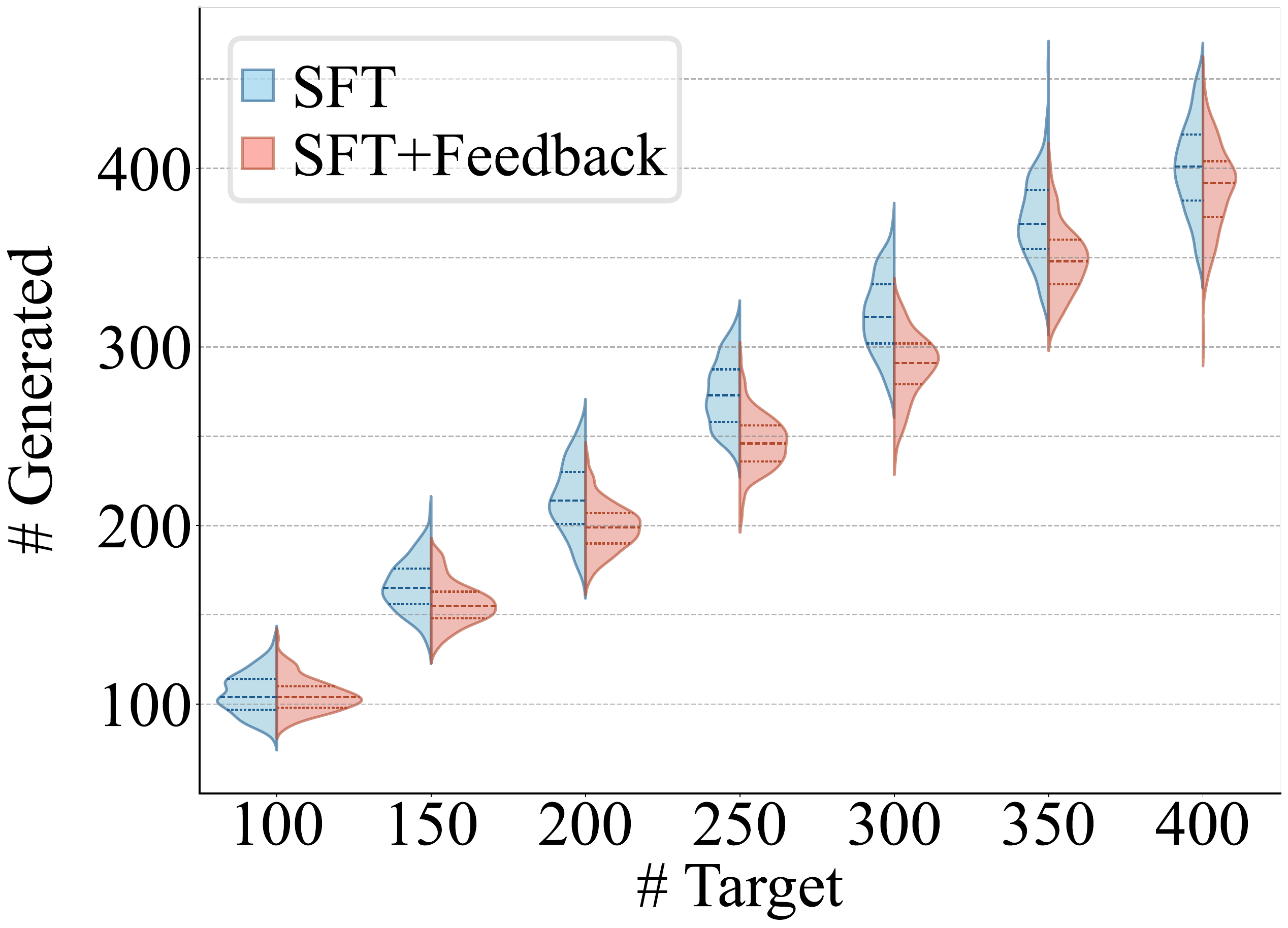} 
        \caption{Token}
        \label{subfig:eli5_a}
    \end{subfigure}
    \hfill
    \begin{subfigure}{.32\textwidth}
        \includegraphics[width=\linewidth]{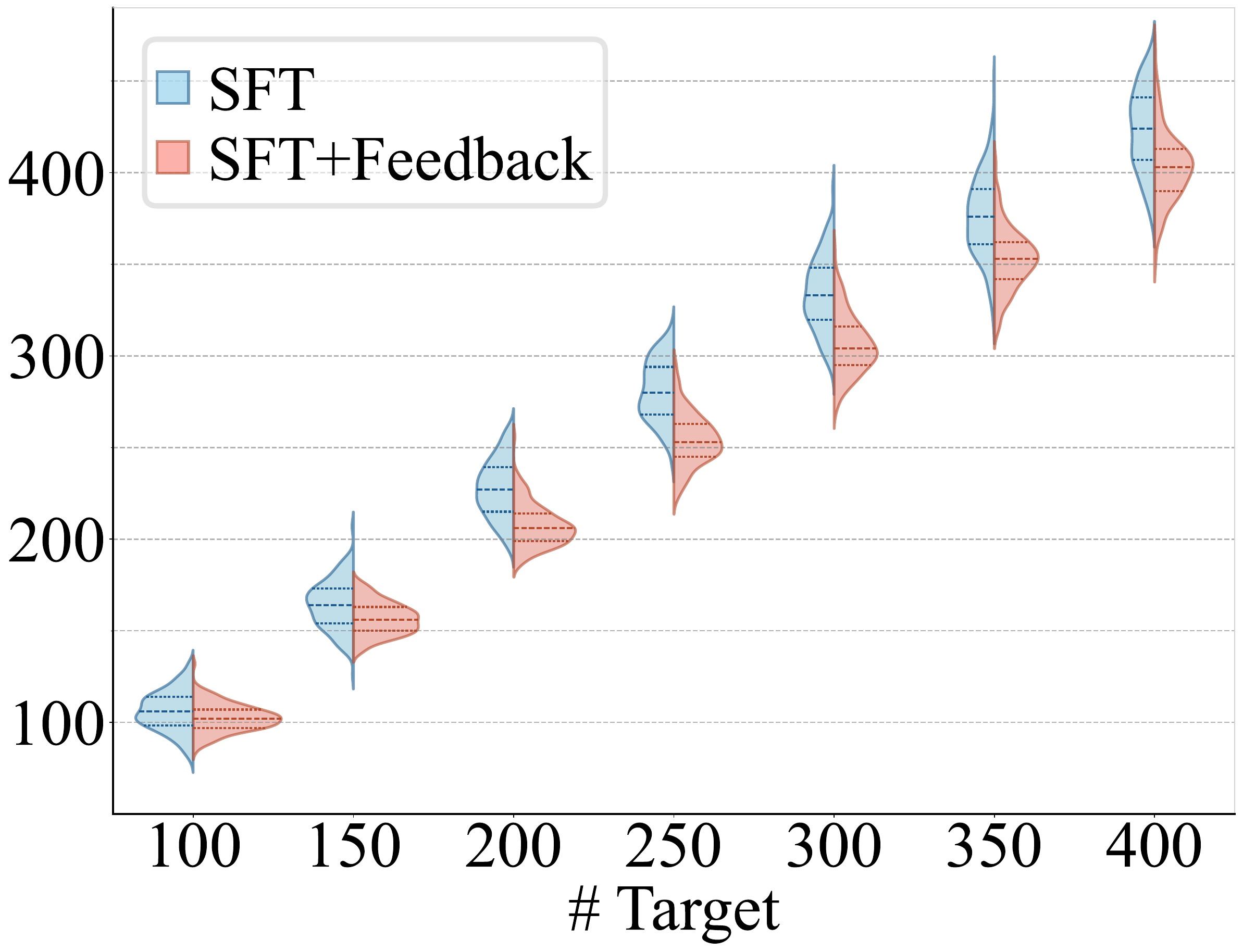} 
        \caption{Word}
        \label{subfig:eli5_b}
    \end{subfigure}
    \hfill
    \begin{subfigure}{.32\textwidth}
        \includegraphics[width=\linewidth]{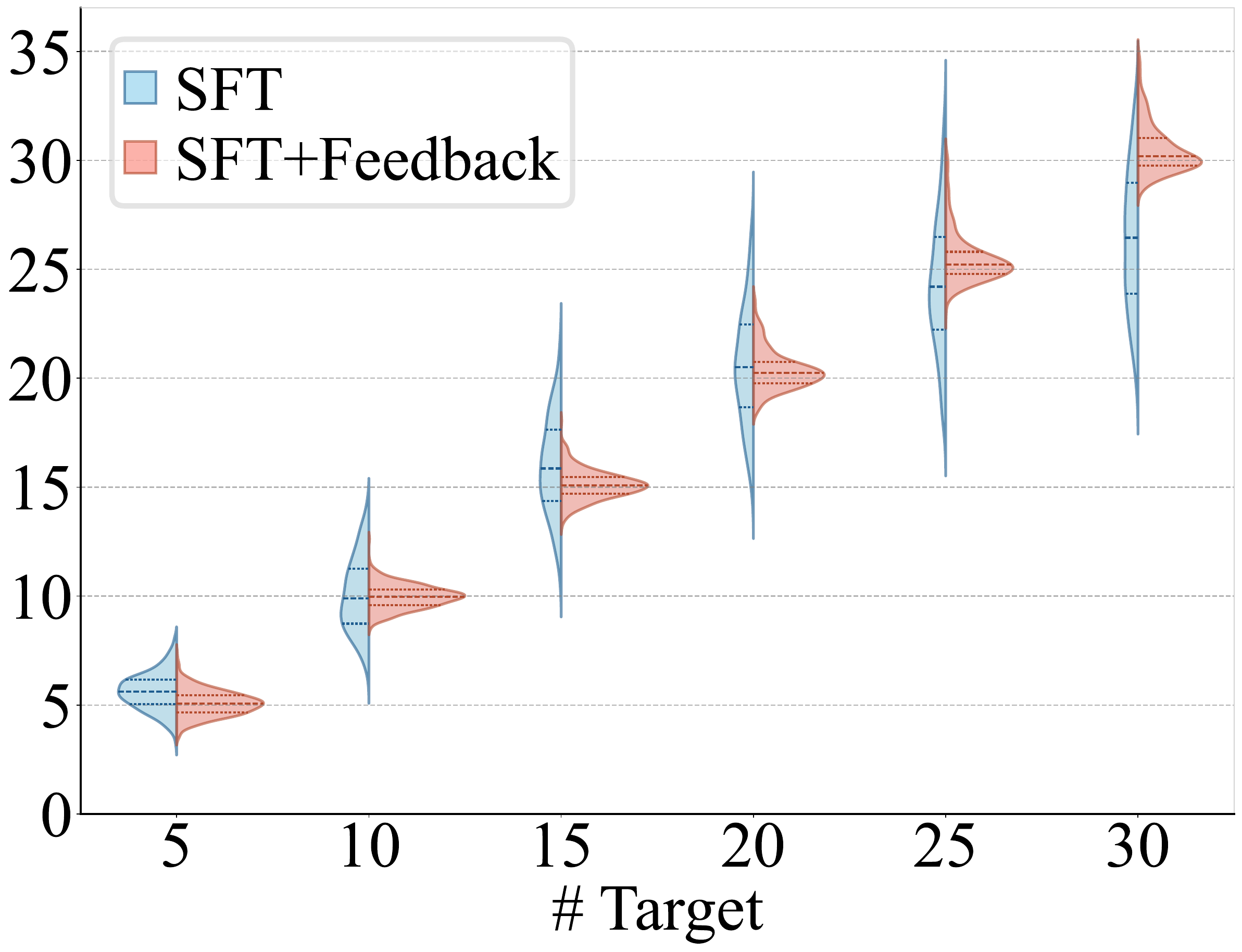} 
        \caption{Sentence}
        \label{subfig:eli5_c}
    \end{subfigure}
    \caption{Generated length distributions of SFT and SFT+Feedback over varying target lengths on the ELI5 test set.}
    \label{fig:eli5}
\end{figure*}

\begin{algorithm}[t]
\small
\caption{Target Length Update}
\label{alg:sft_dataset}
\begin{algorithmic}[1]
    \REQUIRE Length calculation function \textsc{Count}(·)
    \renewcommand{\algorithmicrequire}{\textbf{Input:}}
    \renewcommand{\algorithmicensure}{\textbf{Output:}}
    \REQUIRE An LLM-generated response $\mathbf{y}$ for one question
    \ENSURE Type $\mathcal{T}$, Target length $L_{\text{target}}$
    \STATE $\mathcal{T} \leftarrow \textsc{RandomChoice}(\{\text{token}, \text{word}, \text{sentence}\})$
    \STATE $L_{\text{generated}} \leftarrow \textsc{Count}(\mathbf{y}, \mathcal{T})$ 
    \IF{$\mathcal{T} = \texttt{token}$}
        \STATE $\Delta \leftarrow \textsc{RandomInt}(-10, 10)$
    \ELSIF{$\mathcal{T} = \texttt{word}$}
        \STATE $\Delta \leftarrow \textsc{RandomInt}(-5, 5)$
    \ELSE
        \STATE $\Delta \leftarrow 0$
    \ENDIF
    \STATE $L_{\text{target}} \leftarrow L_{\text{generated}} + \Delta$
    \RETURN $\mathcal{T}, L_{\text{target}}$
\end{algorithmic}
\end{algorithm}

For each instance in our prepared ELI5 training set, we assign a randomly generated target length to create a length requirement. We then provide both the user question and this requirement to the LLM for response generation. Due to the original LLM’s limited ability to follow length constraints, the sampled responses often deviate from the specified target length. To address this, we adjust the target length in the prompt based on the response length, as specified in Algorithm~\ref{alg:sft_dataset}.
For simplicity, we do not filter any response based on its quality due to the costly evaluation. Nevertheless, we show that this can effectively preserve the performance.
In the prompt-based SFT, we use the original model responses without modification. For the feedback-based setting, we first segment responses into sentences, then append length feedback signals after each sentence.
In addition, we implement RULER~\cite{li2024ruler}, a method that employs a meta-token for word-level length control. We utilize the official code\footnote{\url{https://github.com/Geaming2002/Ruler}} for this implementation. To ensure a fair comparison, we train RULER using the same dataset employed for prompt-based SFT. Training details are reported in Appendix~\ref{sec:train_sft}.

Results on ELI5 are presented in Table~\ref{tab:main_1} and Figure~\ref{fig:eli5}. The following conclusions can be drawn:

\paragraph{Incorporating length feedback without fine-tuning sometimes degrades performance on the general domain}
Mirroring the results on specific tasks, incorporating length feedback without fine-tuning continues to yield consistent improvements in open-domain question answering. However, we observe a decline in text quality for certain ELI5 settings, particularly with Qwen3-8B. This suggests the added feedback can sometimes mislead standard generation.
We find, however, that ICL can partly alleviate this problem for Qwen3-8B, underscoring the need for further optimization to preserve output quality.

\input{table/data_efficiency}
\paragraph{SFT effectively balances length control and generation quality}
Applying SFT generally improves length control performance for both the Baseline and Feedback at the token and word levels, echoing the effectiveness of prior approaches.
RULER achieves better length control performance than SFT, however, at the cost of response quality, possibly due to the introduced the meta-token, influncing the orignal generation.
For SFT+Feedback, results demonstrate notable improvement in Qual. scores, performing on par with the Baseline. However, we observe that its sentence-level length control is poorer than that of Feedback. This may be because it is harder to balance the objectives of sentence number and generation quality on this dataset.
Nevertheless, our method still substantially outperforms the Baseline.

\paragraph{Length feedback boosts post-training efficiency}
Intuitively, length feedback frees LLMs from the burden of tracking output length, allowing them to achieve greater training efficacy in length control. To demonstrate this, we analyze the training dynamics of Qwen3-8B at token-level granularity in Figure~\ref{fig:training_efficacy_token}. Results show that SFT+Feedback is at least 4$\times$ more efficient than SFT, achieving competitive performance in just 32 steps. Moreover, we observe that SFT+Feedback has not yet fully converged, suggesting that further scaling the training set could lead to even better performance.
\begin{figure}[t!]
\centering
\vspace{2pt}
\includegraphics[width=0.9\columnwidth]{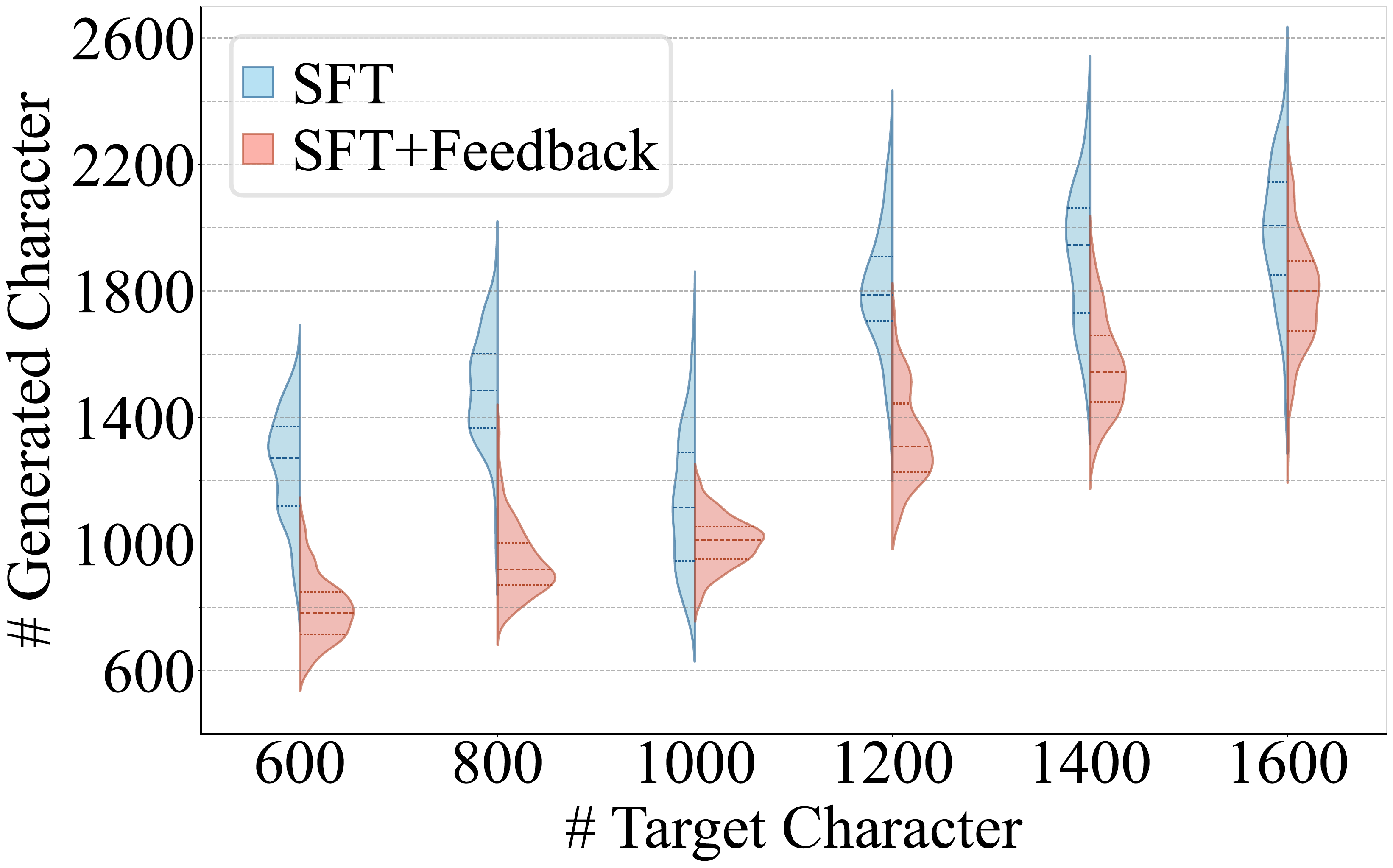} 
\caption{Generated length distributions across target character lengths on the ELI5 test set using Qwen3-8B.}
\label{fig:eli5_align_character}
\end{figure}
\paragraph{Feedback-based SFT exhibits stronger generalization across length units and tasks}
We apply trained models on character-level granularity. Results in Figure~\ref{fig:eli5_align_character} demonstrate that SFT+Feedback consistently outperforms the baseline, with a higher density around the target length in each violin plot. Furthermore, we assess cross-task generalization on TruthfulQA~\cite{lin-etal-2022-truthfulqa}. As shown in Appendix~\ref{appendix:truthfulqa}, our model achieves better cross-task generalization. We believe enriching the training set with more diverse length constraints and tasks could further improve performance. We leave this exploration for future work.

%% file: table/main.tex
\begin{table*}[t]
\small
\centering
\begin{tabular}{lccccccccc}
\toprule
& \multicolumn{3}{c}{Token} & \multicolumn{3}{c}{Word} & \multicolumn{3}{c}{Sentence} \\ 
\cmidrule(lr){2-4} \cmidrule(lr){5-7} \cmidrule(lr){8-10}
& MAE $\downarrow$ & PM $\uparrow$ & Qual. $\uparrow$ & MAE $\downarrow$ & PM $\uparrow$ & Qual. $\uparrow$ & MAE $\downarrow$ & PM $\uparrow$ & Qual. $\uparrow$ \\
\midrule
\multicolumn{10}{c}{\textbf{GovReport}} \\
\midrule
Qwen3-4B & 58.19 & 11.94 & \textbf{4.94} & 70.74 & 8.31 & \textbf{4.95} & 15.85 & 16.36 & \textbf{4.74} \\
\textcolor{gray}{$\hookrightarrow$}~~ICL & 59.07 & 13.30 & 4.77 & 104.20 & 3.88 & 4.80 & 8.20 & 18.41 & 4.53 \\
\textcolor{gray}{$\hookrightarrow$}~~Feedback & 61.30 & 21.35 & 4.64 & 31.96 & 38.53 & 4.84 & 1.38 & 47.83 & 4.62 \\
\textcolor{gray}{$\hookrightarrow$}~~ICL$+$Feedback & \textbf{26.33} & \textbf{34.46} & 4.71 & \textbf{26.63} & \textbf{39.91} & 4.83 & \textbf{0.51} & \textbf{58.40} & 4.67 \\
\cmidrule(lr){1-10}
Qwen3-8B & 52.04 & 15.20 & \textbf{4.96} & 60.61 & 6.06 & \textbf{4.95} & 3.83 & 35.59 & \textbf{4.91} \\
\textcolor{gray}{$\hookrightarrow$}~~ICL & 59.51 & 11.16 & 4.76 & 141.20 & 0.16 & 4.92 & 4.09 & 27.98 & 4.89 \\
\textcolor{gray}{$\hookrightarrow$}~~Feedback & 22.81 & \textbf{58.24} & 4.84 & \textbf{21.23} & \textbf{47.50} & 4.93 & 0.57 & 79.47 & 4.82 \\
\textcolor{gray}{$\hookrightarrow$}~~ICL$+$Feedback & \textbf{17.40} & 50.55 & 4.76 & 51.17 & 17.29 & 4.89 & \textbf{0.10} & \textbf{94.81} & 4.88 \\
\hdashline
\addlinespace[0.8pt]
\textcolor{gray}{$\hookrightarrow$}~~Markergen & 44.26 & 37.27 & 4.29 & 42.19 & 20.91 & 4.62 & 1.87 & 38.61 & 4.74 \\
\cmidrule(lr){1-10}
LLaMA-3.1-8B-Instruct & 108.65 & 5.93 & \textbf{4.78} & 32.83 & 14.32 & \textbf{4.81} & 2.94 & 46.64 & 4.58 \\
\textcolor{gray}{$\hookrightarrow$}~~ICL & 118.27 & 3.31 & 4.70 & 50.60 & 8.66 & 4.79 & 3.32 & 37.77 & 4.54 \\
\textcolor{gray}{$\hookrightarrow$}~~Feedback & 33.88 & 41.30 & 4.64 & \textbf{16.22} & 43.17 & 4.72 & 0.25 & 80.10 & \textbf{4.63} \\
\textcolor{gray}{$\hookrightarrow$}~~ICL$+$Feedback & \textbf{25.94} & 38.80 & 4.53 & 22.93 & 35.69 & 4.72 & \textbf{0.22} & \textbf{82.06} & 4.54 \\
\hdashline
\addlinespace[0.8pt]
\textcolor{gray}{$\hookrightarrow$}~~Markergen & 41.45 & \textbf{56.67} & 3.11 & 34.20 & \textbf{54.50} & 3.54 & 3.36 & 31.22 & 3.66 \\
\midrule
\multicolumn{10}{c}{\textbf{Biographies}} \\
\midrule
Qwen3-4B & 80.46 & 8.24 & 2.94 & 27.66 & 29.82 & \textbf{3.04} & 10.94 & 20.58 & 2.80 \\
\textcolor{gray}{$\hookrightarrow$}~~ICL & 25.07 & 29.01 & 2.94 & 29.41 & 30.45 & 2.98 & 3.36 & 21.68 & 2.62 \\
\textcolor{gray}{$\hookrightarrow$}~~Feedback & 93.36 & 35.68 & 2.82 & 19.37 & 51.99 & 2.99 & 0.28 & 81.66 & \textbf{2.93} \\
\textcolor{gray}{$\hookrightarrow$}~~ICL$+$Feedback & \textbf{14.49} & \textbf{59.41} & \textbf{2.98} & \textbf{7.19} & \textbf{79.31} & 3.02 & \textbf{0.17} & \textbf{82.60} & 2.78 \\
\cmidrule(lr){1-10}
Qwen3-8B & 78.85 & 8.74 & \textbf{3.14} & 55.11 & 8.59 & 3.12 & 2.67 & 26.44 & \textbf{3.11} \\
\textcolor{gray}{$\hookrightarrow$}~~ICL & 27.70 & 28.98 & 3.11 & 38.86 & 16.25 & 3.09 & 1.25 & 50.82 & 3.07 \\
\textcolor{gray}{$\hookrightarrow$}~~Feedback & 22.78 & 66.01 & 3.12 & \textbf{9.22} & \textbf{73.66} & \textbf{3.14} & 0.05 & 97.91 & 3.09 \\
\textcolor{gray}{$\hookrightarrow$}~~ICL$+$Feedback & \textbf{11.22} & \textbf{70.49} & 3.13 & 12.75 & 53.16 & 3.12 & \textbf{0.01} & \textbf{98.54} & 3.06 \\
\hdashline
\addlinespace[0.8pt]
\textcolor{gray}{$\hookrightarrow$}~~Markergen & 42.91 & 44.40 & 2.49 & 26.79 & 40.20 & 2.94 & 2.08 & 33.26 & 3.02 \\
\cmidrule(lr){1-10}
LLaMA-3.1-8B-Instruct & 97.64 & 6.98 & 3.12 & 51.72 & 10.23 & 3.22 & 4.86 & 24.46 & 3.02 \\
\textcolor{gray}{$\hookrightarrow$}~~ICL & 41.81 & 21.16 & 3.24 & 28.69 & 20.04 & 3.33 & 6.12 & 17.12 & 3.12 \\
\textcolor{gray}{$\hookrightarrow$}~~Feedback & 36.90 & 39.04 & 3.34 & 21.96 & 45.15 & 3.32 & 0.77 & 80.20 & \textbf{3.24} \\
\textcolor{gray}{$\hookrightarrow$}~~ICL$+$Feedback & \textbf{24.11} & 47.54 & \textbf{3.38} & \textbf{11.61} & \textbf{61.92} & \textbf{3.36} & \textbf{0.19} & \textbf{92.53} & 3.22 \\
\hdashline
\addlinespace[0.8pt]
\textcolor{gray}{$\hookrightarrow$}~~Markergen & 69.26 & \textbf{53.83} & 2.91 & 37.16 & 53.94 & 2.97 & 2.68 & 29.55 & 2.90 \\
\bottomrule
\end{tabular}
\caption{Results averaged over target length ranges tailored to each linguistic unit, comparing five training-free methods on the summarization and biography generation tasks using three different LLMs.}
\vspace{-1em}
\label{tab:main}
\end{table*}

%% file: table/eli5.tex
\begin{table*}[ht!]
\small
\centering
\begin{tabular}{lccccccccc}
\toprule
& \multicolumn{3}{c}{Token} & \multicolumn{3}{c}{Word} & \multicolumn{3}{c}{Sentence} \\ 
\cmidrule(lr){2-4} \cmidrule(lr){5-7} \cmidrule(lr){8-10}
& MAE $\downarrow$ & PM $\uparrow$ & Qual. $\uparrow$ & MAE $\downarrow$ & PM $\uparrow$ & Qual. $\uparrow$ & MAE $\downarrow$ & PM $\uparrow$ & Qual. $\uparrow$ \\
\midrule
Qwen3-8B & 59.84 & 9.35 & 7.83 & 46.09 & 14.71 & 8.06 & 1.82 & 48.67 & 7.57 \\
\textcolor{gray}{$\hookrightarrow$}~~ICL & 30.64 & 19.65 & 7.80 & 33.27 & 20.86 & 7.89 & 2.74 & 34.57 & 7.71 \\
\textcolor{gray}{$\hookrightarrow$}~~Feedback & 60.99 & 37.09 & 6.78 & 20.86 & \textbf{56.29} & 7.50 & 0.12 & 91.15 & 7.51 \\
\textcolor{gray}{$\hookrightarrow$}~~ICL$+$Feedback  & \textbf{10.33} & \textbf{68.65} & 7.83 & 20.65 & 39.13 & 7.85 & \textbf{0.08} & \textbf{92.80} & 7.69 \\
\textcolor{gray}{$\hookrightarrow$}~~SFT & 23.72 & 30.59 & \textbf{8.14} & 27.68 & 24.86 & \textbf{8.08} & 2.21 & 19.18 & \textbf{8.18} \\
\textcolor{gray}{$\hookrightarrow$}~~SFT$+$Feedback & 16.41 & 47.90 & 8.01 & \textbf{14.10} & 53.94 & 8.07 & 0.40 & 78.60 & 8.11 \\
\hdashline
\addlinespace[0.8pt]
\textcolor{gray}{$\hookrightarrow$}~~RULER & -- & -- & -- & 20.80 & 34.55 & 6.70 & -- & -- & -- 
\\
\cmidrule(lr){1-10}
LLaMA-3.1-8B-Instruct & 70.03 & 9.05 & 7.44 & 39.58 & 9.80 & 7.45 & 3.52 & 33.45 & 7.26 \\
\textcolor{gray}{$\hookrightarrow$}~~ICL & 53.83 & 9.45 & 7.43 & 35.37 & 15.03 & 7.52 & 4.05 & 27.84 & 7.31 \\
\textcolor{gray}{$\hookrightarrow$}~~Feedback & 24.48 & 37.94 & 7.38 & 15.55 & 44.47 & 7.33 & \textbf{0.11} & \textbf{95.75} & 7.05 \\
\textcolor{gray}{$\hookrightarrow$}~~ICL$+$Feedback & 25.28 & 39.18 & 7.25 & \textbf{14.15} & 50.31 & 7.26 & 0.30 & 87.13 & 7.18 \\
\textcolor{gray}{$\hookrightarrow$}~~SFT & 31.91 & 18.60 & \textbf{7.49} & 29.24 & 19.06 & \textbf{7.46} & 3.43 & 11.12 & \textbf{7.50} \\
\textcolor{gray}{$\hookrightarrow$}~~SFT$+$Feedback & \textbf{18.39} & \textbf{46.41} & 7.28 & 15.40 & \textbf{52.53} & 7.40 & 1.54 & 44.40 & 7.45\\
\hdashline
\addlinespace[0.8pt]
\textcolor{gray}{$\hookrightarrow$}~~RULER & -- & -- & -- & 27.31 & 29.46 & 6.10 & -- & -- & -- \\
\bottomrule
\end{tabular}
\caption{Results averaged over target length ranges tailored to each linguistic unit on the ELI5 test set. Note that RULER's meta-token design is exclusive to word-level evaluation.}
\label{tab:main_1}
\end{table*}

%% file: table/data_efficiency.tex
\begin{figure}[t!]
\hspace*{-0.23cm}
\centering
\scalebox{0.88}{\begin{tikzpicture}
\begin{axis}[
    xlabel={\# Training Steps},
    ylabel={
        \begin{tabular}{c}
        \tikz{\draw[fill=purple!45!white,draw=purple!45!white] (0,0) rectangle (0.6em,0.6em);}~~MAE~$\downarrow$
        \end{tabular}
    },
    xmin=6, xmax=170,
    ymin=5, ymax=59,
    xtick={16,32,...,160},
    ytick={10,20,...,60},
    width=8cm,
    height=5.25cm,
    legend style={at={(0.5, 1.18)}, anchor=north, legend columns=2},
    y label style={yshift=-0.5em},
    axis y line*=left,
    axis x line*=bottom,
    enlargelimits=false,
    grid=both,
    ymajorgrids=false,
    every axis plot/.append style={very thick},
]
\addlegendimage{black, mark=square}
\addlegendentry{SFT}
\addlegendimage{black, mark=square*}
\addlegendentry{SFT+Feedback}

\addplot[
color=purple!45!white,
mark=square,
]
coordinates {
    (16,32.9)(32,28.3)(48,26.4)(64,25.4)(80,23.3)
    (96,23.9)(112,23.8)(128,22.5)(144,24.6)(160,23.7)
};

\addplot[
    color=purple!45!white,
    mark=square*,
]
coordinates {
    (16,55.8)(32,22.3)(48,23.2)(64,20.4)(80,18.5)
    (96,18.0)(112,17.8)(128,17.1)(144,16.2)(160,16.4)
};

\end{axis}

\begin{axis}[
    xlabel={Training Steps},
    ylabel={
        \begin{tabular}{c}
            \tikz{\draw[fill=teal!50!white,draw=teal!50!white] (0,0) rectangle (0.6em,0.6em);}~~PM~$\uparrow$
        \end{tabular}
    },
    xmin=6, xmax=170,
    ymin=5, ymax=59,
    xtick={8,16,32,...,160,168},
    ytick={10,20,...,50},
    axis y line*=right,
    axis x line=none,
    y label style={yshift=0.5em},
    width=8cm,
    height=5.25cm,
    enlargelimits=false,
    grid=none,
    every axis plot/.append style={very thick},
]

\addplot[
    color=teal!50!white,
    mark=square,
]
coordinates {
    (16,23.2)(32,27.3)(48,28.1)(64,28.4)(80,33.8)
    (96,32.5)(112,31.2)(128,32.9)(144,31.2)(160,30.6)
};

\addplot[
    color=teal!50!white,
    mark=square*,
]
coordinates {
    (16,13.8)(32,40.6)(48,40.2)(64,43.3)(80,44.1)
    (96,45.7)(112,45.8)(128,48.1)(144,49.9)(160,47.9)
};

\end{axis}
\end{tikzpicture}}
\vspace{-16pt}
\caption{Training dynamics of token-length control performance on the ELI5 task using Qwen3-8B.}
\label{fig:training_efficacy_token}
\vspace{-0.5em}
\end{figure}

%% file: section/5_related.tex
\section{Related Work}
\subsection{Length Control for Text Generation}
Controlling output length is crucial to meet diverse user requirements.
Early studies addressed this challenge by designing early-stopping decoding strategies~\cite{rush2015neural} or incorporating specialized length-aware module~\cite{liu2018controlling,takase2019positional,yu2021lenatten} to guide generation.
With the emergence of LLMs~\cite{achiam2023gpt}, their advanced instruction-following capability enables length control through direct prompting~\cite{markergen}.
However, existing methods still rely on extensive post-training, such as SFT~\cite{yuan2024following,li2024ruler,butcher_precise_2025} or RL~\cite{singhal2023long,jie_prompt-based_2024}, to achieve precise length control.
In contrast, our method accomplishes this goal by incorporating length feedback, which is either training-free or training-efficient. This approach enables strong generalization under diverse length constraints.

\subsection{Feedback-Guided Text Generation}
Since LLMs are essentially probabilistic models, they often struggle with tasks involving complex mathematical reasoning~\cite{zhou2022least} or up-to-date factual knowledge~\cite{luo2023systematic}. A straightforward way to overcome these limitations is to equip LLMs with the ability to use external tools~\cite{parisi2022talm,schick2023toolformer}, such as search engines, calculators, or compilers. Using external feedback from these tools, LLMs' capabilities can be substantially extended, allowing them to focus better on tasks they inherently excel at.
Markergen~\cite{markergen} was the first to integrate length feedback directly into the generation process. However, its multi-stage framework introduces considerable complexity and inefficiency.
In contrast, empirical results demonstrate that our method achieves better length control while maintaining text quality in a single pass. This streamlined process naturally allows further optimization through training.

%% file: section/6_conclu.tex
\section{Conclusion}
In this work, we propose incorporating external length feedback at the end of sentences during LLM generation.
To enable models to internalize this mechanism, we further construct a dataset enriched with length feedback and perform SFT, promoting generalization across diverse tasks and length specifications.
We evaluate our approach on both domain-specific and general-domain tasks across different models.
Results show our method achieves a more favorable balance among inference efficiency, precise length control, and high text quality. Furthermore, unlike previous work, our method efficiently adapts to various length types.
In the future, we plan to extend our method to constraints beyond output length, further enhancing LLMs' controllability over their outputs.

%% file: section/7_appendix.tex
\appendix
\section{Additional Results from the Preliminary Study}
\label{appendix:preliminary}
We extend the investigation from token-level to word and sentence granularities. Following the token-level setup, we collect responses of varying lengths by prompting LLMs under the constraint of different target lengths.
Specifically, the target lengths at the word level are randomly sampled within the range of 100 to 400 words, whereas sentence‐level target lengths are randomly sampled between 5 and 30 sentences.

Detailed results for the word-level evaluation are shown in Figure~\ref{fig:pilot_study_word}, and the sentence-level results appear in Figure~\ref{fig:pilot_study_sentence}.
Consistently across all granularities, longer texts exhibit higher internal estimation errors, indicating a degraded self-assessment capability that impairs precise length control.
\input{table/length_estimation_word}
\input{table/length_estimation_sentence}
\section{Analysis of Model Judgment Reliability}
\label{appendix:human_evaluation}
To assess the reliability of model-based quality judgments, we conduct a human evaluation study utilizing 50 summarization samples.

We recruited five graduate students who verified their fluency in English and familiarity with fundamental text evaluation concepts. Each annotator was compensated at a flat rate of \$15 per hour. To ensure uniform understanding of the comparative evaluation criteria, all annotators received detailed written instructions and participated in a brief training session. The full text of the unified instructions is provided in Figure~\ref{fig:human_unified_guideline}. For each sample, human annotators are presented with two summaries: one generated by the feedback-based method and one generated by the baseline method. The annotators are blinded to the source of each summary. Their task is to compare the two summaries along three dimensions: Coherence, Consistency, and Relevance.

We compare these human comparative judgments with the ratings provided by two LLM evaluators, GPT-4.1 and DeepSeek-V3, respectively. Both models assign individual scores to each of the two summaries separately, a practice that generates comparative signals which enable alignment with the comparative judgments derived from human evaluators.

As shown in Table~\ref{tab:human_eval}, GPT-4.1 achieves a slightly higher consistency ratio for Coherence whereas DeepSeek-V3 outperforms GPT-4.1 in the other two critical dimensions as it attains an 86\% consistency ratio for Consistency and a 78\% ratio for Relevance. This finding indicates that DeepSeek-V3 possesses greater credibility in capturing human-perceived text quality, leading to its adoption for quality evaluation across all our experiments.


\begin{figure*}[ht!]
\centering
\begin{tcolorbox}[
colback=gray!5!white,
colframe=gray!50!black,
coltitle=white,
fonttitle=\bfseries\normalsize,
fontupper=\small,
title=Human Annotation Guideline
]
\textbf{Task:}
\\ \\
You will be given one source government report and two summaries (Summary A and Summary B) written for that report. Your task is to read the source report and both summaries carefully, then compare the two summaries and make judgments on three dimensions: Coherence, Consistency, and Relevance. For each dimension, decide which summary is better, or if they are of equal quality. Evaluate independently based on the criteria below.
\\ \\
\textbf{Evaluation Dimensions \& Criteria:}
\\ \\
\textbf{Coherence} – This dimension assesses the overall flow, structure, and logical organization of the summary. A coherent summary should read as a unified whole, not a list of disjointed statements.
\begin{itemize}
\item Summary A is better if it is significantly more coherent (well-structured, smooth flow).
\item Summary B is better if it is significantly more coherent.
\item Equal if both summaries are similar in coherence.
\end{itemize}
\textbf{Consistency} – This dimension assesses the factual alignment between the summary and the source document. A consistent summary should contain only information that is directly stated or logically entailed by the source, with no hallucinations or contradictions.
\begin{itemize}
\item Summary A is better if it has significantly fewer factual errors or hallucinations.
\item Summary B is better if it has significantly fewer factual errors or hallucinations.
\item Equal if both summaries have similar factual consistency.
\end{itemize}
\textbf{Relevance} – This dimension assesses the selection of important and salient content from the source document. A relevant summary should focus on the core topic and key information, avoiding redundancy and irrelevant details.
\begin{itemize}
\item Summary A is better if it more effectively covers important content with less redundancy.
\item Summary B is better if it more effectively covers important content with less redundancy.
\item Equal if both summaries are similar in relevance and coverage.
\end{itemize}
\textbf{Evaluation Steps:}
\begin{enumerate}
\item \textbf{Read Source:} Read the source government report and identify its main topic, key points, and important facts.
\item \textbf{Read Both Summaries:} Read both Summary A and Summary B carefully.
\item \textbf{Compare \& Judge:} For each dimension above, compare the two summaries and determine which summary performs better on this dimension, selecting from A, B, or Equal.
\end{enumerate}
For each sample, record your comparative judgment for Coherence, Consistency, and Relevance (e.g., Sample 01: Coherence: A, Consistency: Equal, Relevance: B).
\end{tcolorbox}
\vspace{-4pt}
\caption{Human annotation guideline for summary quality evaluation.}
\label{fig:human_unified_guideline}
\end{figure*}

\begin{table}[ht!]
\centering
\small
\begin{tabular}{lccc}
\toprule
 & Coherence & Consistency & Relevance \\
\midrule
DeepSeek-V3 & 72\% & 86\% & 78\% \\
\addlinespace[2.8pt]
GPT-4.1 & 74\% & 80\% & 68\% \\
\bottomrule
\end{tabular}
\caption{Consistency ratios between human and model judgments.}
\label{tab:human_eval}
\end{table}

\section{Additional Analyses on Training-Free Length Control}
\label{appendix:train-free}
We present detailed length distributions on GovReport for Qwen3-4B (Figure~\ref{fig:summary_qwen3_4B_align}), Qwen3-8B (Figure~\ref{fig:summary_qwen3_8B_align}) and LLaMA-3.1-8B-Instruct (Figure~\ref{fig:summary_llama_align}).
Each figure compares four training-free strategies at the token, word, and sentence granularities across a corresponding range of target lengths.
Taking Qwen3-4B  as an example, Figure~\ref{subfig:summary_qwen3_4B_prompt_a} to~\ref{subfig:summary_qwen3_4B_prompt_c} contrast Baseline and Feedback, while Figure~\ref{subfig:summary_qwen3_4B_icl_a} to~\ref{subfig:summary_qwen3_4B_icl_c} compare ICL and ICL+Feedback.
Corresponding evaluations on Biographies appear in Figures~\ref{fig:bio_qwen3_4B_align},~\ref{fig:bio_qwen3_8B_align} and~\ref{fig:bio_llama_align}.
Across various experimental setups, integrating length feedback (i.e., Feedback and ICL+Feedback) consistently yields output length distributions that more closely adhere to target ranges than their counterparts (i.e., Baseline and ICL), especially for larger models.
This robust performance underscores the effectiveness and generalizability of the feedback mechanism in achieving precise output length regulation without requiring training.
\input{appendix_fig/summary_qwen3_4B}
\input{appendix_fig/summary_qwen3_8B}
\input{appendix_fig/summary_llama}
\input{appendix_fig/bio_qwen3_4B}
\input{appendix_fig/bio_qwen3_8B}
\input{appendix_fig/bio_llama}
\section{Additional Analyses on Training-Based Length Control}
\label{appendix:train-based}
Figure~\ref{fig:eli5_llama} illustrates the length distributions produced by LLaMA-3.1-8B-Instruct with SFT+Feedback and SFT across various target lengths.
We observe that LLaMA-3.1-8B-Instruct with SFT+Feedback consistently produces distributions that are more tightly concentrated around the specified target lengths than SFT, indicating superior precision in length control.
\input{appendix_fig/eli5_llama}
\section{Hyperparameters for SFT Training}
\label{sec:train_sft}
For all models, we perform SFT using the \texttt{TRL} library\footnote{https://github.com/huggingface/trl} on four A100 GPUs. We set the learning rate to \(1 \times 10^{-6}\), with a batch size of 64 for training one epoch.
\section{Cross-Task Generalization}
\label{appendix:truthfulqa}
To verify the model’s cross-task generalization ability after training, we conduct evaluations on TruthfulQA, a dataset unseen during training. As shown in Table~\ref{tab:truthfulqa}, our SFT+Feedback model achieves superior length control performance, demonstrating enhanced generalization capability.
\begin{table*}[ht!]
\centering
\small
\begin{tabular}{lccccccccc}
\toprule
& \multicolumn{3}{c}{Token} & \multicolumn{3}{c}{Word} & \multicolumn{3}{c}{Sentence} \\ 
\cmidrule(lr){2-4} \cmidrule(lr){5-7} \cmidrule(lr){8-10}
& MAE $\downarrow$ & PM $\uparrow$ & Qual. $\uparrow$ & MAE $\downarrow$ & PM $\uparrow$ & Qual. $\uparrow$ & MAE $\downarrow$ & PM $\uparrow$ & Qual. $\uparrow$ \\
\midrule
SFT & 25.43 & 28.63 & 3.87 & 29.71 & 22.69 & 3.89 & 2.73 & 16.84 & 3.88 \\
\addlinespace[2.8pt]
SFT+Feedback & 17.83 & 49.32 & 3.86 & 14.52 & 54.96 & 3.82 & 0.52 & 78.10 & 3.76 \\
\bottomrule
\end{tabular}
\caption{Results of the ELI5-trained models on the TruthfulQA task.}
\label{tab:truthfulqa}
\end{table*}
\section{Prompt Templates}
\subsection{Prompt Template for Token Counting}
In the preliminary study, we adopt the prompt in Figure~\ref{fig:token_count} to enable LLMs to estimate their output length.

\begin{figure*}[ht!]
\centering
\begin{tcolorbox}[
    colback=gray!5!white, 
    colframe=gray!50!black, 
    coltitle=white,
    fonttitle=\bfseries\normalsize,  
    fontupper=\small,
    title=User Prompt for Token Counting
]
You will be given a piece of text. Please answer how many tokens are used in the text. Output the number of tokens within \textbackslash boxed\{\}.
\\ \\
Text: \texttt{[Summary]}
\end{tcolorbox}
\caption{Prompt template for token counting.}
\label{fig:token_count}
\end{figure*}

\subsection{Prompt Templates for Quality Evaluation}
Prompt templates for assessing response quality across various generation tasks are depicted in Figures~\ref{fig:eli5_evaluation} through~\ref{fig:biography_factuality_evaluation}. In the summarization task, the coherence, consistency and relevance evaluation templates are respectively shown in Figures~\ref{fig:summary_coherence_evaluation}, \ref{fig:summary_consistency_evaluation} and \ref{fig:summary_relevance_evaluation}.
For the biography generation task, Figure~\ref{fig:biography_coherence_evaluation} displays the coherence evaluation template, and Figure~\ref{fig:biography_factuality_evaluation} presents the template for factuality evaluation.
Additionally, Figure~\ref{fig:eli5_evaluation} shows the quality evaluation template applicable to general-domain text generation.
We present all prompt templates employed for text quality evaluation across various generation tasks.
Specifically, general-domain texts are measured in terms of correctness and helpfulness. Summaries are evaluated for coherence, consistency, and relevance, while biographies are assessed for coherence and factuality.
These prompt templates facilitate a rigorous and uniform approach to measuring key quality dimensions across different generation settings.

\begin{figure*}[ht!]
\centering
\begin{tcolorbox}[
    colback=gray!5!white, 
    colframe=gray!50!black, 
    coltitle=white,
    fonttitle=\bfseries\normalsize,
    fontupper=\small,
    title=Prompt Template for General-Domain Text Quality Evaluation
]
\relax[Instruction]

Please act as an impartial judge and evaluate the quality of the response provided by an AI assistant to the user question displayed below. Your evaluation should consider correctness and helpfulness. You will be given a reference answer and the assistant's answer. Begin your evaluation by comparing the assistant's answer with the reference answer. Identify and correct any mistakes. Be as objective as possible. After providing your explanation, you must rate the response on a scale of 1 to 10 by strictly following this format: ``\textbackslash  boxed\{rating\}", for example: ``Rating: \textbackslash boxed\{5\}".
\\ \\
\relax[Question]

\{question\}
\\ \\
\relax[The Start of Reference Answer]

\{reference\}

\relax[The End of Reference Answer]
\\ \\
\relax[The Start of Assistant's Answer]

\{prediction\}

\relax[The End of Assistant's Answer]
\end{tcolorbox}
\vspace{-4pt}
\caption{Prompt template for general-domain text quality evaluation.}
\label{fig:eli5_evaluation}
\end{figure*}
\begin{figure*}[!ht]
\centering
\begin{tcolorbox}[
    colback=gray!5!white, 
    colframe=gray!50!black, 
    coltitle=white,
    fonttitle=\bfseries\normalsize,
    fontupper=\small,
    title=Prompt Template for Summary Coherence Evaluation
]
You will be given one summary written for a government report.
\\ \\
Your task is to rate the summary on one metric.
\\ \\
Please make sure you read and understand these instructions carefully. Please keep this document open while reviewing, and refer to it as needed.
\\ \\
Evaluation Criteria:
\\ \\
Coherence (1-5) - the collective quality of all sentences. We align this dimension with the DUC quality question of structure and coherence whereby ``the summary should be well-structured and well-organized. The summary should not just be a heap of related information, but should build from sentence to a coherent body of information about a topic."
\\ \\
Evaluation Steps:
\\ \\
1. Read the government report carefully and identify the main topic and key points.

2. Read the summary and compare it to the government report. Check if the summary covers the main topic and key points of the government report, and if it presents them in a clear and logical order.

3. Assign a score for coherence on a scale of 1 to 5, where 1 is the lowest and 5 is the highest based on the Evaluation Criteria.
\\ \\
Example:
\\ \\
Source Text:

\texttt{[Document]}
\\ \\
Summary:

\texttt{[Summary]}
\\ \\
Follow the evaluation steps and then output the coherence score within \textbackslash boxed\{\}.
\end{tcolorbox}
\vspace{-4pt}
\caption{Prompt template for summary coherence evaluation.}
\label{fig:summary_coherence_evaluation}
\end{figure*}

\begin{figure*}[ht!]
\centering
\begin{tcolorbox}[
    colback=gray!5!white, 
    colframe=gray!50!black, 
    coltitle=white,
    fonttitle=\bfseries\normalsize,
    fontupper=\small,
    title=Prompt Template for Summary Consistency Evaluation
]
You will be given a government report. You will then be given one summary written for this report.
\\ \\
Your task is to rate the summary on one metric.
\\ \\
Please make sure you read and understand these instructions carefully. Please keep this document open while reviewing, and refer to it as needed.
\\ \\
Evaluation Criteria:
\\ \\
Consistency (1-5) - the factual alignment between the summary and the summarized source. A factually consistent summary contains only statements that are entailed by the source document. Annotators were also asked to penalize summaries that contained hallucinated facts. 
\\ \\
Evaluation Steps:
\\ \\
1. Read the government report carefully and identify the main facts and details it presents.

2. Read the summary and compare it to the report. Check if the summary contains any factual errors that are not supported by the report.

3. Assign a score for consistency based on the Evaluation Criteria.
\\ \\
Example:
\\ \\
Source Text: 

\texttt{[Document]}
\\ \\
Summary: 

\texttt{[Summary]}
\\ \\
Follow the evaluation steps and then output the consistency score within \textbackslash boxed\{\}.
\end{tcolorbox}
\vspace{-4pt}
\caption{Prompt template for summary consistency evaluation.}
\label{fig:summary_consistency_evaluation}
\end{figure*}

\begin{figure*}[!ht]
\centering
\begin{tcolorbox}[
    colback=gray!5!white, 
    colframe=gray!50!black, 
    coltitle=white,
    fonttitle=\bfseries\normalsize,
    fontupper=\small,
    title=Prompt Template for Summary Relevance Evaluation
]
You will be given one summary written for a government report.
\\ \\
Your task is to rate the summary on one metric.
\\ \\
Please make sure you read and understand these instructions carefully. Please keep this document open while reviewing, and refer to it as needed.
\\ \\
Evaluation Criteria:
\\ \\
Relevance (1-5) - selection of important content from the source. The summary should include only important information from the source document. Annotators were instructed to penalize summaries which contained redundancies and excess information.
\\ \\
Evaluation Steps:
\\ \\
1. Read the summary and the source document carefully.

2. Compare the summary to the source document and identify the main points of the report.

3. Assess how well the summary covers the main points of the report, and how much irrelevant or redundant information it contains.
4. Assign a relevance score from 1 to 5.
\\ \\
Example:
\\ \\
Source Text:

\texttt{[Document]}
\\ \\
Summary:

\texttt{[Summary]}
\\ \\
Follow the evaluation steps and then output the relevance score within \textbackslash boxed\{\}.
\end{tcolorbox}
\vspace{-4pt}
\caption{Prompt template for summary relevance evaluation.}
\label{fig:summary_relevance_evaluation}
\end{figure*}
\begin{figure*}[ht!]
\centering
\begin{tcolorbox}[
    colback=gray!5!white, 
    colframe=gray!50!black, 
    coltitle=white,
    fonttitle=\bfseries\normalsize,
    fontupper=\small,
    title=Prompt Template for Biography Coherence Evaluation
]
You will be given a biography of a person.
\\ \\
Your task is to rate the biography on one metric.
\\ \\
Please make sure you read and understand these instructions carefully.
\\ \\
Evaluation Criteria:
\\ \\
Coherence (1-5) - The collective quality of all sentences. The biography should be well-structured and logically organized. The biography should not just be a heap of related information, but should build from sentence to sentence into a coherent body of information about the person.
\\ \\
Evaluation Steps:
\\ \\
1. Read the biography carefully and assess whether it presents them in a clear and logical order.

2. Assign a score for coherence on a scale of 1 to 5, where 1 is the lowest and 5 is the highest based on the Evaluation Criteria.
\\ \\
Example:
\\ \\
Person:

\texttt{[Person]}
\\ \\
Biography:

\texttt{[Biography]}
\\ \\
Follow the evaluation steps and then output the coherence score within \textbackslash boxed\{\}.
\end{tcolorbox}
\vspace{-4pt}
\caption{Prompt template for biography coherence evaluation.}
\label{fig:biography_coherence_evaluation}
\end{figure*}

\begin{figure*}[ht!]
\centering
\begin{tcolorbox}[
    colback=gray!5!white, 
    colframe=gray!50!black, 
    coltitle=white,
    fonttitle=\bfseries\normalsize,
    fontupper=\small,
    title=Prompt Template for Biography Factuality Evaluation
]
You will be given two texts about a person:

1. An authoritative Wikipedia text that contains verifiable factual information about the person.

2. A generated biography written by an AI assistant.
\\ \\
Your task is to rate the generated biography on one metric.
\\ \\
Please make sure you read and understand these instructions carefully.
\\ \\
Evaluation Criteria:
\\ \\
Factual Accuracy (1-5) — The degree to which verifiable claims in the generated biography align with the information provided in the Wikipedia text. Please use the following scale:

  - 5 (Excellent): All key factual claims are fully supported by the Wikipedia text. Minor details may be missing, but there are no factual errors.
  
  - 4 (Good): Most claims are supported. There may be minor factual errors or omissions, but they do not affect the overall understanding.
  
  - 3 (Fair): A moderate number of claims are supported. Some factual errors or omissions are present, but the basic understanding remains.
  
  - 2 (Poor): Few claims are supported, and some factual errors or unverified statements may impact the core understanding.
  
  - 1 (Very Poor): Every claim is either factually false or entirely unsupported, constituting largely fabricated content.
\\ \\
Evaluation Steps:
\\ \\
1. Carefully read the Wikipedia text.

2. Read the generated biography carefully and identify all factual claims in the biography.

3. For each claim, check whether it is supported, contradicted, or absent in the Wikipedia text.

4. Assign a factual accuracy score on a scale of 1 to 5, where 1 is the lowest and 5 is the highest based on the Evaluation Criteria.
\\ \\
Example:
\\ \\
Person:

\texttt{[Person]}
\\ \\
Wikipedia\_Text:

\texttt{[Wikipedia\_Text]}
\\ \\
Generated\_Biography:

\texttt{[Biography]}
\\ \\
Follow the evaluation steps and then output the factual accuracy score within \textbackslash boxed\{\}.
\end{tcolorbox}
\vspace{-4pt}
\caption{Prompt template for biography factuality evaluation.}
\label{fig:biography_factuality_evaluation}
\end{figure*}

%% file: table/length_estimation_word.tex
\newcommand{\sevenptword}{\fontsize{7.6pt}{9pt}\selectfont}
\pgfplotsset{compat=1.3,
    /pgfplots/ybar legend/.style={
    /pgfplots/legend image code/.code={%
       \draw[##1,/tikz/.cd,yshift=-0.25em]
        (0cm,0cm) rectangle (7pt,0.8em);},
   },
}
\begin{figure*}[!ht]
\hspace*{-0.5cm}
\pgfplotsset{width=6.5cm, height=4.5cm}
\centering
\scalebox{0.90}{\begin{tikzpicture}
\begin{groupplot}[
    group style={
        group name=my plots,
        group size=3 by 1,
        horizontal sep=0.8cm,
    },
]

\nextgroupplot[
    ybar,
    ymin=0,
    ymax=135,
    ytick={0,30,...,120},
    ylabel={MAE~$\downarrow$},
    y label style={yshift=-0.6em},
    symbolic x coords={{[50,150)}, {[150,250)}, {[250,350)}, {[350,450)}},
    xticklabel style={font=\sevenptword},
    xtick=data,
    enlarge x limits=0.2,
    xtick pos=bottom,
    ytick pos=left,
    bar width=11pt,
    title style={yshift=-12.25em},
    title={(a) Qwen3-4B},
    axis lines*= left,
]
\addplot[color=purple!80!black, fill=purple!20!white] coordinates {
    ({[50,150)},11.584) ({[150,250)},21.00526315789474) ({[250,350)},61.05761316872428) ({[350,450)},111.38152610441767)
};
\addplot[color=teal!80!black, fill=teal!20!white] coordinates {
    ({[50,150)},10.776) ({[150,250)},26.310526315789474) ({[250,350)},69.92181069958848) ({[350,450)},87.12449799196787)
};

\nextgroupplot[
    ybar,
    ymin=0,
    ymax=135,
    ytick={0,30,...,120},
    symbolic x coords={{[50,150)}, {[150,250)}, {[250,350)}, {[350,450)}},
    xticklabel style={font=\sevenptword},
    xtick=data,
    enlarge x limits=0.2,
    xtick pos=bottom,
    ytick pos=left,
    bar width=11pt,
    title style={yshift=-12.25em},
    title={(b) Qwen3-8B},
    axis lines*= left,
    legend style={
            at={(0.5,1)},
            anchor=south,
            column sep=0.5ex,
            /tikz/every even column/.append style={column sep=1.5em},
            font=\small,
            draw=none,
            legend columns=2,
        },
]
\addplot[color=purple!80!black, fill=purple!20!white]  coordinates {
    ({[50,150)},30.985507246376812) ({[150,250)},27.144186046511628) ({[250,350)},36.96511627906977) ({[350,450)},90.59851301115242)
};
\addplot[color=teal!80!black, fill=teal!20!white] coordinates {
    ({[50,150)},16.81159420289855) ({[150,250)},42.24186046511628) ({[250,350)},77.10852713178295) ({[350,450)},90.22304832713755)
};
\legend{Estimated vs. Generated, Generated vs. Specified}

\nextgroupplot[
    legend style={at={(0.5,-0.35)}, anchor=north, legend columns=2},
    ybar,
    ymin=0,
    ymax=55,
    ytick={0,10,...,50},
    symbolic x coords={{[50,150)}, {[150,250)}, {[250,350)}, {[350,450)}},
    xticklabel style={font=\sevenptword},
    xtick=data,
    enlarge x limits=0.2,
    xtick pos=bottom,
    ytick pos=left,
    bar width=11pt,
    title style={yshift=-12.25em},
    title={(c) LLaMA-3.1-8B-Instruct},
    axis lines*= left
]
\addplot[color=purple!80!black, fill=purple!20!white] coordinates {
    ({[50,150)},17.252032520325205) ({[150,250)},23.702290076335878) ({[250,350)},34.13084112149533) ({[350,450)},42.81439393939394)
};
\addplot[color=teal!80!black, fill=teal!20!white] coordinates {
    ({[50,150)},8.252032520325203) ({[150,250)},17.545801526717558) ({[250,350)},32.46105919003115) ({[350,450)},43.07954545454545)
};

\end{groupplot}
\end{tikzpicture}}
\caption{Mean Absolute Error (MAE) between (1) the estimated and actual generated length, and (2) the generated and user-specified length, across varying ranges of words for (a) Qwen3-4B, (b) Qwen3-8B, or (c) LLaMA-3.1-8B-Instruct.}
\label{fig:pilot_study_word}
\end{figure*}

%% file: table/length_estimation_sentence.tex
\newcommand{\sevenptsentence}{\fontsize{7.6pt}{9pt}\selectfont}
\pgfplotsset{compat=1.3,
    /pgfplots/ybar legend/.style={
    /pgfplots/legend image code/.code={%
       \draw[##1,/tikz/.cd,yshift=-0.25em]
        (0cm,0cm) rectangle (7pt,0.8em);},
   },
}
\begin{figure*}[!ht]
\hspace*{-0.5cm}
\pgfplotsset{width=6.5cm, height=4.5cm}
\centering
\scalebox{0.90}{\begin{tikzpicture}
\begin{groupplot}[
    group style={
        group name=my plots,
        group size=3 by 1,
        horizontal sep=0.8cm,
    },
]

\nextgroupplot[
    ybar,
    ymin=0,
    ymax=14,
    ytick={0,3,...,12},
    ylabel={MAE~$\downarrow$},
    y label style={yshift=-0.6em},
    symbolic x coords={{[5,10)}, {[10,15)}, {[15,20)}, {[20,25)}, {[25,30)}},
    xticklabel style={font=\sevenptsentence},
    xtick=data,
    enlarge x limits=0.2,
    xtick pos=bottom,
    ytick pos=left,
    bar width=8.5pt,
    title style={yshift=-12.25em},
    title={(a) Qwen3-4B},
    axis lines*= left,
]
\addplot[color=purple!80!black, fill=purple!20!white] coordinates {
    ({[5,10)},2.8962264150943398) ({[10,15)},5.171875) ({[15,20)},6.151515151515151) ({[20,25)},9.061855670103093) ({[25,30)},11.06)
};
\addplot[color=teal!80!black, fill=teal!20!white] coordinates {
    ({[5,10)},0.3113207547169811) ({[10,15)},1.921875) ({[15,20)},5.424242424242424) ({[20,25)},10.371134020618557) ({[25,30)},13.06)
};

\nextgroupplot[
    ybar,
    ymin=0,
    ymax=5.4,
    ytick={0,1,...,5},
    symbolic x coords={{[5,10)}, {[10,15)}, {[15,20)}, {[20,25)}, {[25,30)}},
    xticklabel style={font=\sevenptsentence},
    xtick=data,
    enlarge x limits=0.2,
    xtick pos=bottom,
    ytick pos=left,
    bar width=8.5pt,
    title style={yshift=-12.25em},
    title={(b) Qwen3-8B},
    axis lines*= left,
    legend style={
            at={(0.5,1)},
            anchor=south,
            column sep=0.5ex,
            /tikz/every even column/.append style={column sep=1.5em},
            font=\small,
            draw=none,
            legend columns=2,
        },
]
\addplot[color=purple!80!black, fill=purple!20!white]  coordinates {
    ({[5,10)},0.49264705882352944) ({[10,15)},0.8042328042328042) ({[15,20)},0.976) ({[20,25)},0.9615384615384616) ({[25,30)},2.112676056338028)
};
\addplot[color=teal!80!black, fill=teal!20!white] coordinates {
    ({[5,10)},0.16176470588235295) ({[10,15)},0.5396825396825397) ({[15,20)},1.2) ({[20,25)},2.853846153846154) ({[25,30)},3.316901408450704)
};
\legend{Estimated vs. Generated, Generated vs. Specified}

\nextgroupplot[
    legend style={at={(0.5,-0.35)}, anchor=north, legend columns=2},
    ybar,
    ymin=0,
    ymax=5.4,
    ytick={0,1,...,5},
    symbolic x coords={{[5,10)}, {[10,15)}, {[15,20)}, {[20,25)}, {[25,30)}},
    xticklabel style={font=\sevenptsentence},
    xtick=data,
    enlarge x limits=0.2,
    xtick pos=bottom,
    ytick pos=left,
    bar width=8.5pt,
    title style={yshift=-12.25em},
    title={(c) LLaMA-3.1-8B-Instruct},
    axis lines*= left
]
\addplot[color=purple!80!black, fill=purple!20!white] coordinates {
    ({[5,10)},0.4041095890410959) ({[10,15)},1.3980099502487562) ({[15,20)},1.860759493670886) ({[20,25)},3.1307692307692307) ({[25,30)},5.165137614678899)
};
\addplot[color=teal!80!black, fill=teal!20!white] coordinates {
    ({[5,10)},0.07534246575342465) ({[10,15)},0.1890547263681592) ({[15,20)},0.7025316455696202) ({[20,25)},1.376923076923077) ({[25,30)},3.036697247706422)
};

\end{groupplot}
\end{tikzpicture}}
\caption{Mean Absolute Error (MAE) across varying ranges of sentences.}
\label{fig:pilot_study_sentence}
\end{figure*}

%% file: appendix_fig/summary_qwen3_4B.tex
\begin{figure*}[!ht]
\centering
    \begin{subfigure}{.340\textwidth}
        \includegraphics[width=\linewidth]{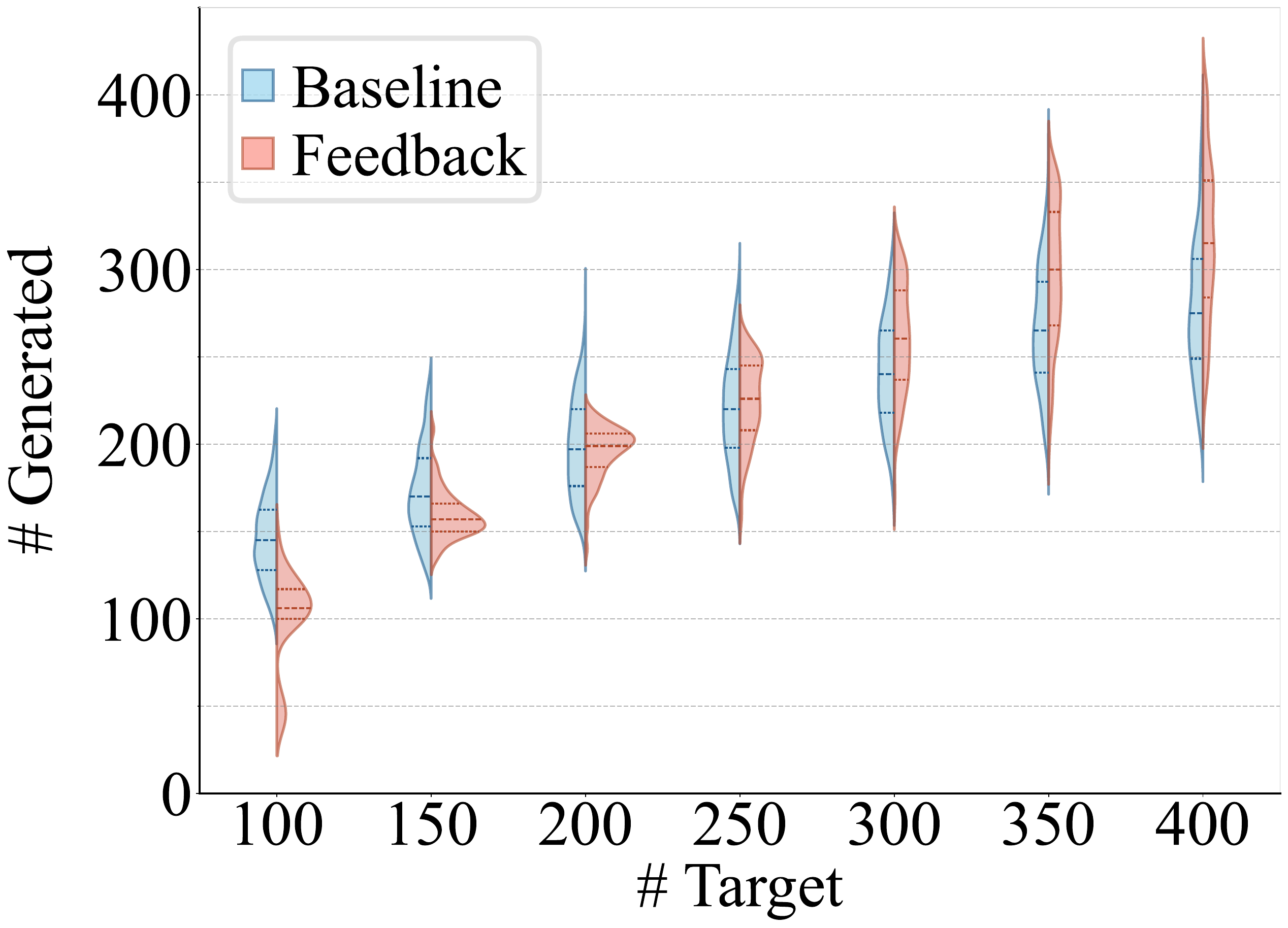} 
        \caption{Token}
        \label{subfig:summary_qwen3_4B_prompt_a}
    \end{subfigure}
    \hfill
    \begin{subfigure}{.32\textwidth}
        \includegraphics[width=\linewidth]{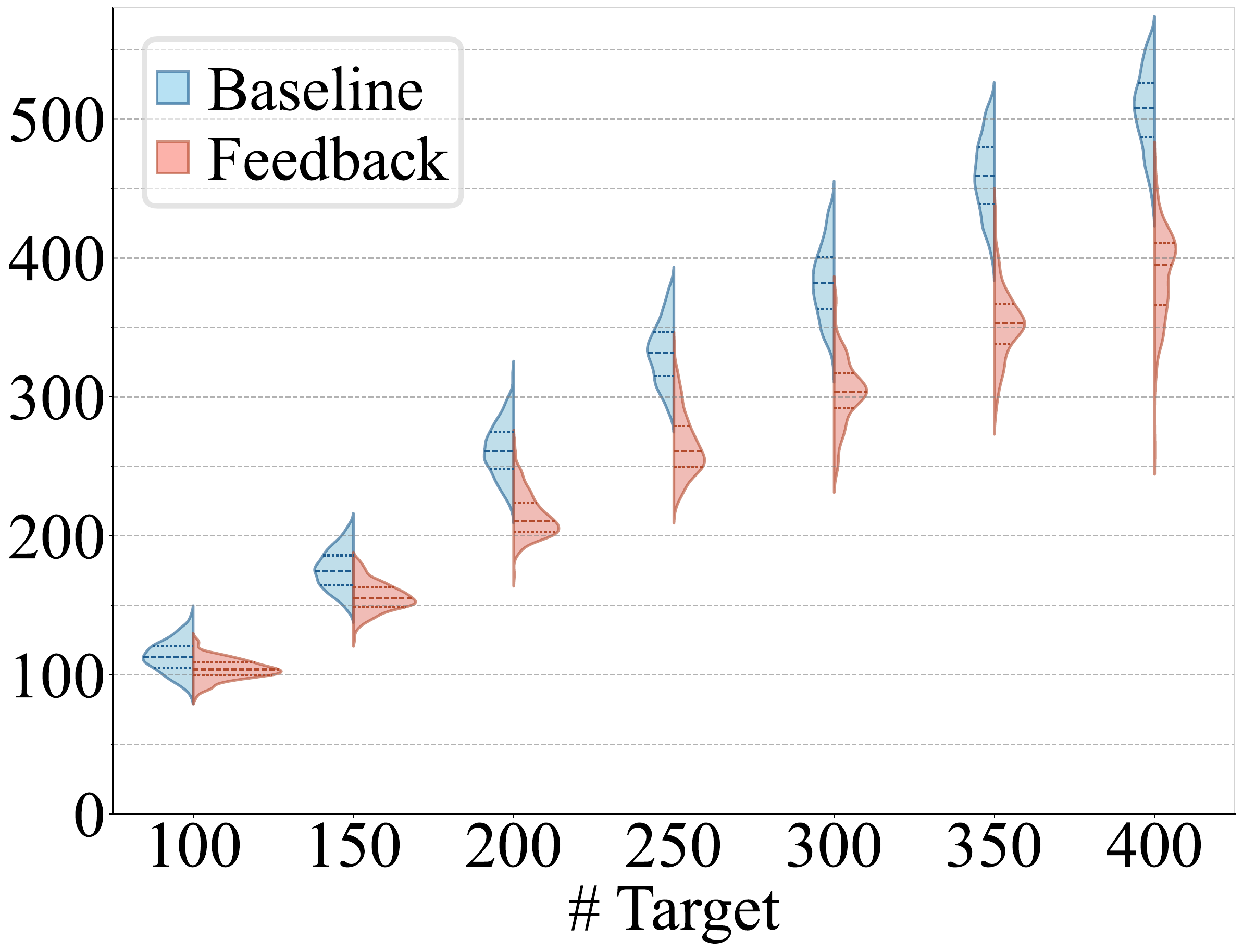} 
        \caption{Word}
        \label{subfig:summary_qwen3_4B_prompt_b}
    \end{subfigure}
    \hfill
    \begin{subfigure}{.32\textwidth}
        \includegraphics[width=\linewidth]{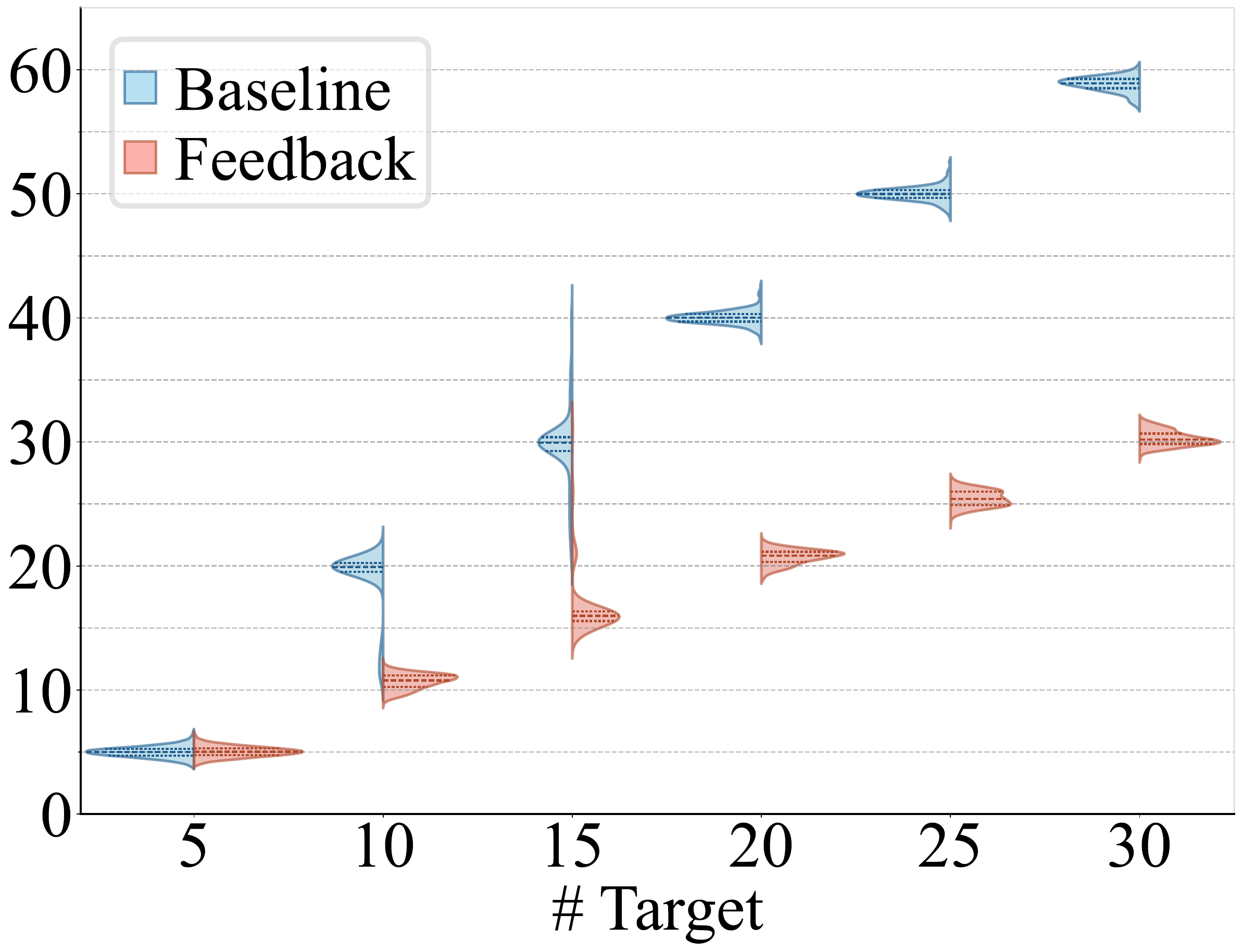} 
        \caption{Sentence}
        \label{subfig:summary_qwen3_4B_prompt_c}
    \end{subfigure}
    \begin{subfigure}{.340\textwidth}
        \includegraphics[width=\linewidth]{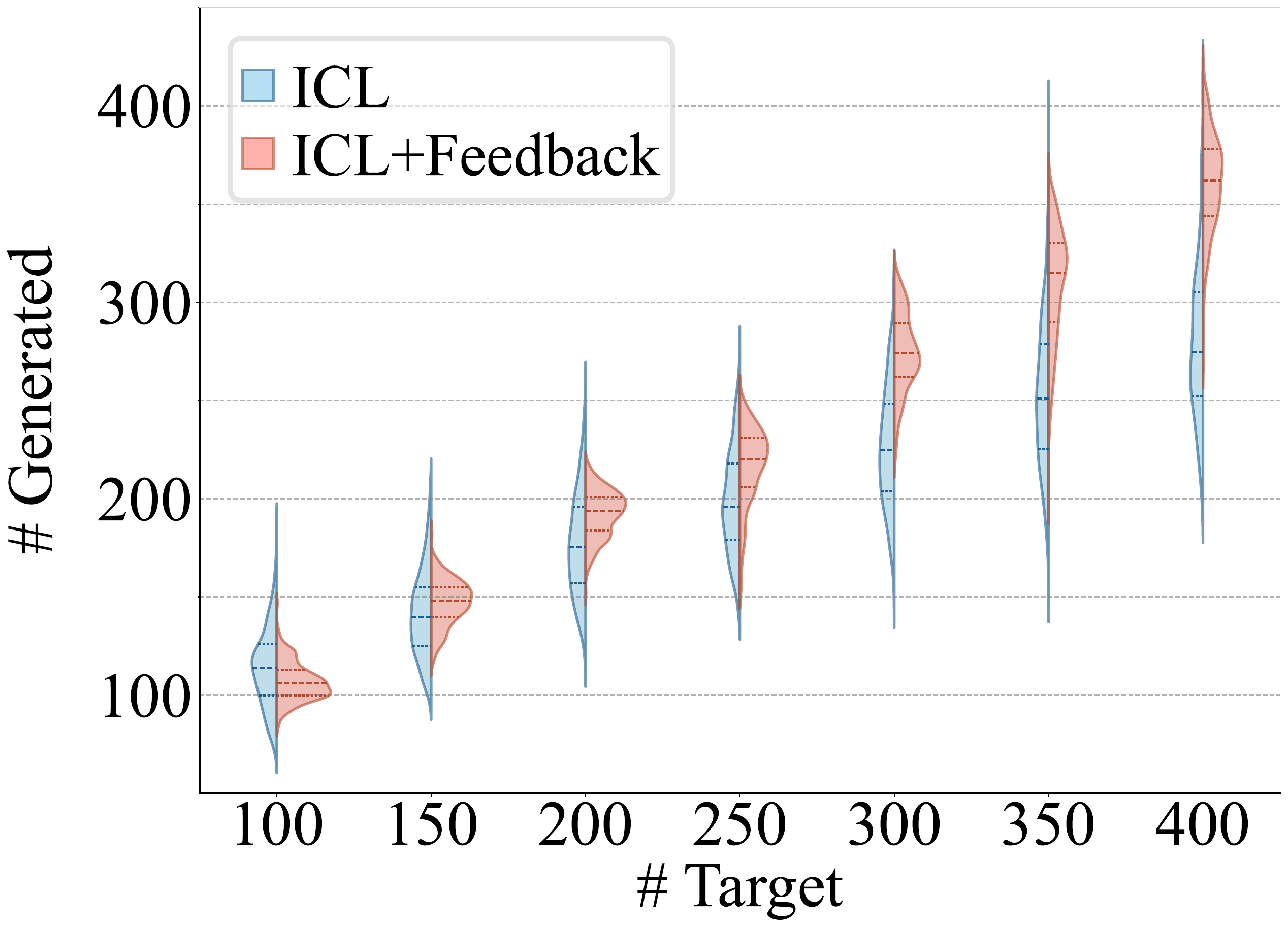} 
        \caption{Token (ICL)}
        \label{subfig:summary_qwen3_4B_icl_a}
    \end{subfigure}
    \hfill
    \begin{subfigure}{.32\textwidth}
        \includegraphics[width=\linewidth]{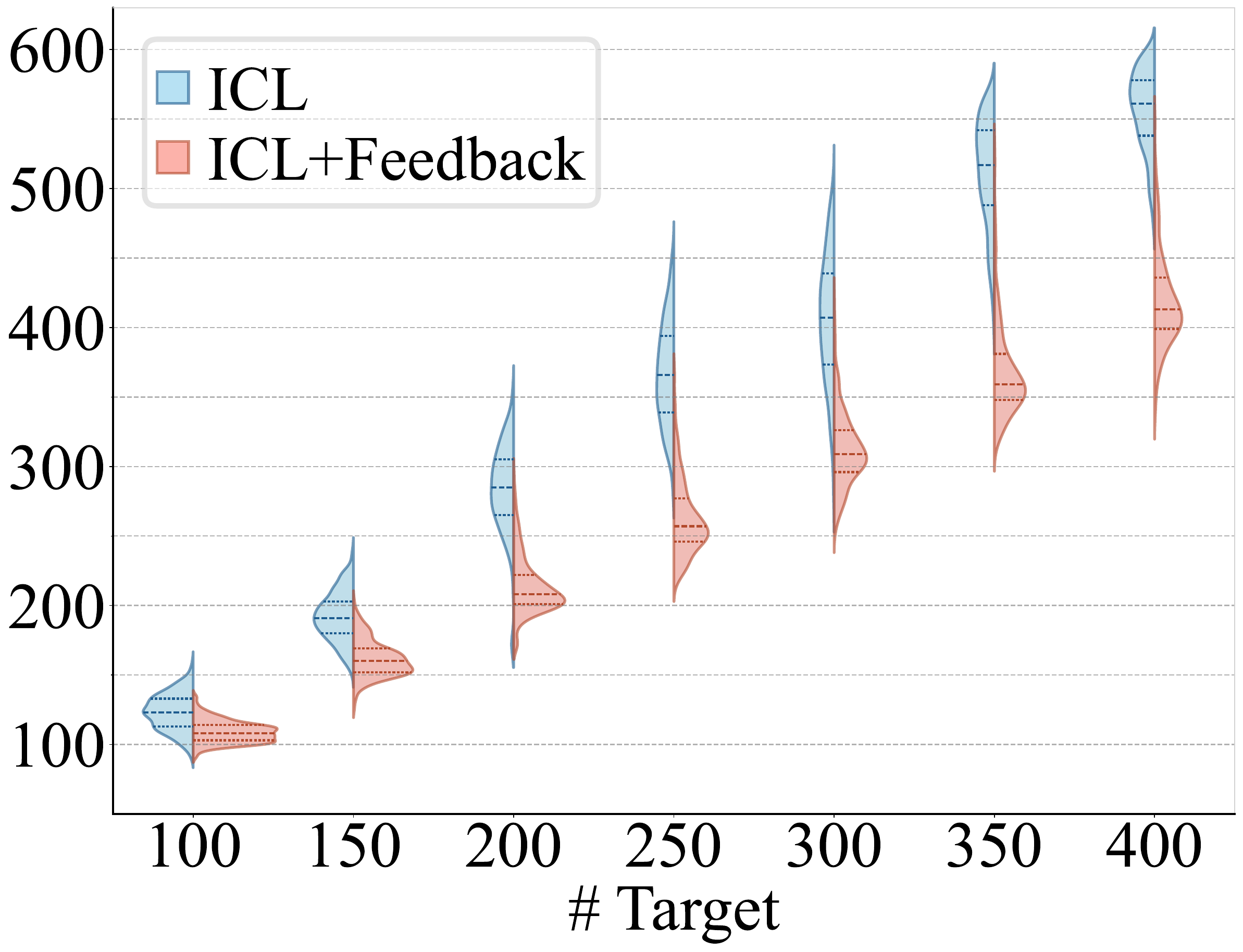} 
        \caption{Word (ICL)}
        \label{subfig:summary_qwen3_4B_icl_b}
    \end{subfigure}
    \hfill
    \begin{subfigure}{.32\textwidth}
        \includegraphics[width=\linewidth]{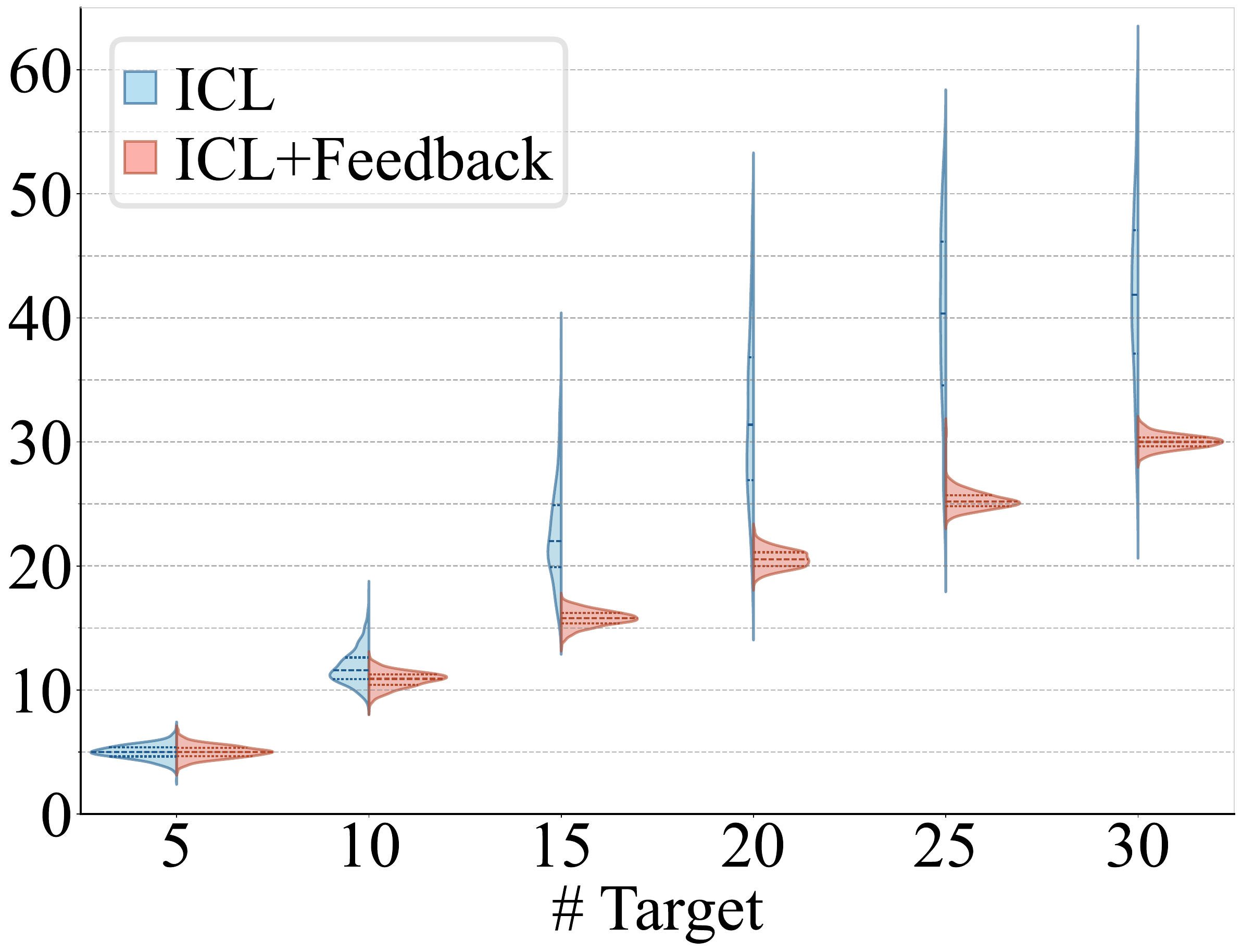} 
        \caption{Sentence (ICL)}
        \label{subfig:summary_qwen3_4B_icl_c}
    \end{subfigure}
    \caption{Generated length distributions under varying target lengths on GovReport using Qwen3-4B.}
    \label{fig:summary_qwen3_4B_align}
\end{figure*}

%% file: appendix_fig/summary_qwen3_8B.tex
\begin{figure*}[!ht]
\centering
    \begin{subfigure}{.340\textwidth}
        \includegraphics[width=\linewidth]{figure/summary_align_token_2.pdf} 
        \caption{Token}
        \label{subfig:summary_qwen3_8B_prompt_a}
    \end{subfigure}
    \hfill
    \begin{subfigure}{.32\textwidth}
        \includegraphics[width=\linewidth]{figure/summary_align_word_2.pdf} 
        \caption{Word}
        \label{subfig:summary_qwen3_8B_prompt_b}
    \end{subfigure}
    \hfill
    \begin{subfigure}{.32\textwidth}
        \includegraphics[width=\linewidth]{figure/summary_align_sentence_2.pdf} 
        \caption{Sentence}
        \label{subfig:summary_qwen3_8B_prompt_c}
    \end{subfigure}
    \begin{subfigure}{.340\textwidth}
        \includegraphics[width=\linewidth]{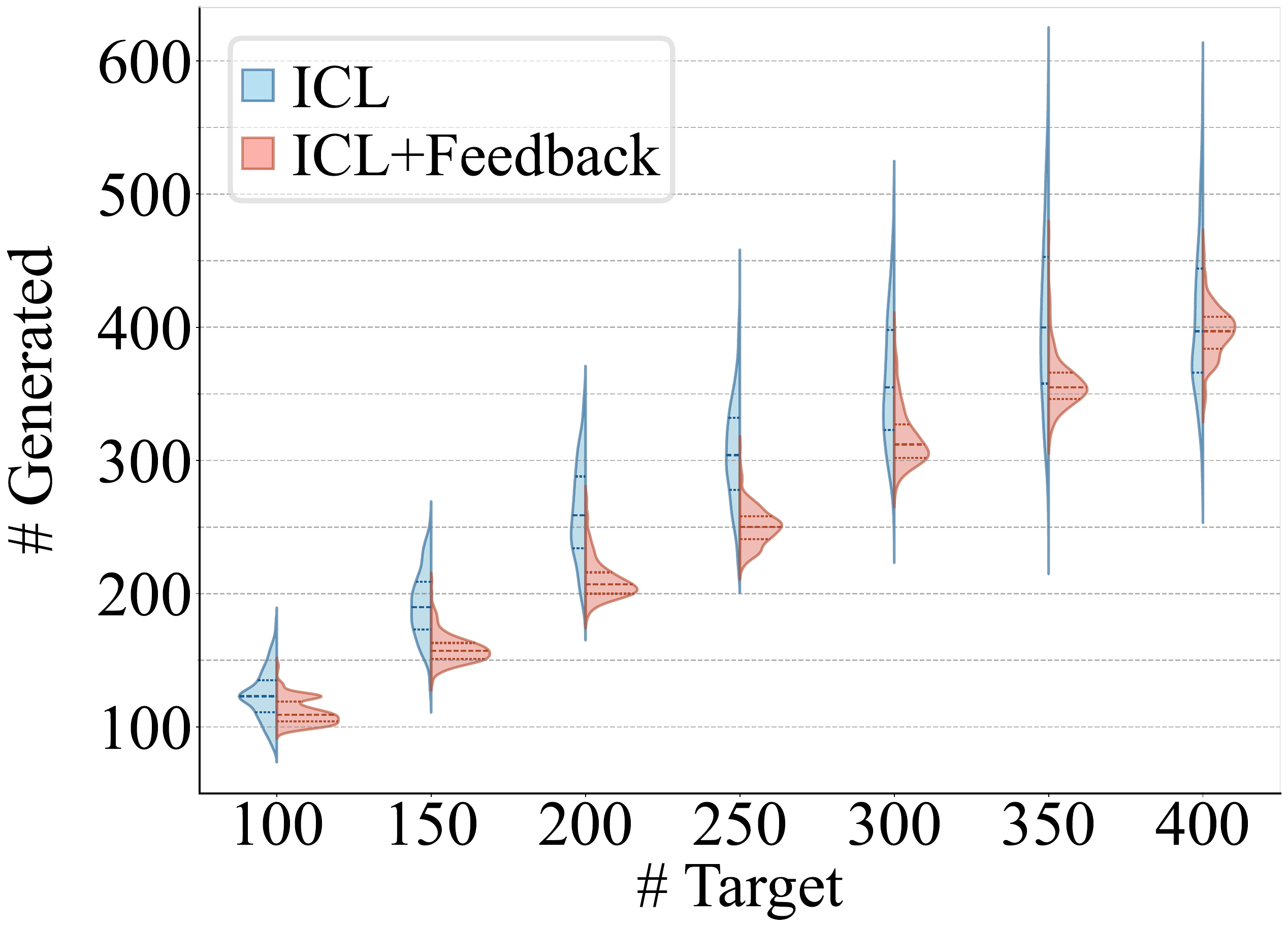} 
        \caption{Token (ICL)}
        \label{subfig:summary_qwen3_8B_icl_a}
    \end{subfigure}
    \hfill
    \begin{subfigure}{.32\textwidth}
        \includegraphics[width=\linewidth]{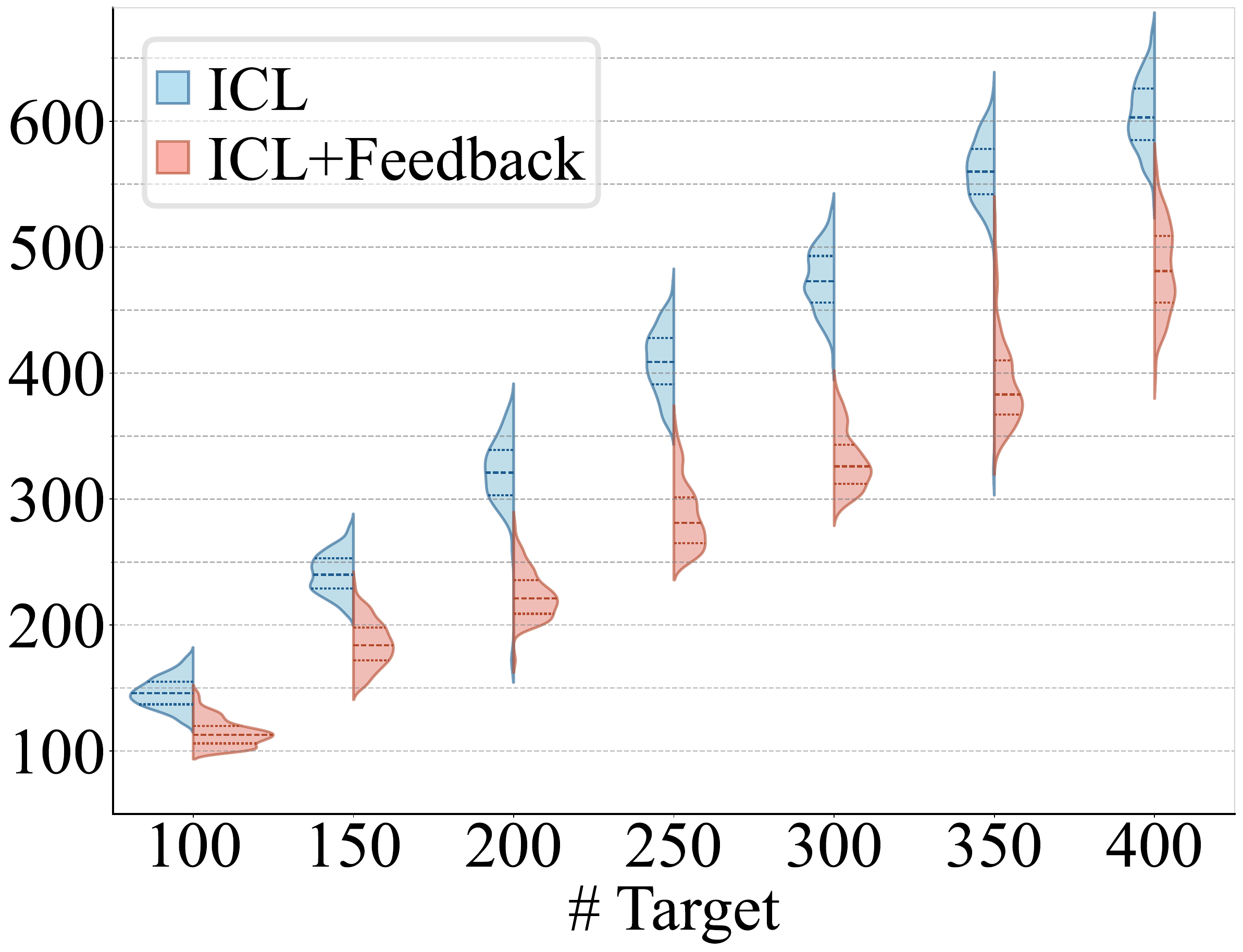} 
        \caption{Word (ICL)}
        \label{subfig:summary_qwen3_8B_icl_b}
    \end{subfigure}
    \hfill
    \begin{subfigure}{.32\textwidth}
        \includegraphics[width=\linewidth]{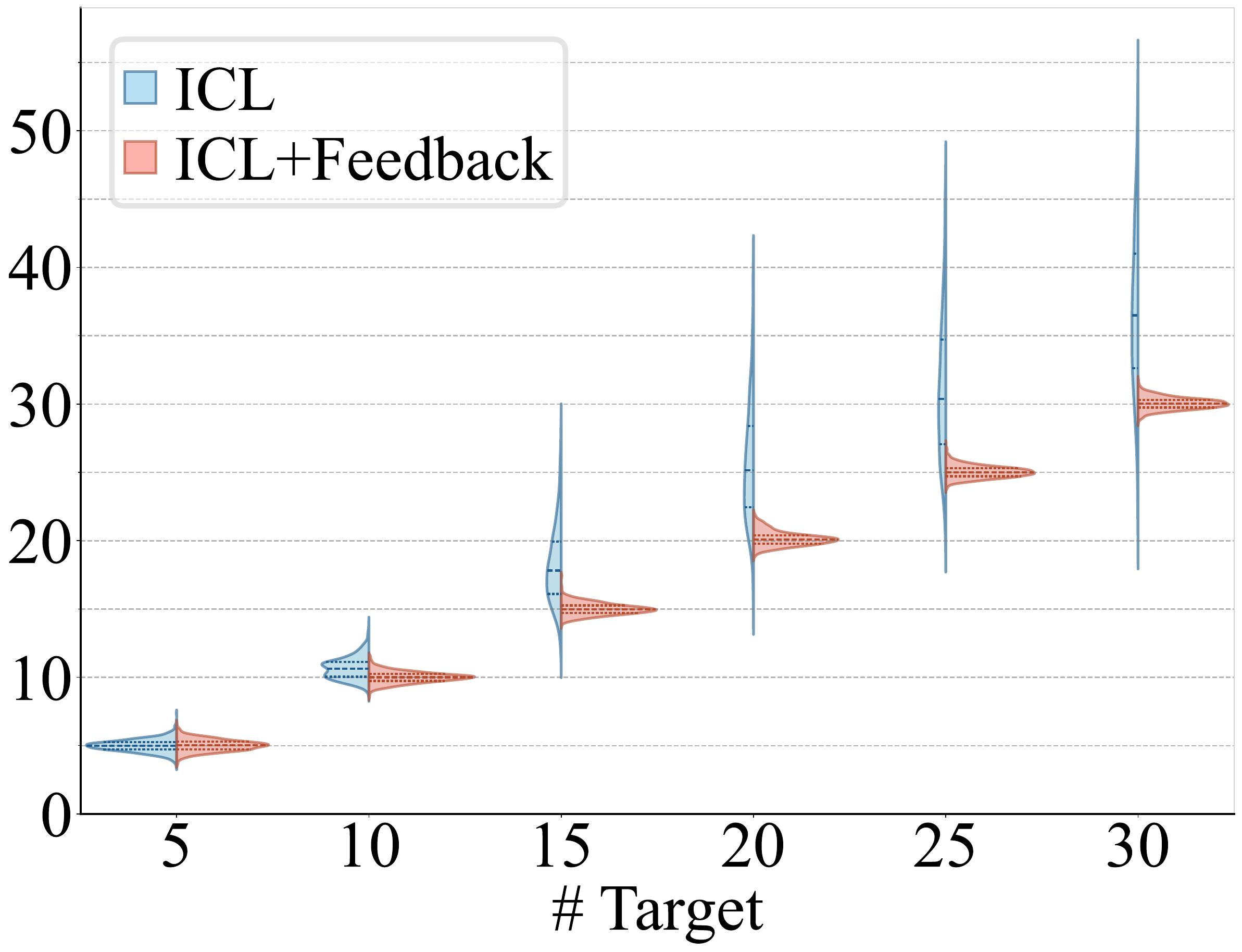} 
        \caption{Sentence (ICL)}
        \label{subfig:summary_qwen3_8B_icl_c}
    \end{subfigure}
    \caption{Generated length distributions under varying target lengths on GovReport using Qwen3-8B.}
    \label{fig:summary_qwen3_8B_align}
\end{figure*}

%% file: appendix_fig/summary_llama.tex
\begin{figure*}[!ht]
\centering
    \begin{subfigure}{.340\textwidth}
        \includegraphics[width=\linewidth]{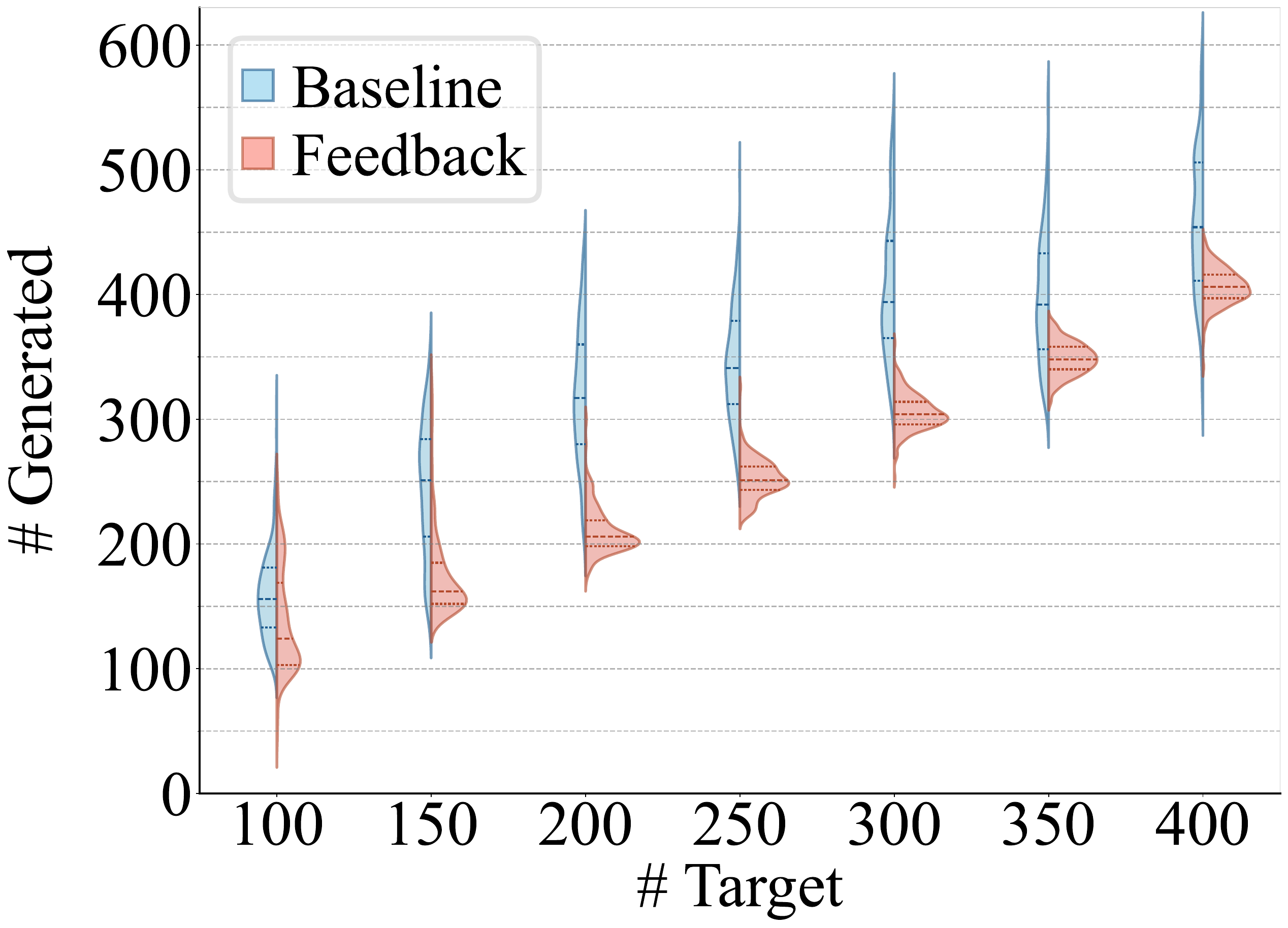} 
        \caption{Token}
        \label{subfig:summary_llama_prompt_a}
    \end{subfigure}
    \hfill
    \begin{subfigure}{.32\textwidth}
        \includegraphics[width=\linewidth]{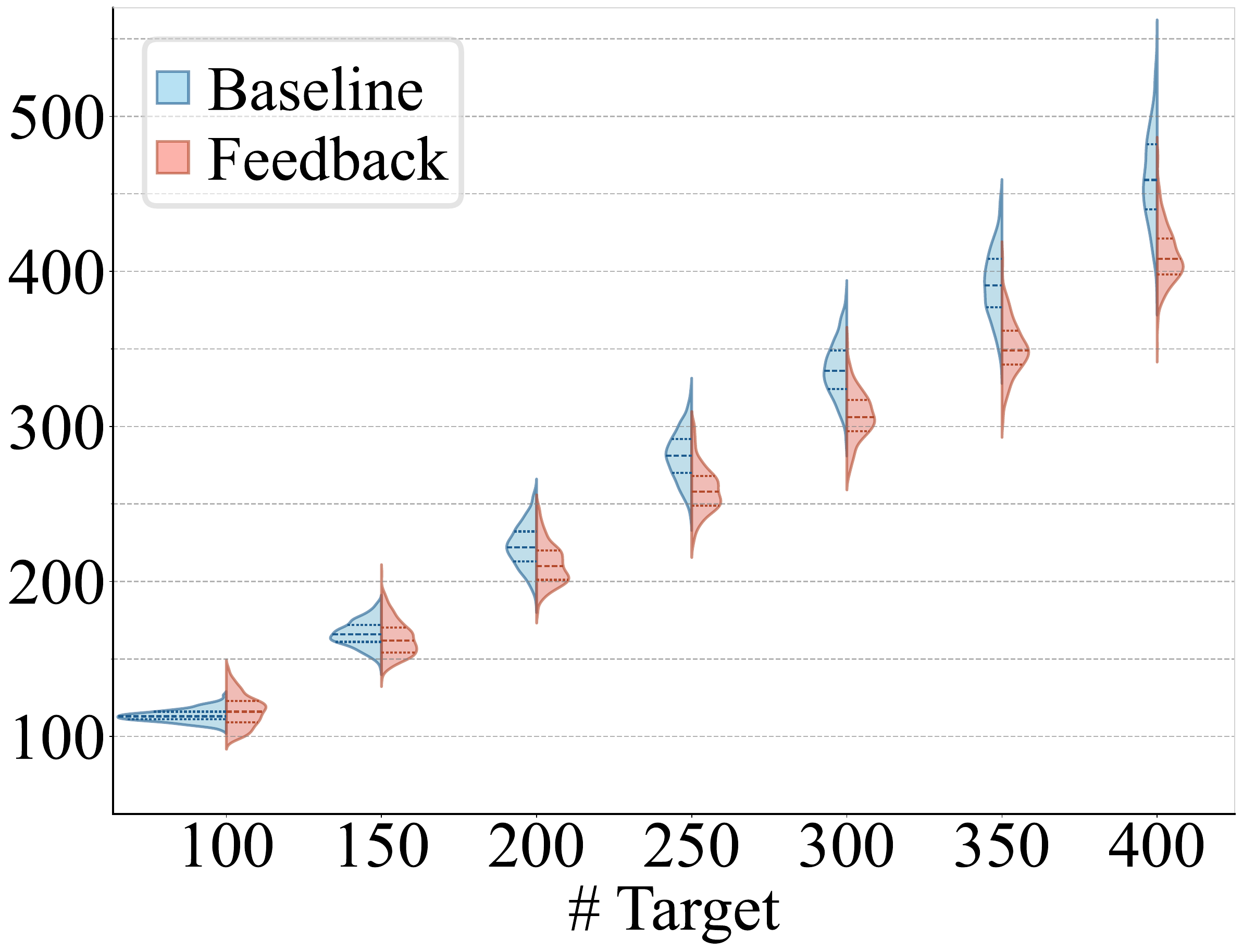} 
        \caption{Word}
        \label{subfig:summary_llama_prompt_b}
    \end{subfigure}
    \hfill
    \begin{subfigure}{.32\textwidth}
        \includegraphics[width=\linewidth]{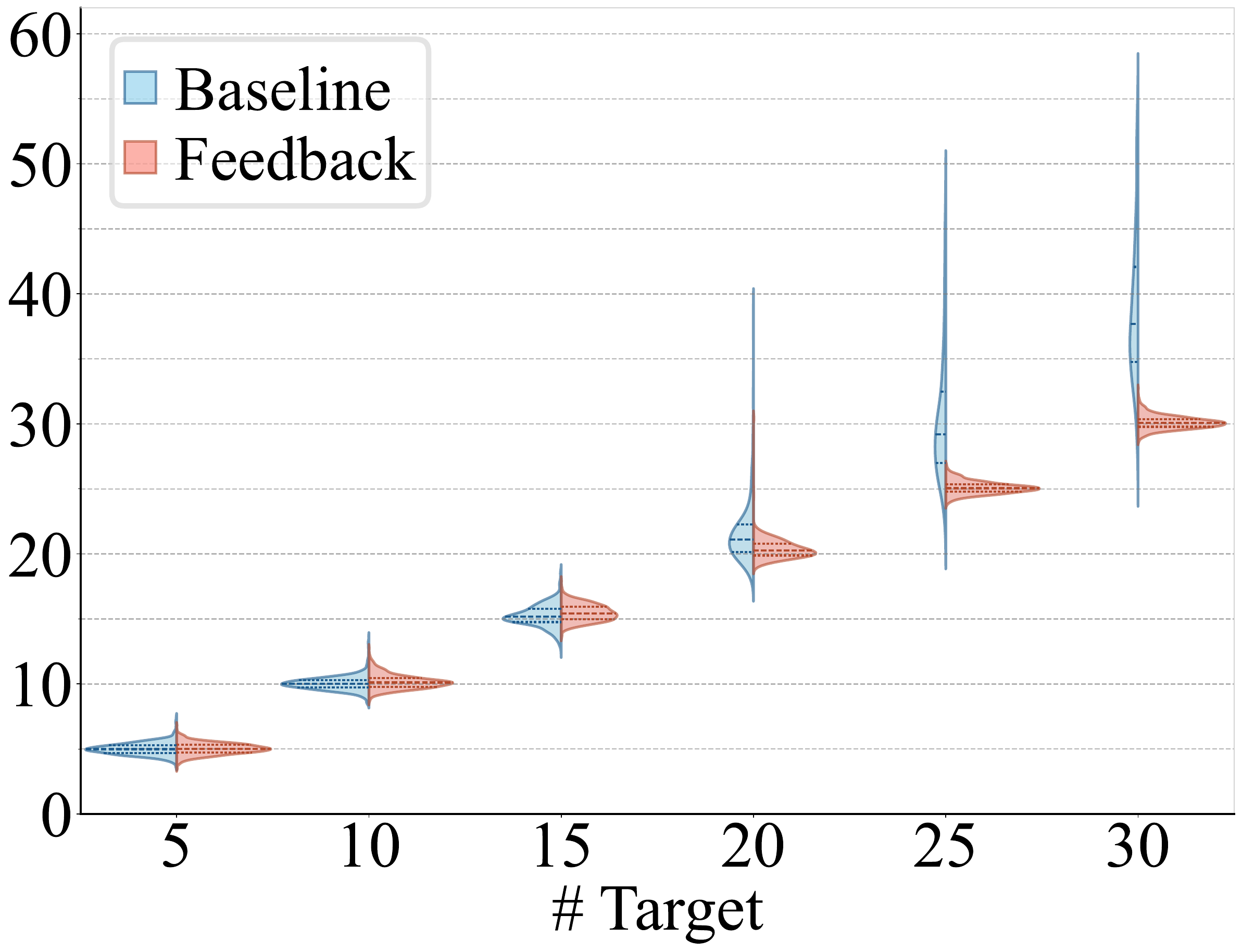} 
        \caption{Sentence}
        \label{subfig:summary_llama_prompt_c}
    \end{subfigure}
    \begin{subfigure}{.340\textwidth}
        \includegraphics[width=\linewidth]{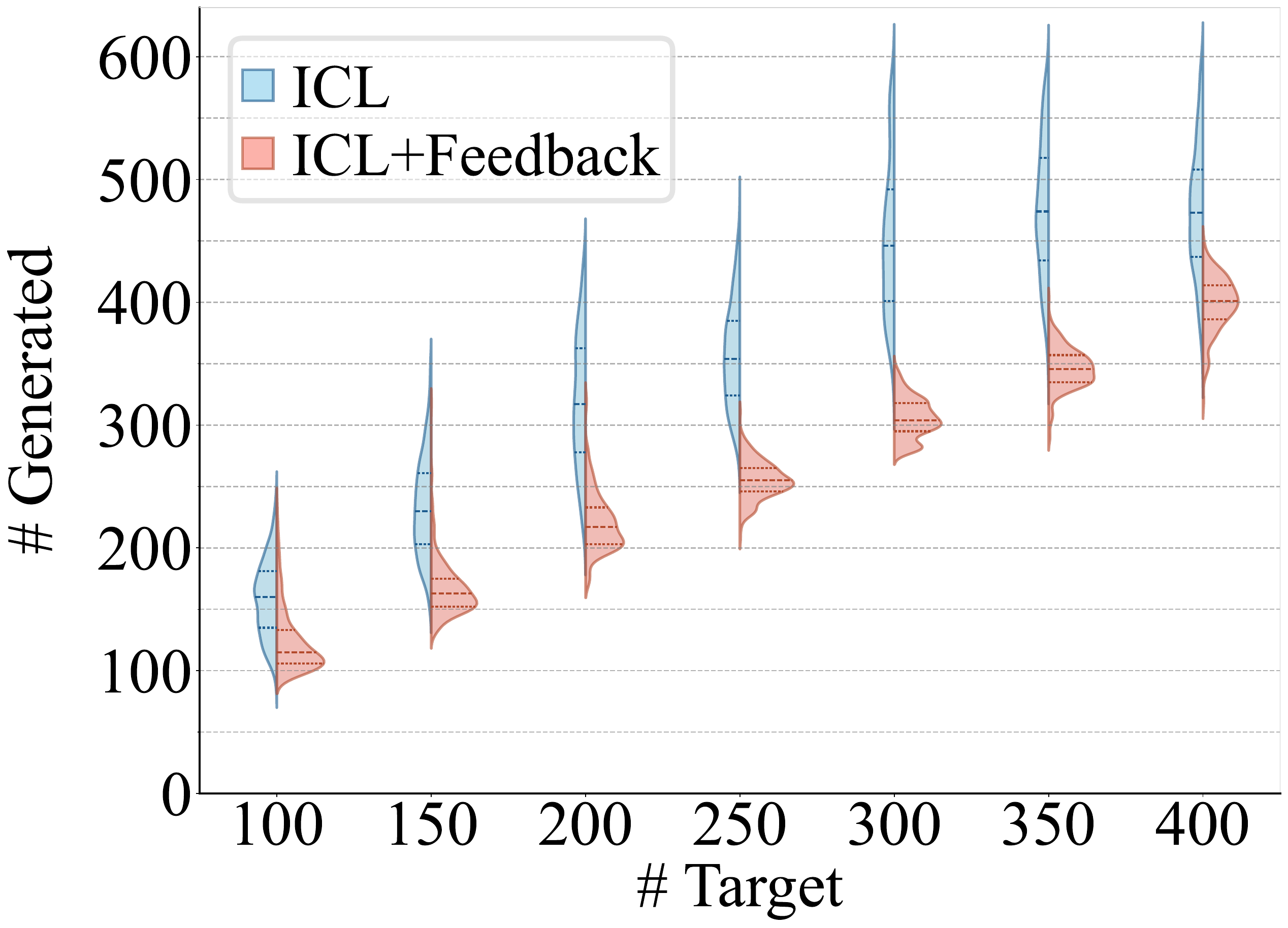} 
        \caption{Token (ICL)}
        \label{subfig:summary_llama_icl_a}
    \end{subfigure}
    \hfill
    \begin{subfigure}{.32\textwidth}
        \includegraphics[width=\linewidth]{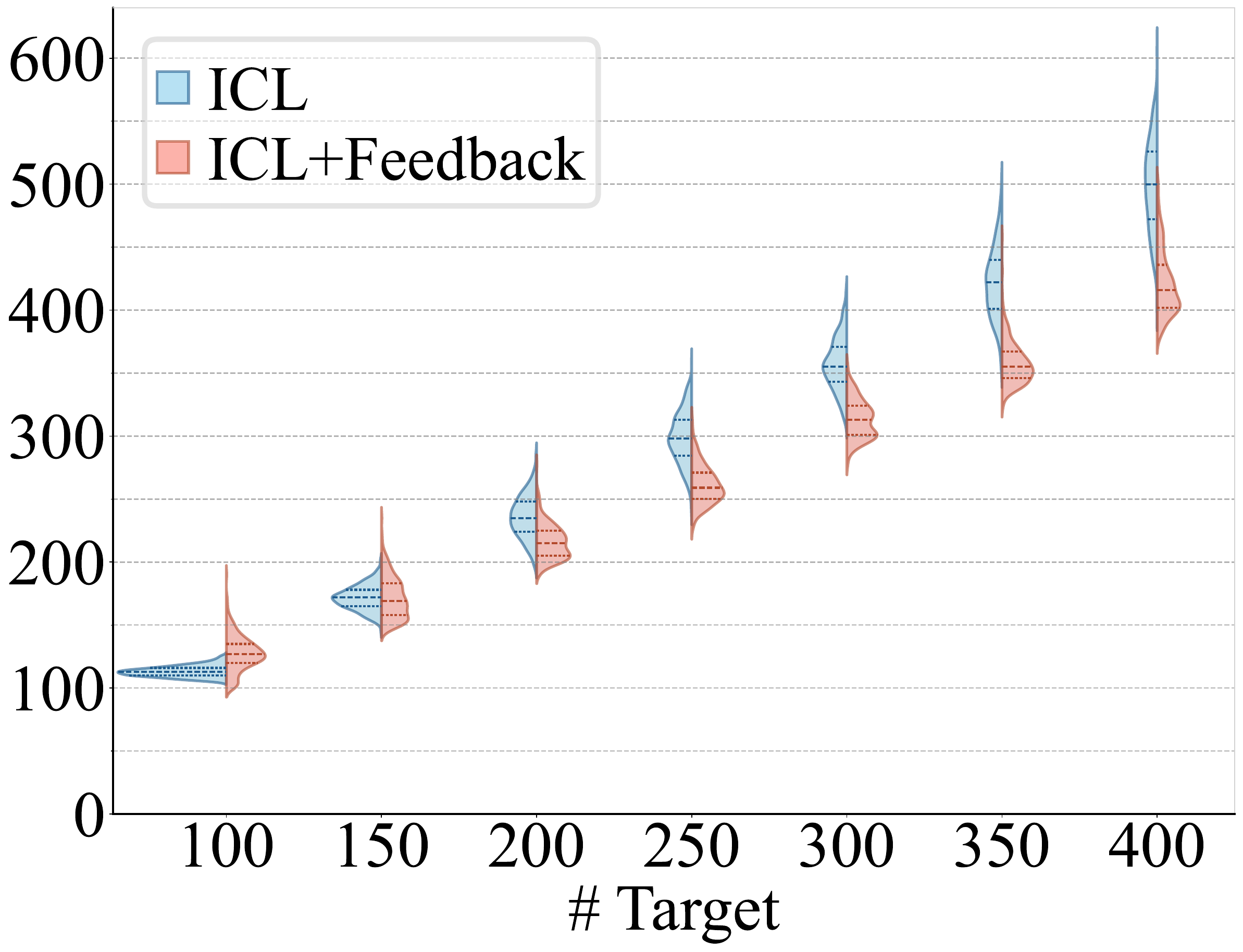} 
        \caption{Word (ICL)}
        \label{subfig:summary_llama_icl_b}
    \end{subfigure}
    \hfill
    \begin{subfigure}{.32\textwidth}
        \includegraphics[width=\linewidth]{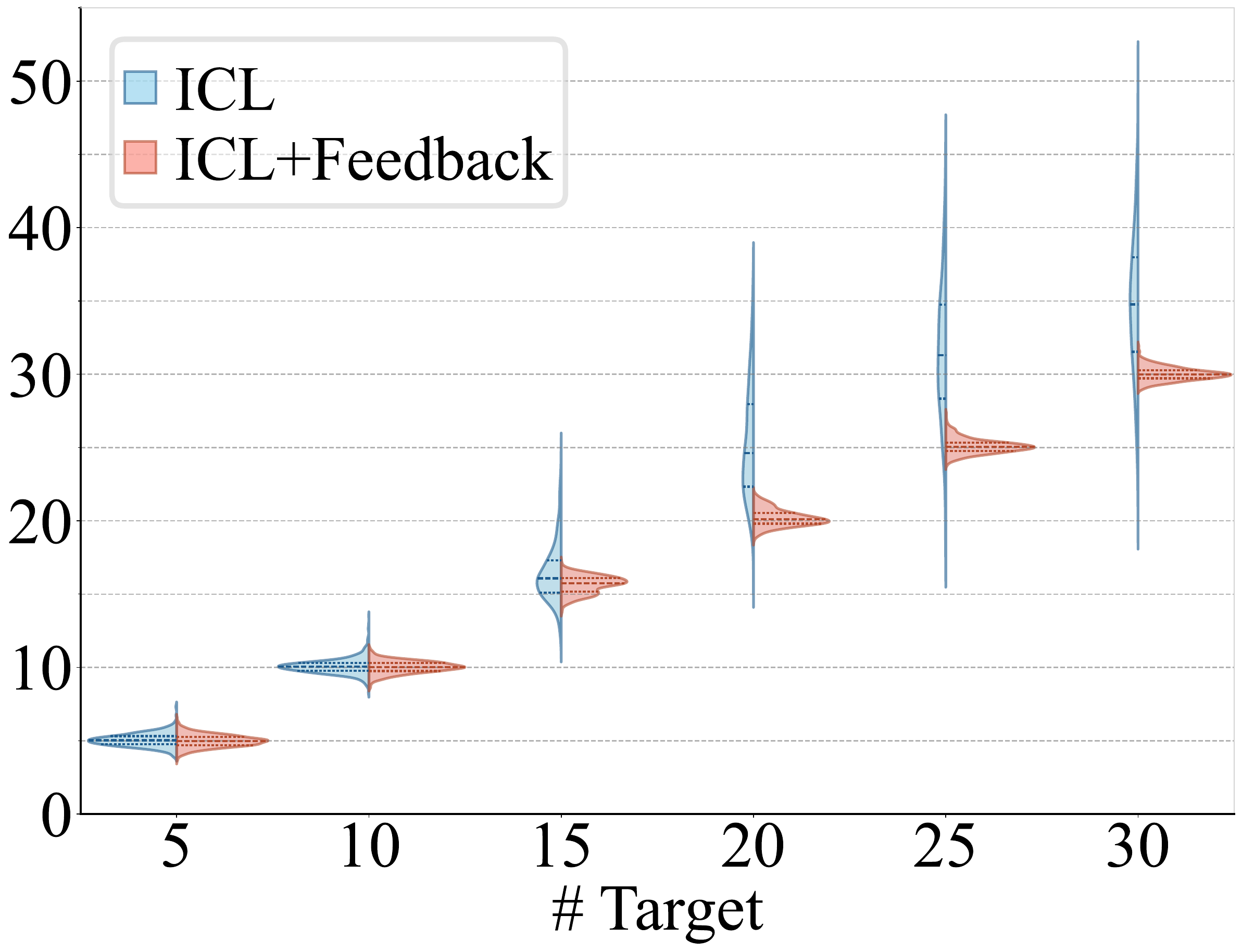} 
        \caption{Sentence (ICL)}
        \label{subfig:summary_llama_icl_c}
    \end{subfigure}
    \caption{Generated length distributions under varying target lengths on GovReport using LLaMA-3.1-8B-Instruct.}
    \label{fig:summary_llama_align}
\end{figure*}

%% file: appendix_fig/bio_qwen3_4B.tex
\begin{figure*}[!ht]
\centering
    \begin{subfigure}{.340\textwidth}
        \includegraphics[width=\linewidth]{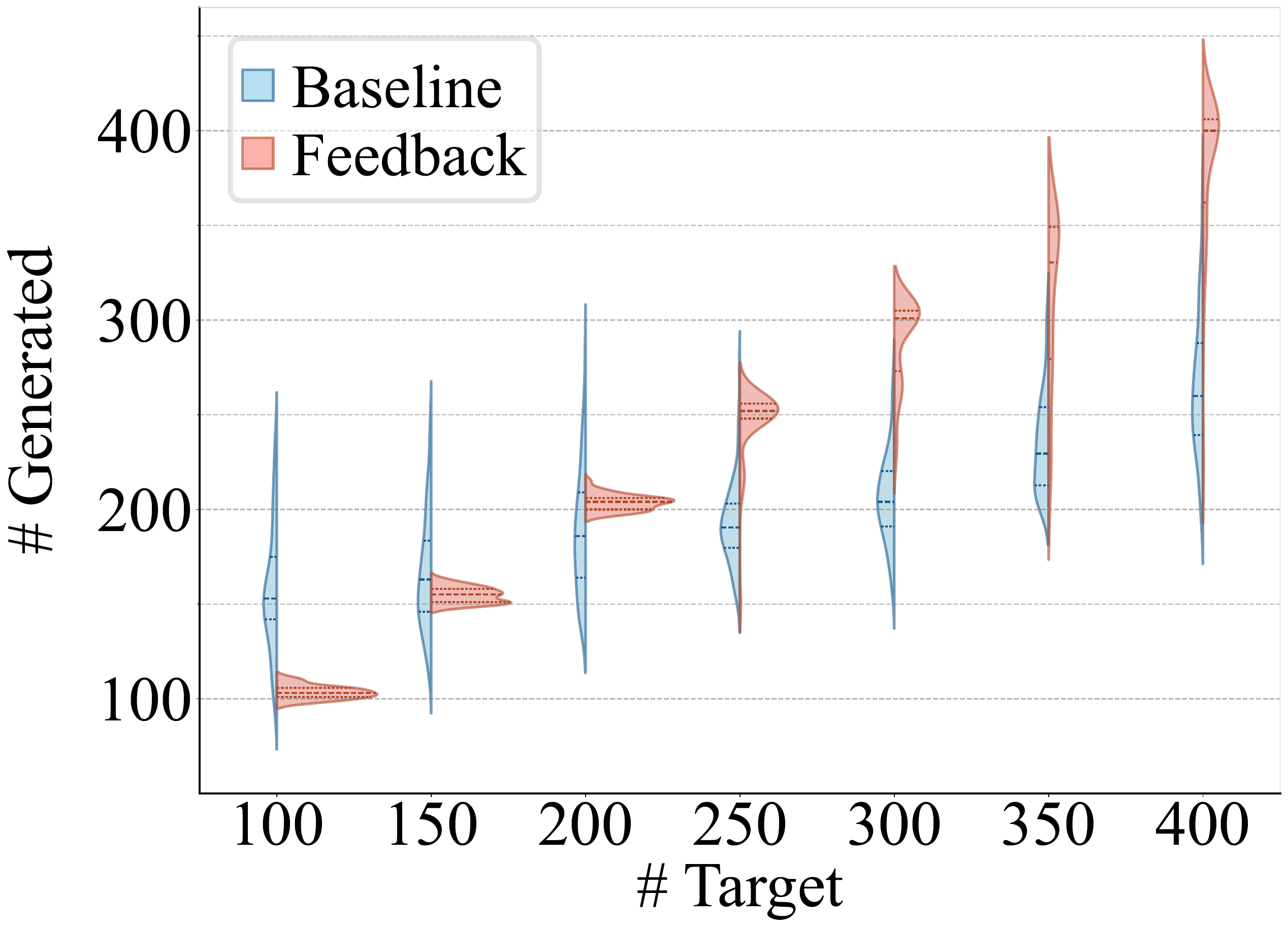} 
        \caption{Token}
        \label{subfig:bio_qwen3_4B_prompt_a}
    \end{subfigure}
    \hfill
    \begin{subfigure}{.32\textwidth}
        \includegraphics[width=\linewidth]{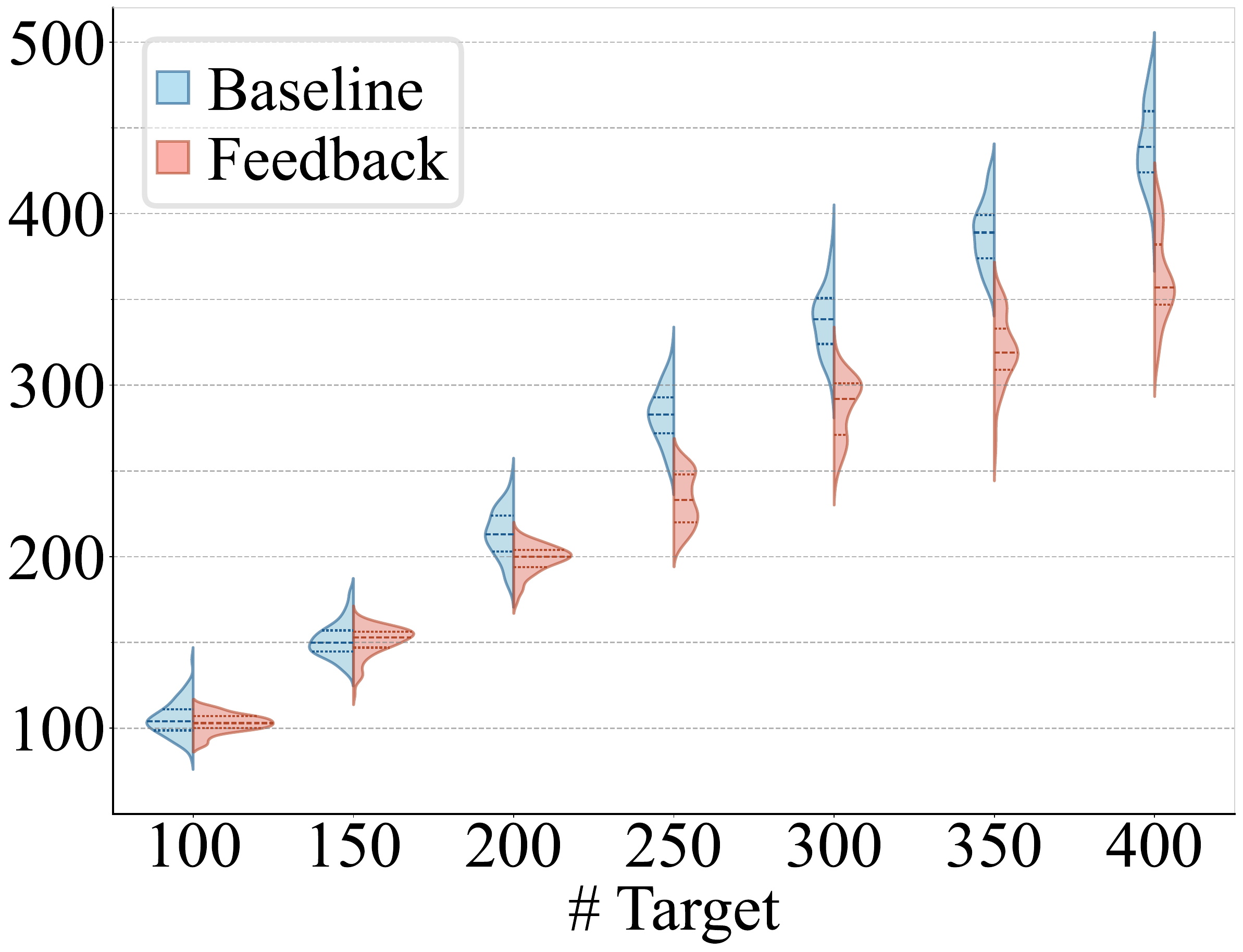} 
        \caption{Word}
        \label{subfig:bio_qwen3_4B_prompt_b}
    \end{subfigure}
    \hfill
    \begin{subfigure}{.32\textwidth}
        \includegraphics[width=\linewidth]{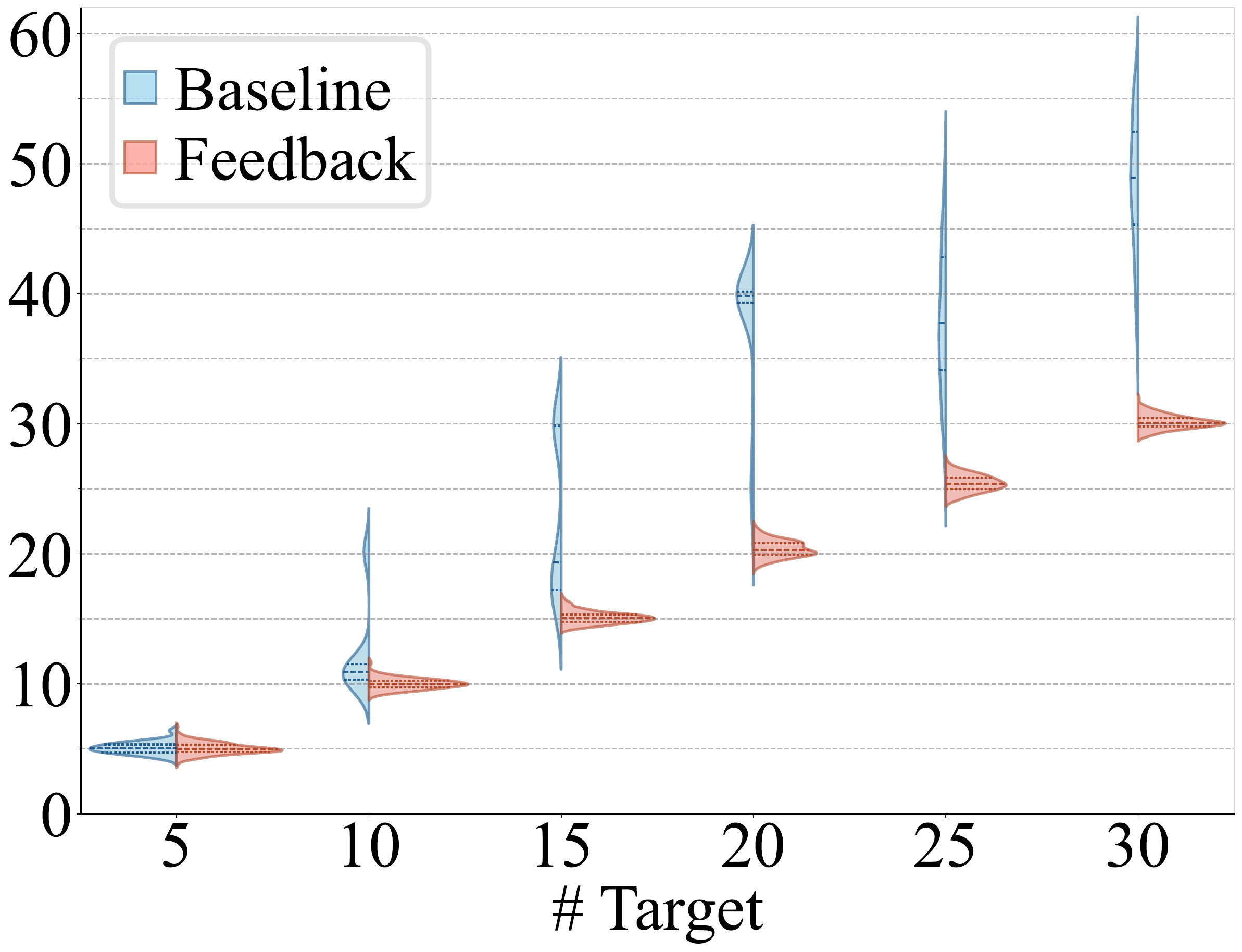} 
        \caption{Sentence}
        \label{subfig:bio_qwen3_4B_prompt_c}
    \end{subfigure}
    \begin{subfigure}{.340\textwidth}
        \includegraphics[width=\linewidth]{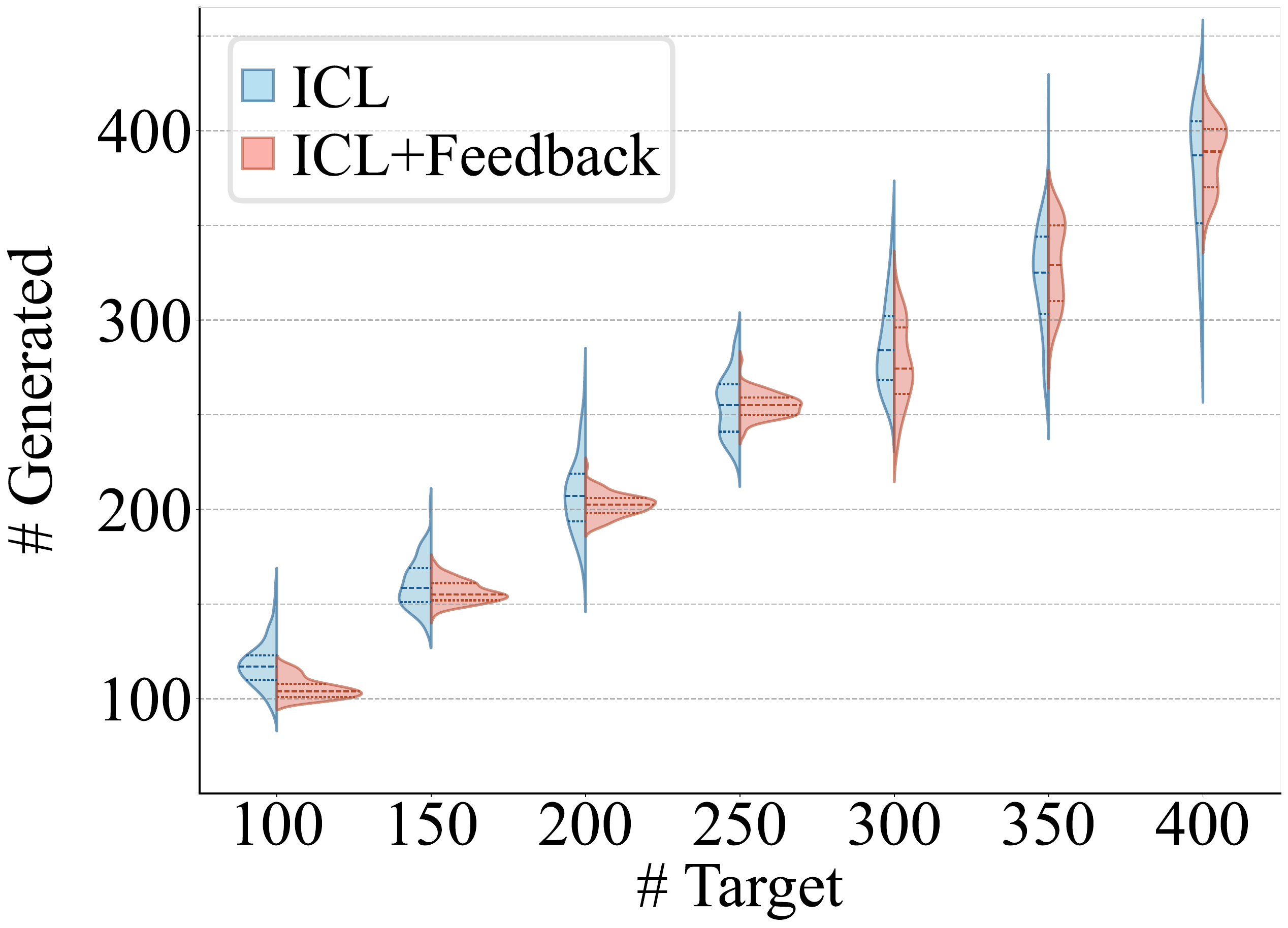} 
        \caption{Token (ICL)}
        \label{subfig:bio_qwen3_4B_icl_a}
    \end{subfigure}
    \hfill
    \begin{subfigure}{.32\textwidth}
        \includegraphics[width=\linewidth]{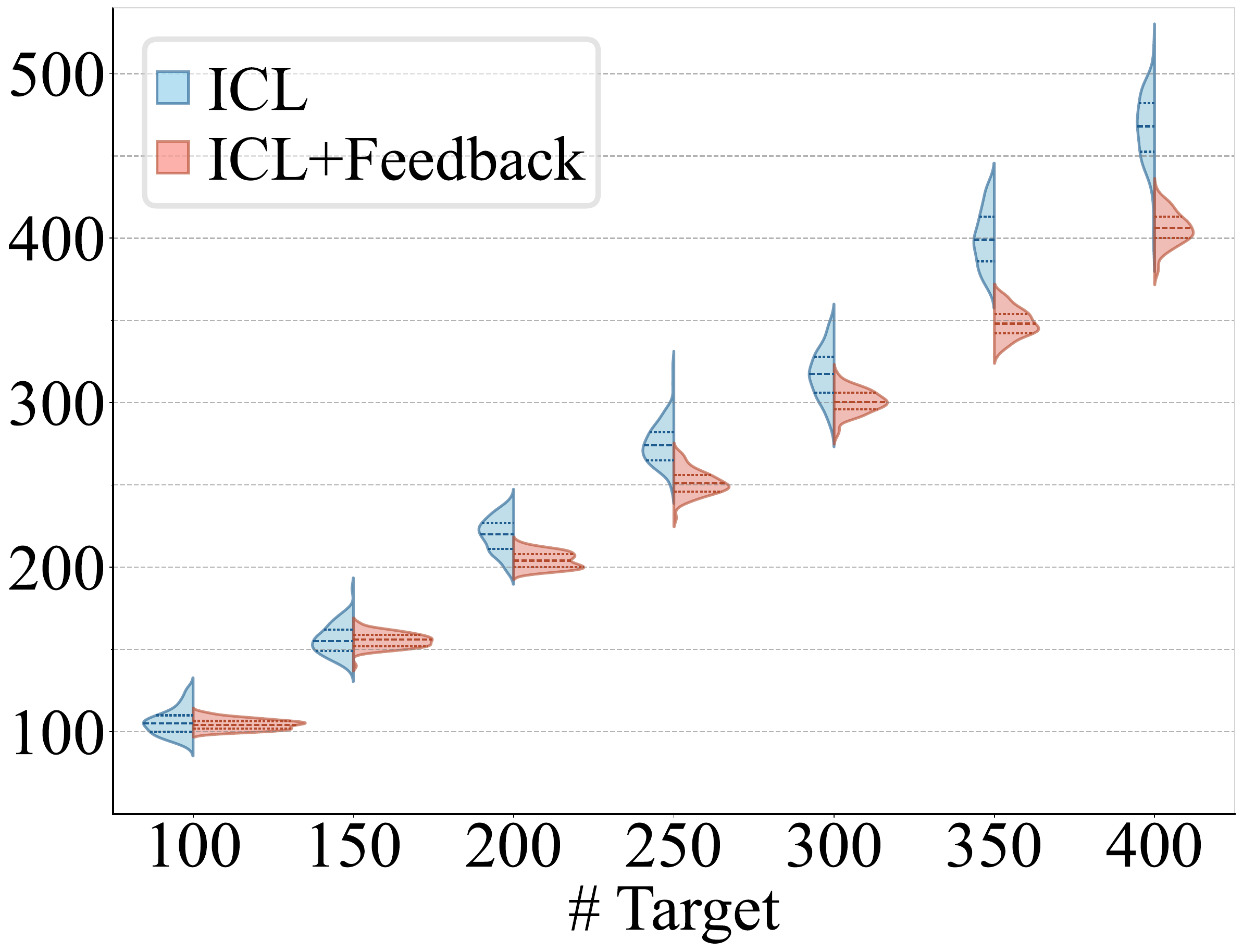} 
        \caption{Word (ICL)}
        \label{subfig:bio_qwen3_4B_icl_b}
    \end{subfigure}
    \hfill
    \begin{subfigure}{.32\textwidth}
        \includegraphics[width=\linewidth]{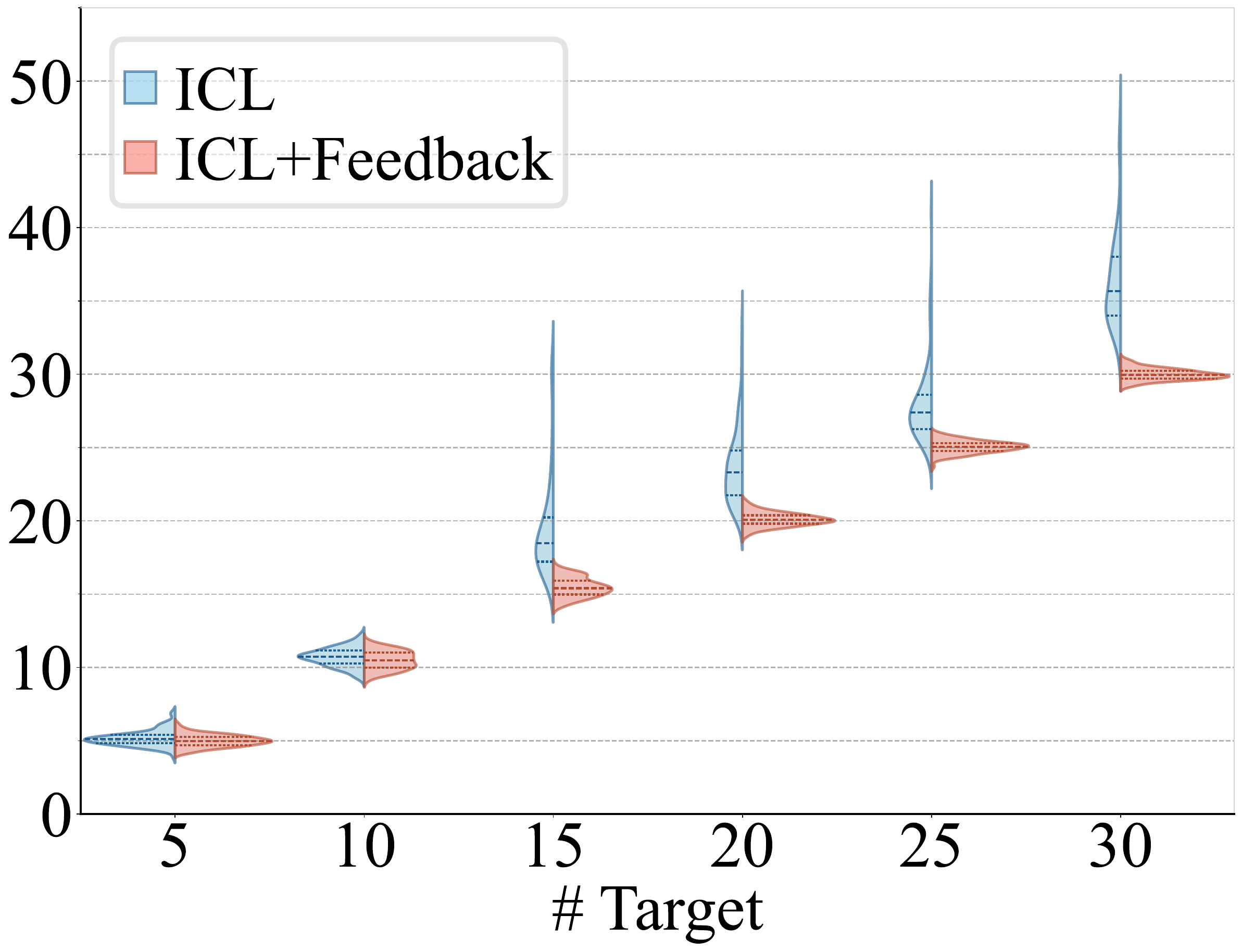} 
        \caption{Sentence (ICL)}
        \label{subfig:bio_qwen3_4B_icl_c}
    \end{subfigure}
    \caption{Generated length distributions under varying target lengths on Biographies using Qwen3-4B.}
    \label{fig:bio_qwen3_4B_align}
\end{figure*}

%% file: appendix_fig/bio_qwen3_8B.tex
\begin{figure*}[!ht]
\centering
    \begin{subfigure}{.340\textwidth}
        \includegraphics[width=\linewidth]{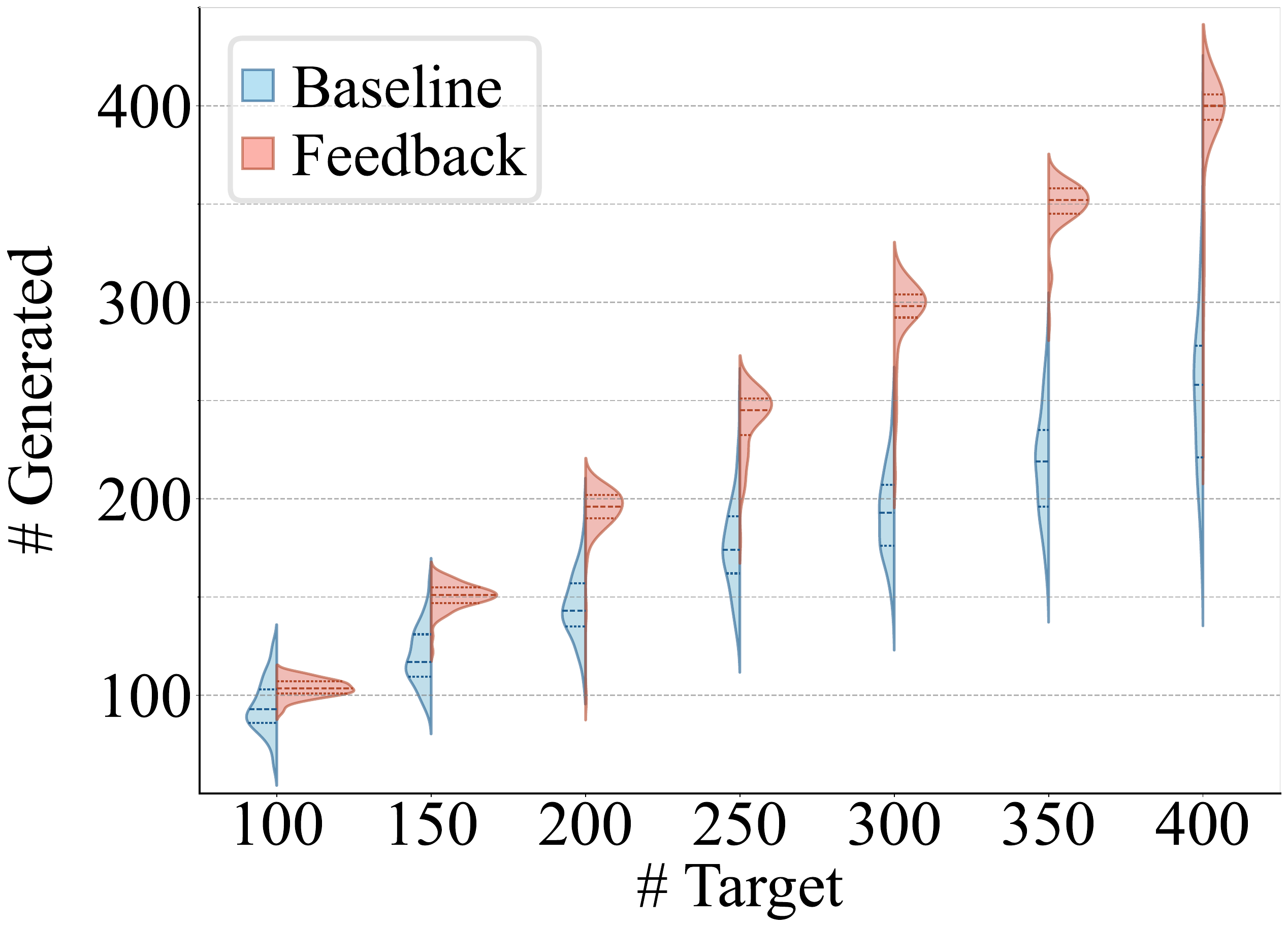} 
        \caption{Token}
        \label{subfig:bio_qwen3_8B_prompt_a}
    \end{subfigure}
    \hfill
    \begin{subfigure}{.32\textwidth}
        \includegraphics[width=\linewidth]{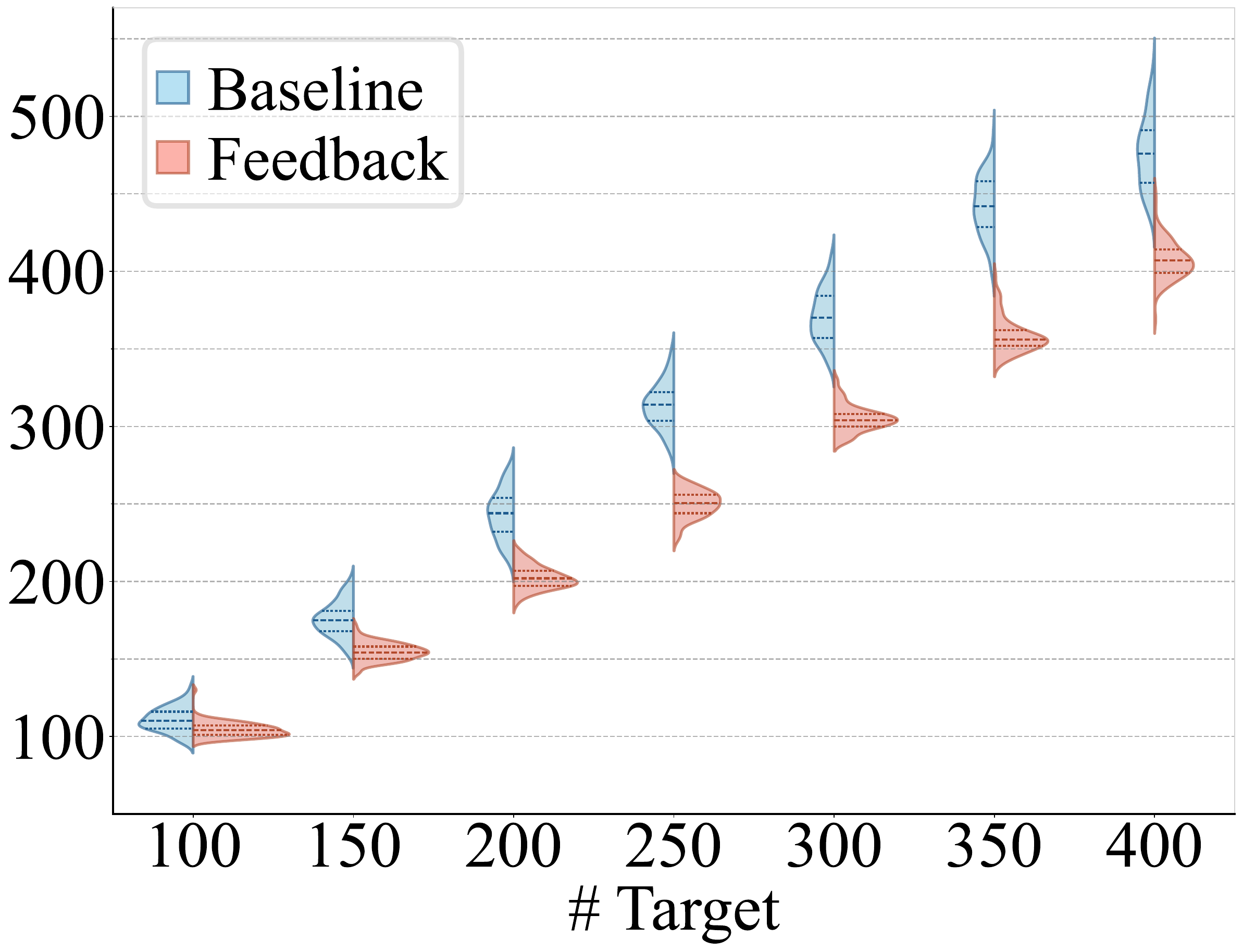} 
        \caption{Word}
        \label{subfig:bio_qwen3_8B_prompt_b}
    \end{subfigure}
    \hfill
    \begin{subfigure}{.32\textwidth}
        \includegraphics[width=\linewidth]{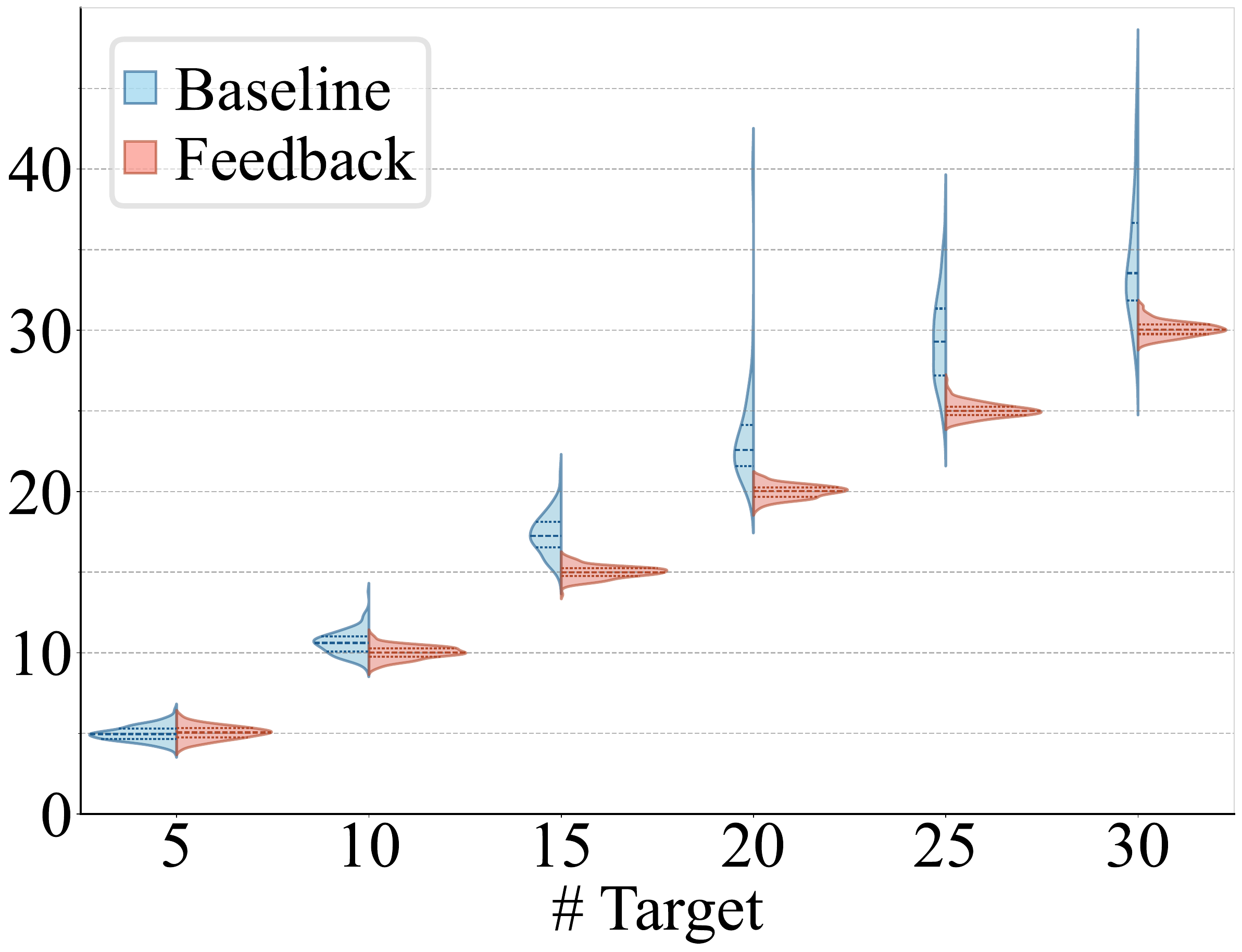} 
        \caption{Sentence}
        \label{subfig:bio_qwen3_8B_prompt_c}
    \end{subfigure}
    \begin{subfigure}{.340\textwidth}
        \includegraphics[width=\linewidth]{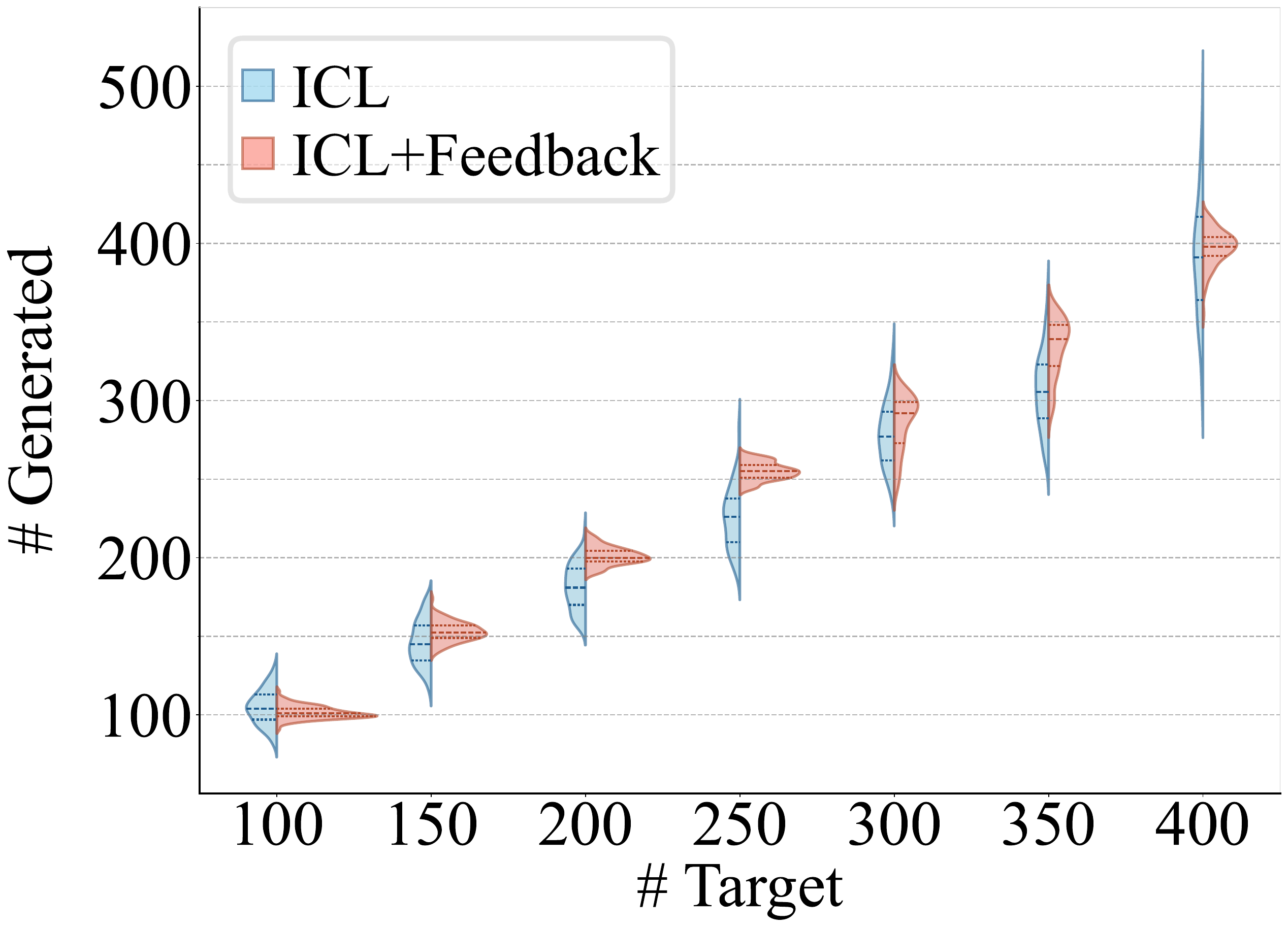} 
        \caption{Token (ICL)}
        \label{subfig:bio_qwen3_8B_icl_a}
    \end{subfigure}
    \hfill
    \begin{subfigure}{.32\textwidth}
        \includegraphics[width=\linewidth]{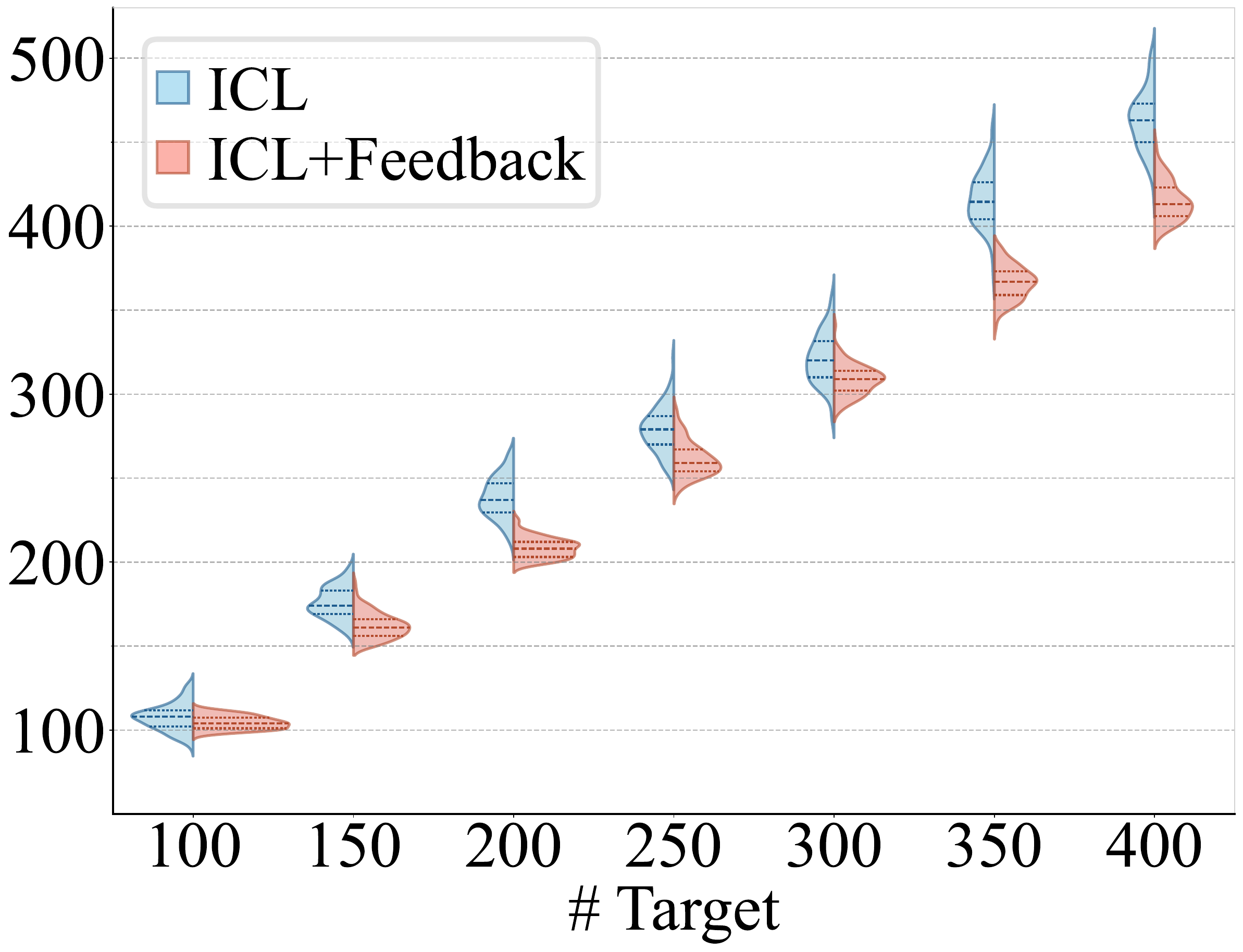} 
        \caption{Word (ICL)}
        \label{subfig:bio_qwen3_8B_icl_b}
    \end{subfigure}
    \hfill
    \begin{subfigure}{.32\textwidth}
        \includegraphics[width=\linewidth]{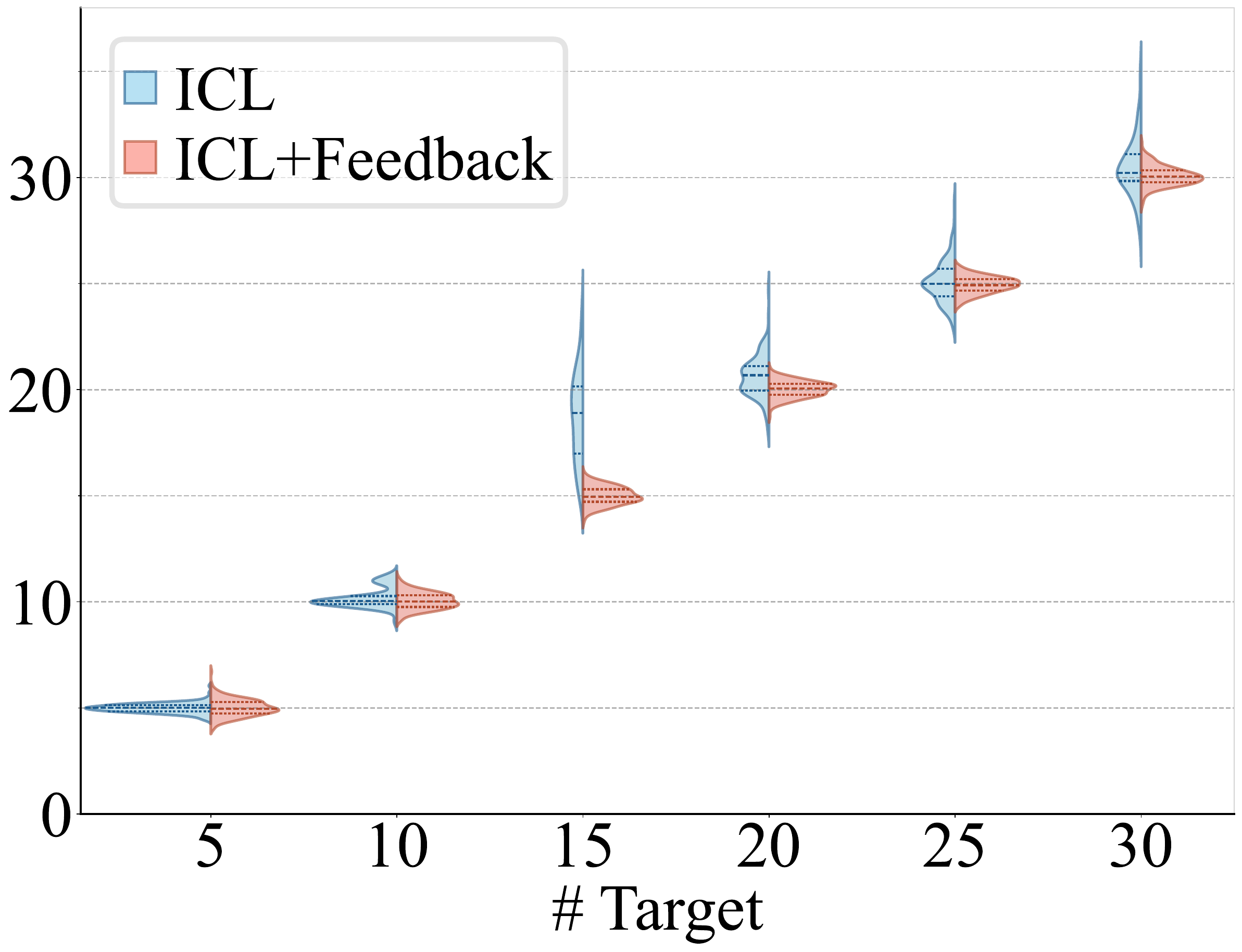} 
        \caption{Sentence (ICL)}
        \label{subfig:bio_qwen3_8B_icl_c}
    \end{subfigure}
    \caption{Generated length distributions under varying target lengths on Biographies using Qwen3-8B.}
    \label{fig:bio_qwen3_8B_align}
\end{figure*}

%% file: appendix_fig/bio_llama.tex
\begin{figure*}[!ht]
\centering
    \begin{subfigure}{.340\textwidth}
        \includegraphics[width=\linewidth]{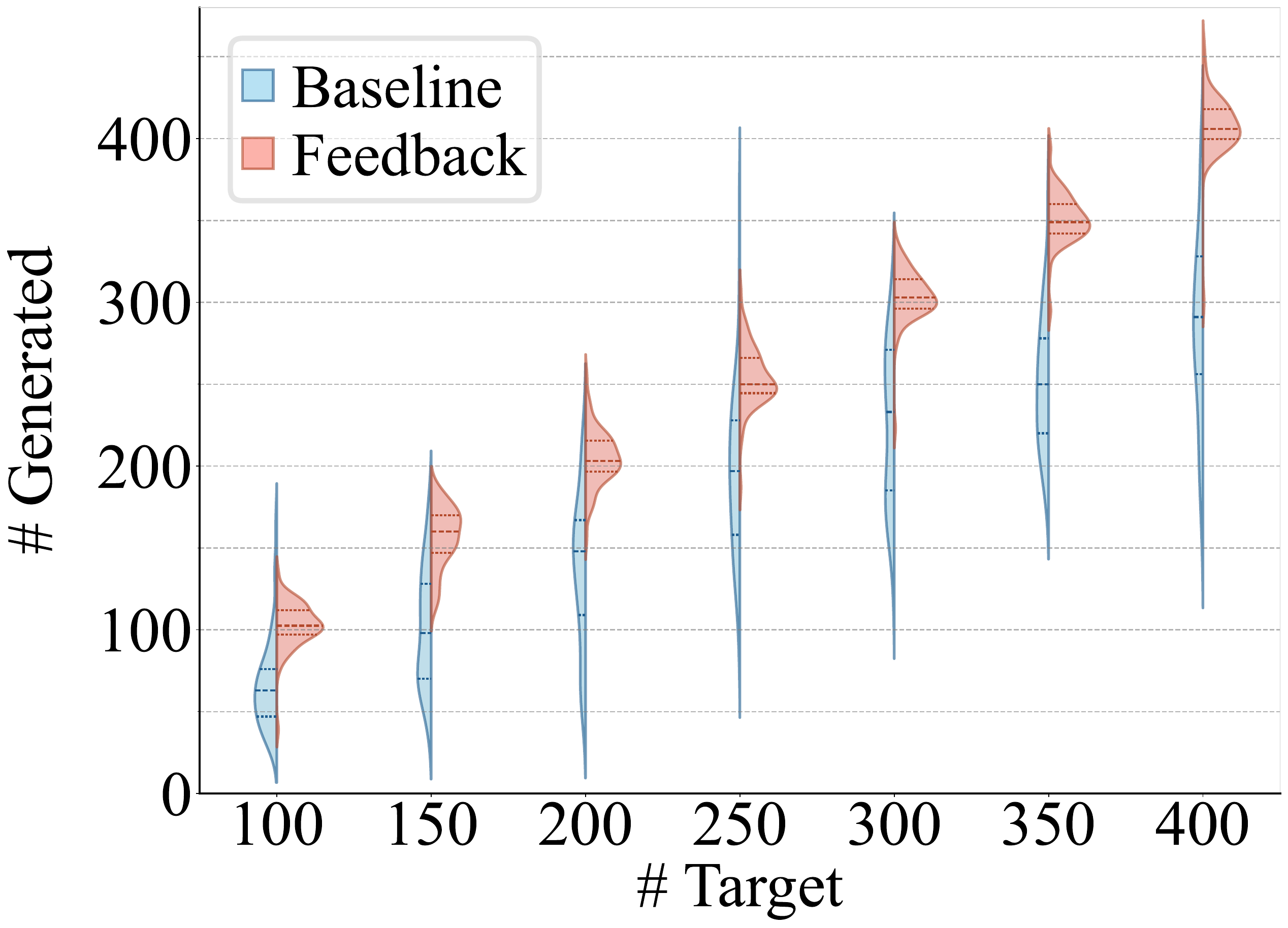} 
        \caption{Token}
        \label{subfig:bio_llama_prompt_a}
    \end{subfigure}
    \hfill
    \begin{subfigure}{.32\textwidth}
        \includegraphics[width=\linewidth]{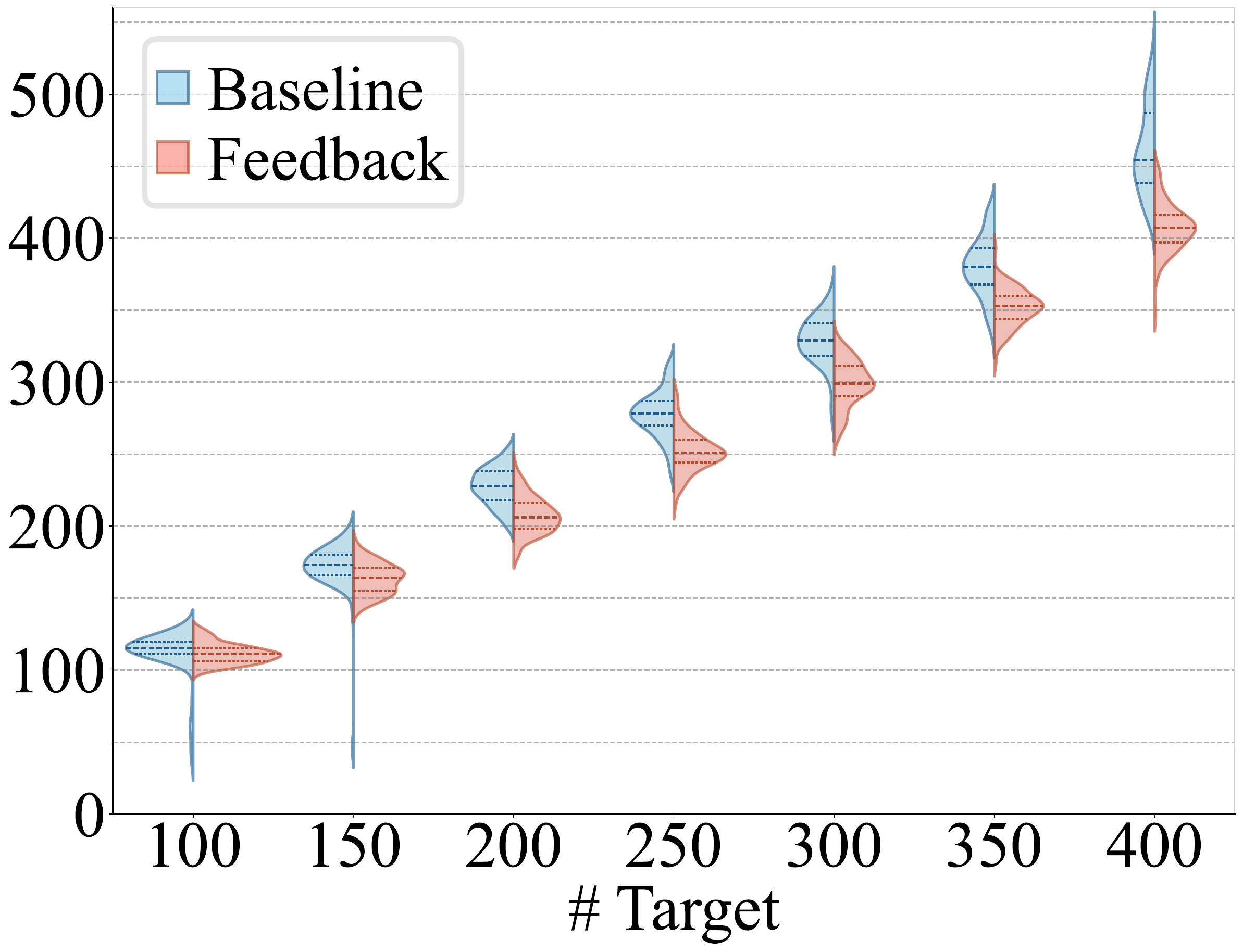} 
        \caption{Word}
        \label{subfig:bio_llama_prompt_b}
    \end{subfigure}
    \hfill
    \begin{subfigure}{.32\textwidth}
        \includegraphics[width=\linewidth]{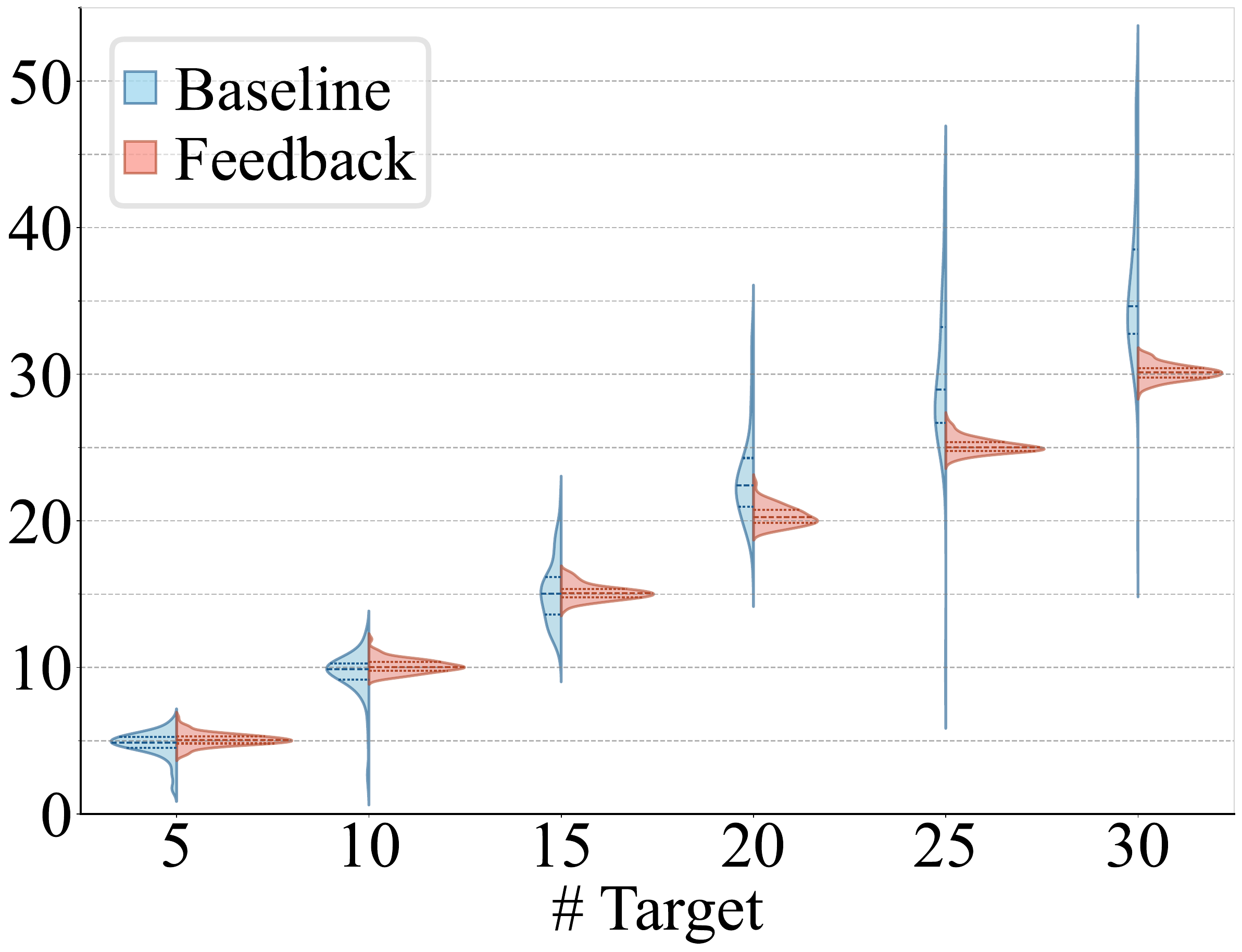} 
        \caption{Sentence}
        \label{subfig:bio_llama_prompt_c}
    \end{subfigure}
    \begin{subfigure}{.340\textwidth}
        \includegraphics[width=\linewidth]{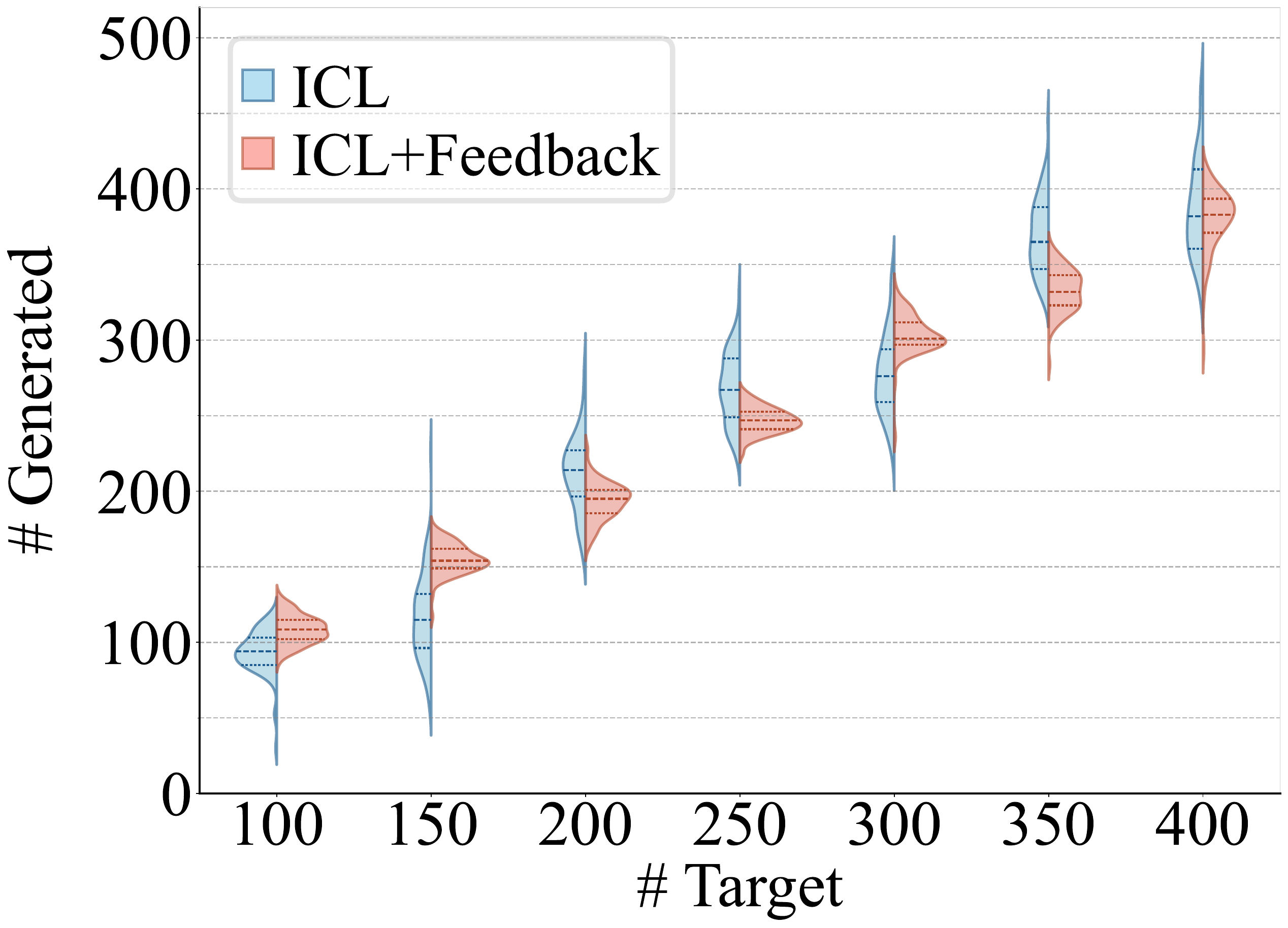} 
        \caption{Token (ICL)}
        \label{subfig:bio_llama_icl_a}
    \end{subfigure}
    \hfill
    \begin{subfigure}{.32\textwidth}
        \includegraphics[width=\linewidth]{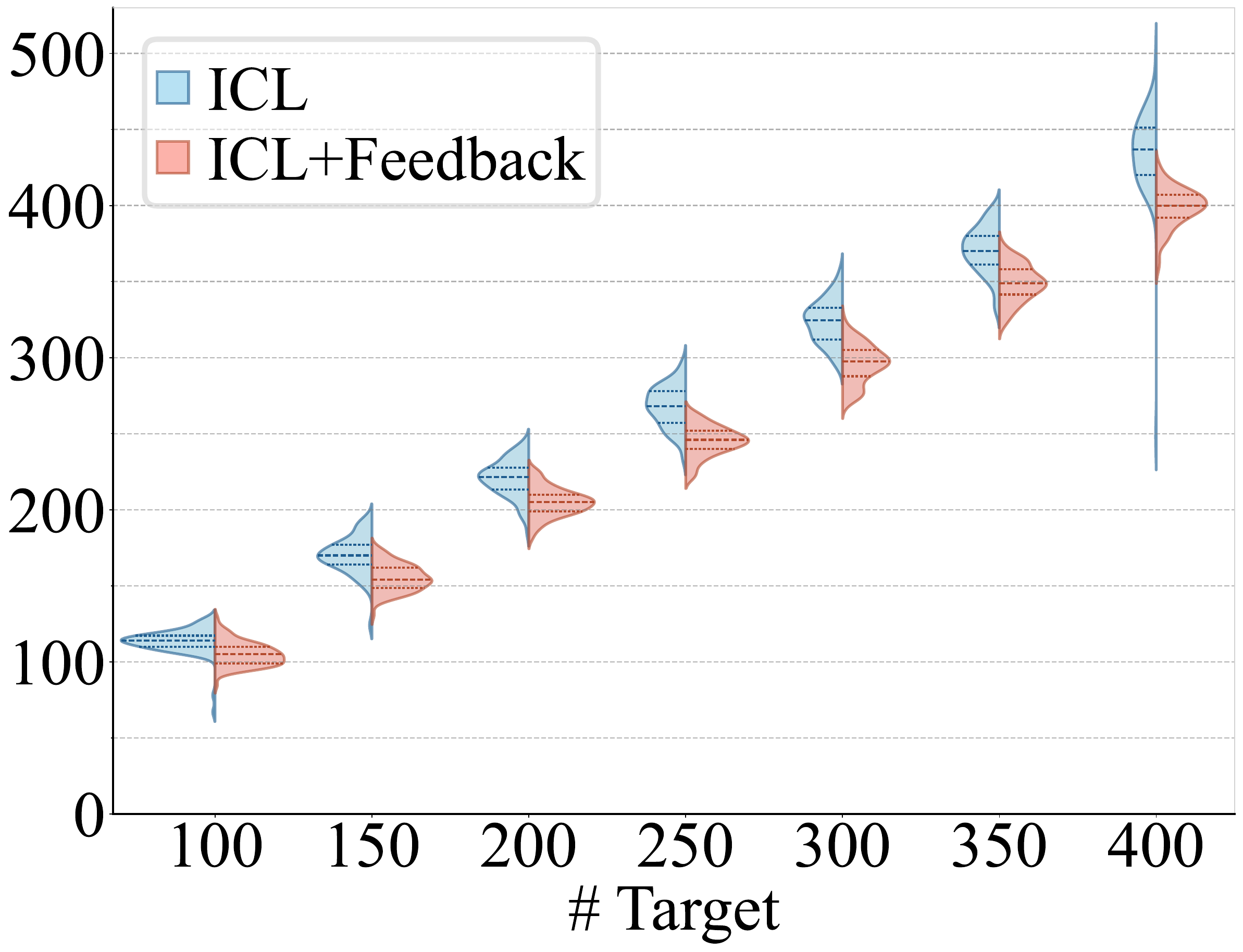} 
        \caption{Word (ICL)}
        \label{subfig:bio_llama_icl_b}
    \end{subfigure}
    \hfill
    \begin{subfigure}{.32\textwidth}
        \includegraphics[width=\linewidth]{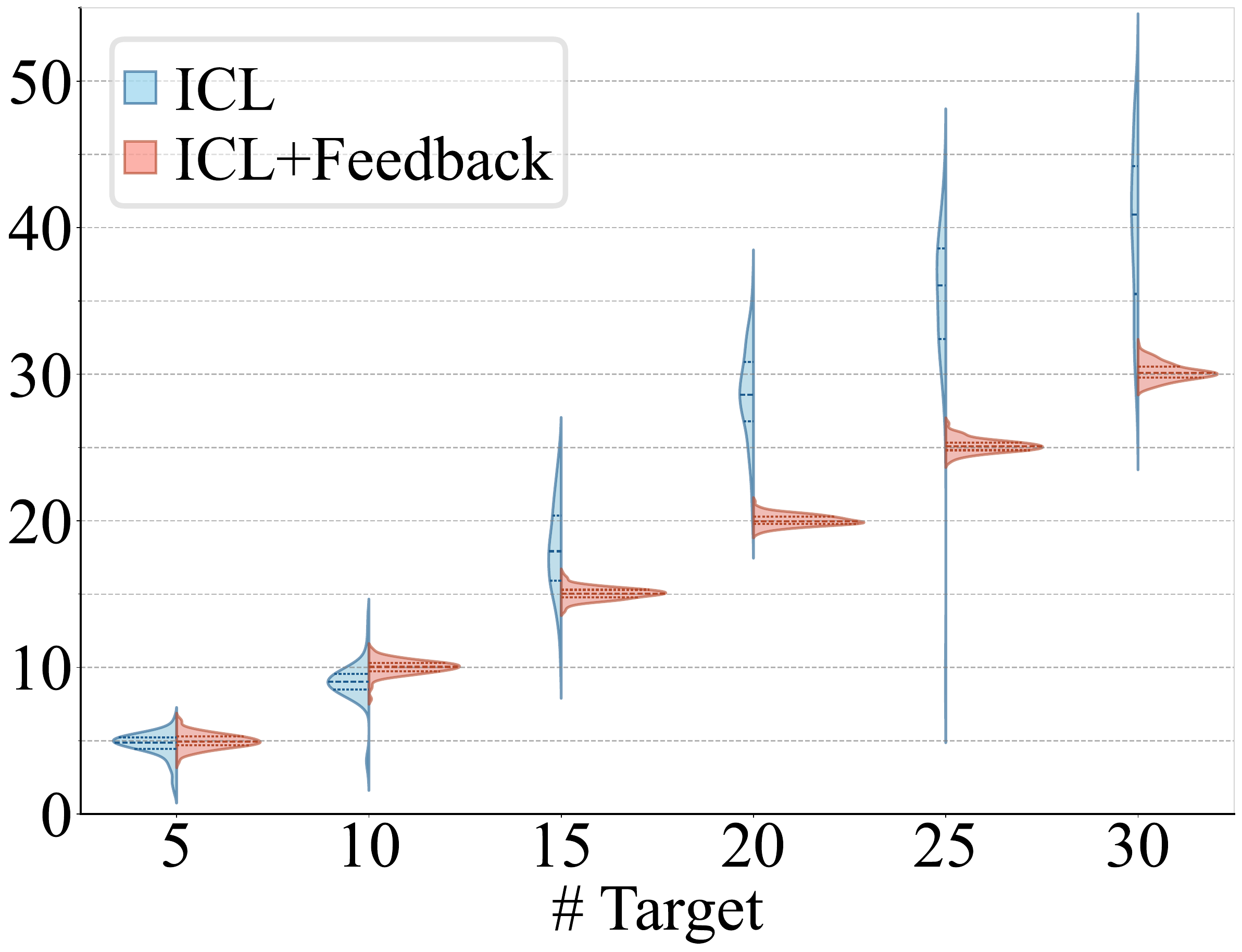} 
        \caption{Sentence (ICL)}
        \label{subfig:bio_llama_icl_c}
    \end{subfigure}
    \caption{Generated length distributions under varying target lengths on Biographies using LLaMA-3.1-8B-Instruct.}
    \label{fig:bio_llama_align}
\end{figure*}

%% file: appendix_fig/eli5_llama.tex
\begin{figure*}[!ht]
\centering
    \begin{subfigure}{.340\textwidth}
        \includegraphics[width=\linewidth]{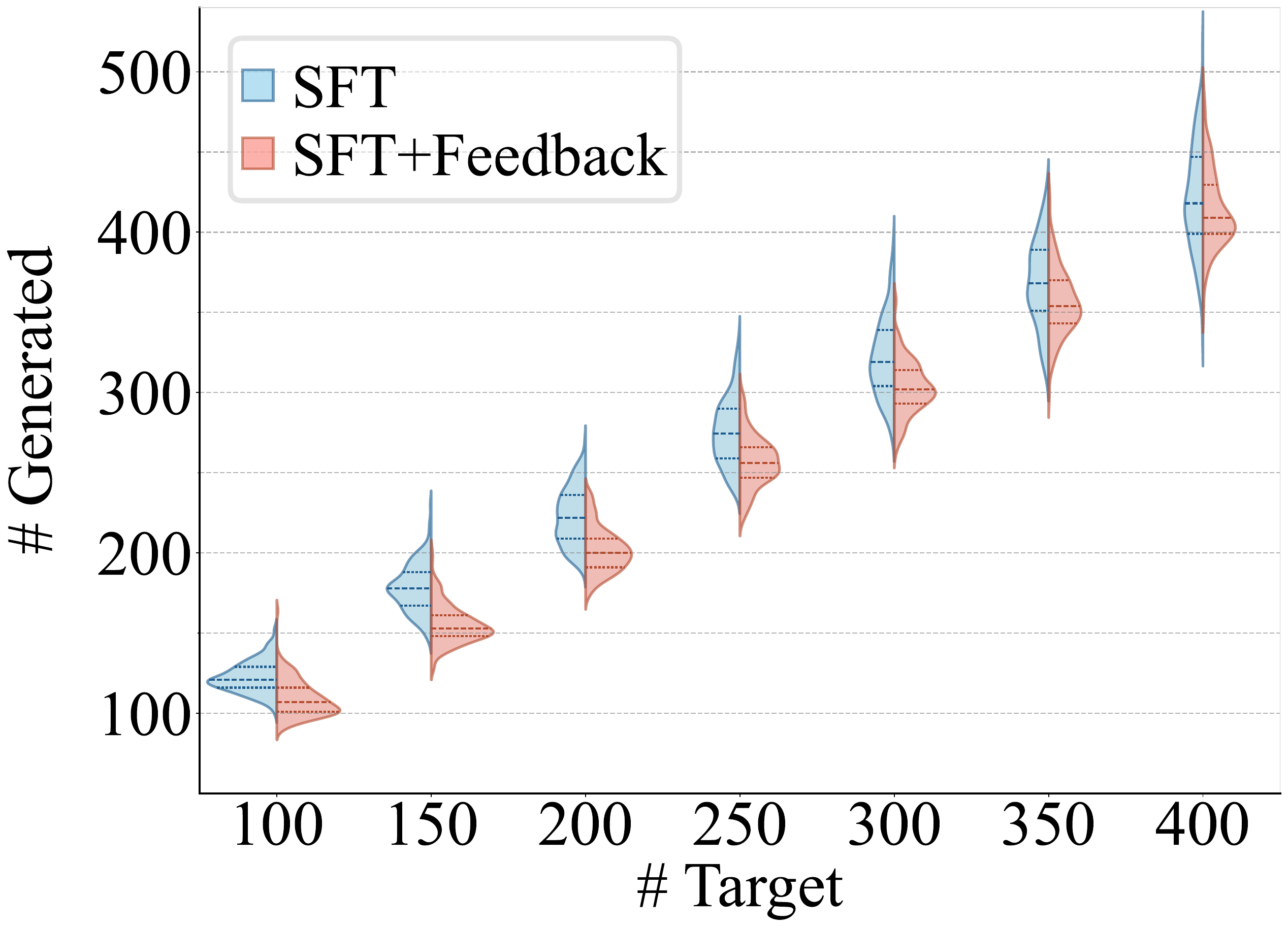} 
        \caption{Token}
        \label{subfig:eli5_llama_a}
    \end{subfigure}
    \hfill
    \begin{subfigure}{.32\textwidth}
        \includegraphics[width=\linewidth]{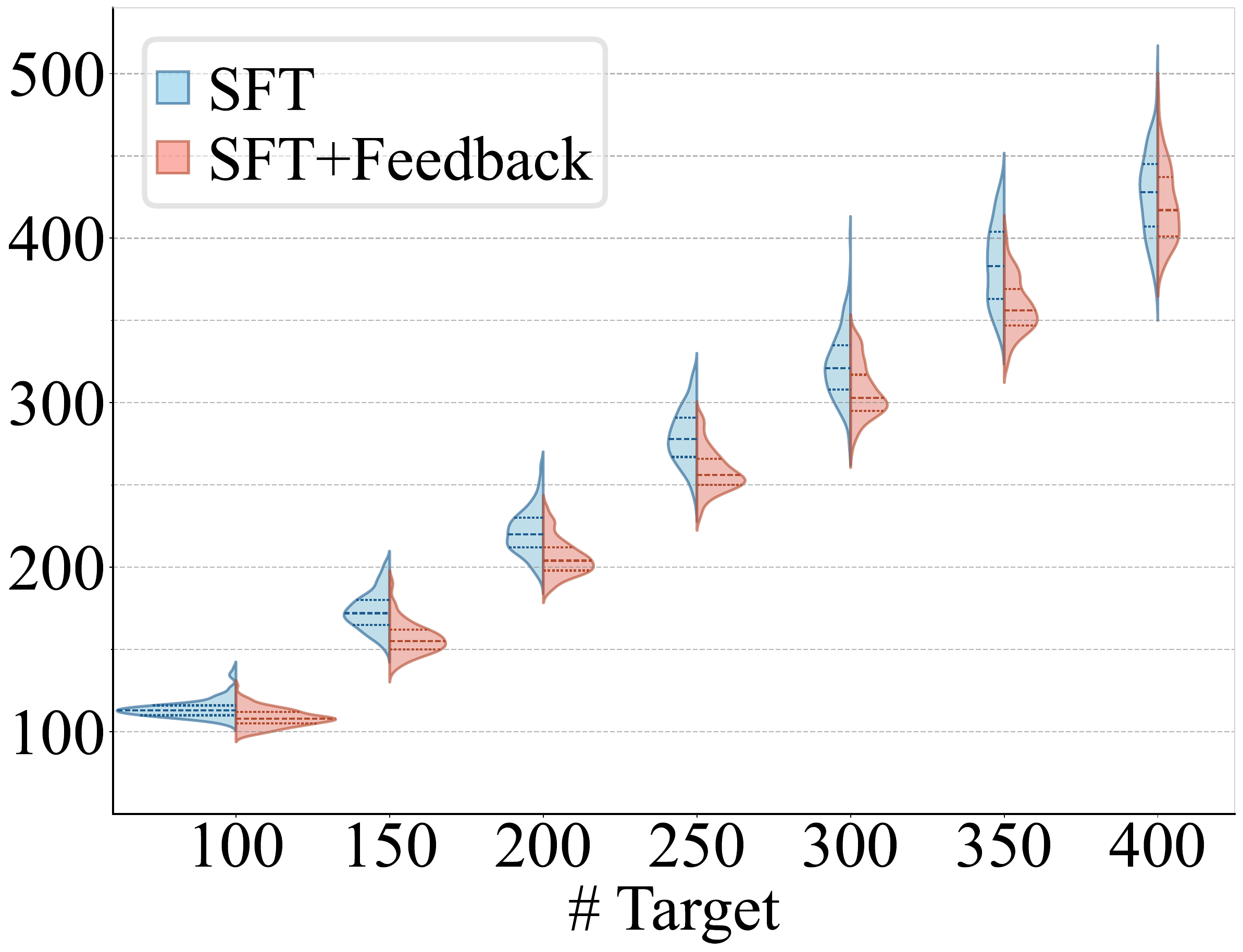} 
        \caption{Word}
        \label{subfig:eli5_llama_b}
    \end{subfigure}
    \hfill
    \begin{subfigure}{.32\textwidth}
        \includegraphics[width=\linewidth]{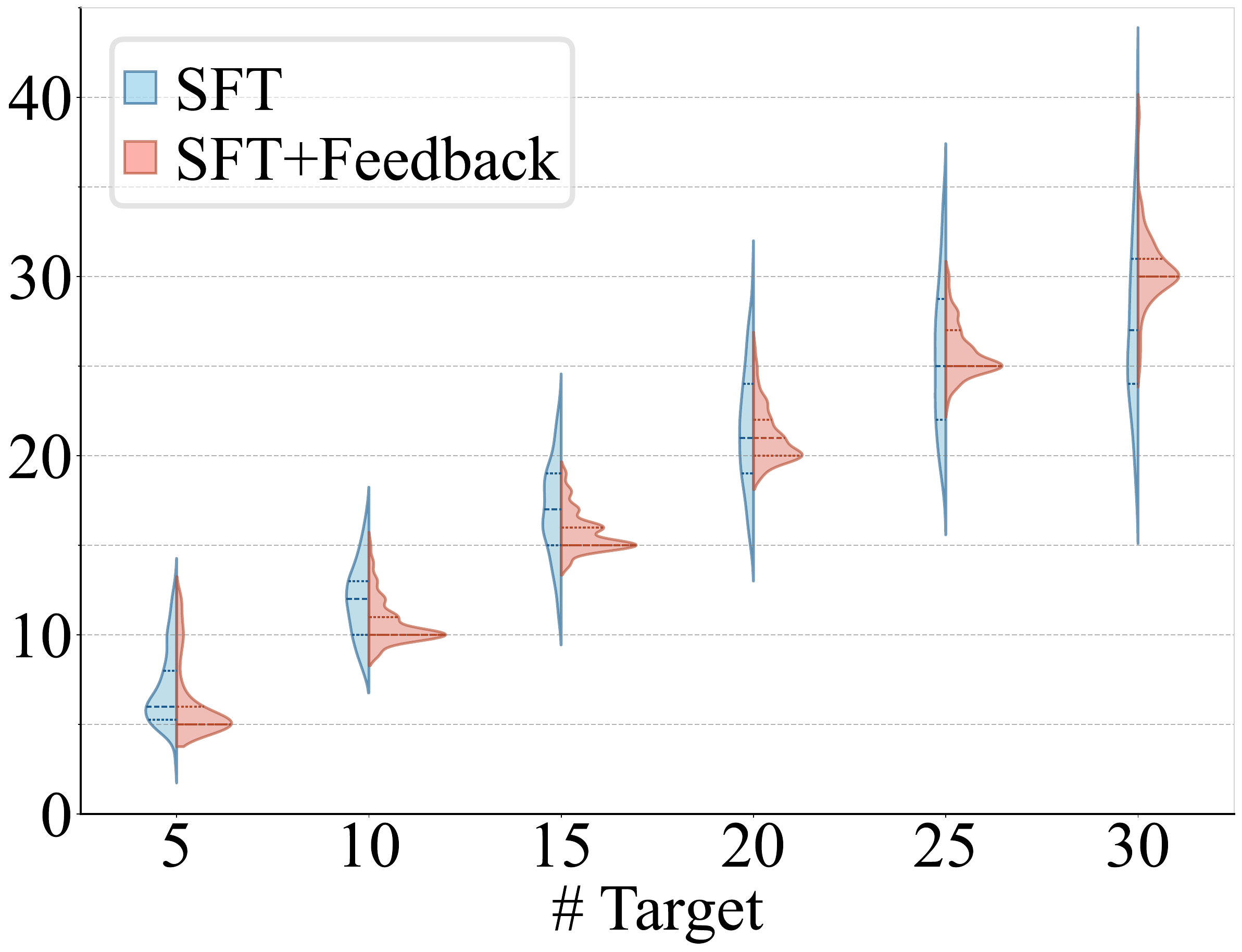} 
        \caption{Sentence}
        \label{subfig:eli5_llama_c}
    \end{subfigure}
    \caption{Generated length distributions of SFT and SFT+Feedback across varying target lengths on the ELI5 test set using LLaMA-3.1-8B-Instruct.}
    \label{fig:eli5_llama}
\end{figure*}